\documentclass[sn-mathphys]{sn-jnl}

\jyear{2021}%

\usepackage{tikz}
\usetikzlibrary{positioning,arrows,calc}
\tikzset{modal/.style={>=stealth',shorten >=1pt,shorten <=1pt,auto,node distance=1.5cm,semithick},world/.style={circle,draw,minimum size=0.5cm,fill=gray!15},point/.style={circle,draw,inner sep=0.5mm,fill=black},reflexive above/.style={->,loop,looseness=7,in=120,out=60},reflexive below/.style={->,loop,looseness=7,in=240,out=300},reflexive left/.style={->,loop,looseness=7,in=150,out=210},reflexive right/.style={->,loop,looseness=7,in=30,out=330}}

\usepackage{logicpuzzle}
\usepackage{subcaption}
\usepackage{wasysym}
\usepackage{pdfpages}

\theoremstyle{thmstyleone}%
\theoremstyle{thmstyletwo}%
\newtheorem{example}{Example}%

\theoremstyle{thmstylethree}%

\begin{document}

\title[Measuring reasoning of LLMs]{Measuring reasoning capabilities of ChatGPT}


\author*[1]{\fnm{Adrian} \sur{Groza}}\email{adrian.groza@cs.utcluj.ro}



\affil*[1]{\orgdiv{Department of Computer Science}, 
\orgname{Technical University of Cluj-Napoca}, 
\orgaddress{\street{28 Memorandumului}, 
\city{Cluj-Napoca}, 
\postcode{400114}, 
\country{Romania}}}




\abstract{
I shall quantify the logical faults generated by ChatGPT when applied to reasoning tasks. 
For experiments, I use the 144 puzzles from the library \url{https://users.utcluj.ro/~agroza/puzzles/maloga}~\cite{groza:fol}. 
The library contains puzzles of various types, including arithmetic puzzles, logical equations, Sudoku-like puzzles, zebra-like puzzles, truth-telling puzzles, grid puzzles, strange numbers, or self-reference puzzles. 
The correct solutions for these puzzles were checked using the theorem prover Prover9~\cite{mccune2005release} and the finite models finder Mace4~\cite{mccune2003mace4} based on human-modelling in Equational First Order Logic. 

A first output of this study is the benchmark of 100 logical puzzles. 
For this dataset ChatGPT provided both correct answer and justification for 7\% only. 
Since the dataset seems challenging, 
the researchers are invited to test the dataset on more advanced or tuned models than ChatGPT3.5 with more crafted prompts. 

A second output is the classification of reasoning faults conveyed by ChatGPT.  
This classification forms a basis for a taxonomy of reasoning faults generated by large language models. 
I have identified 67 such logical faults, among which: inconsistencies, implication does not hold, unsupported claim, lack of commonsense, wrong justification.
The 100 solutions generated by ChatGPT contain 698 logical faults. 
That is on average, 7 fallacies for each reasoning task.

A third ouput is the annotated answers of the ChatGPT with the corresponding logical faults. 
Each wrong statement within the ChatGPT answer was manually annotated, aiming to quantify the amount of faulty text generated by the language model. 
On average, 26.03\% from the generated text was a logical fault.

}



\maketitle


\section{A dataset for reasoning with LLMs}\label{sec1}
For experiments I started with the 144 puzzles from the MALOGA library from ~\cite{groza:fol}. 
The library contains puzzles of various types, including arithmetic puzzles, logical equations, Sudoku-like puzzles, zebra-like puzzles, truth telling puzzles, grid puzzles, strange numbers, or self-reference puzzles. 
The correct solutions were obtained using Prover9 and Mace4 based on human-modelling in Equational First Order Logic 
(FOL)\footnote{The formalisations of these puzzles in FOL is available at \url{https://users.utcluj.ro/~agroza/puzzles/maloga}}.

The aim was to investigate how many of these 144 puzzles are actually solved by LLMs. 
Part of these puzzles have graphical input that makes them difficult to be textualised for 
ChatGPT, e.g. Futoshiki, Kakurasu, Takuzo, Kakuro. After eliminating these graphical puzzles, a set of 100 logical puzzles remains (see Table~\ref{tab:dataset}).

\begin{table}
\caption{Removing puzzles that have graphical input\label{tab:dataset}}
\centering
\begin{tabular}{rlll}\hline
No &   Type &  MALOGA ~\cite{groza:fol} & Reasoning dataset\\ \hline
1 &  Micro arithmetic puzzles  & 12 & 12 \\
2 &  Strange numbers  & 12 & 12 \\
3 &  Practical puzzles   & 12 & 12 \\
4 &  Ladies and tigers     & 12 & 12 \\
5 &  Einstein puzzles        & 12 & 12 \\
6 &  Island of truth          & 12 & 12 \\
7 &  Love and marriage            & 12 & 12 \\
8 &  Grid puzzles   & 12 & ~~5 \\
9 &   Japanese puzzles               & 12 & ~~0 \\
10 &   Russian puzzles                 & 12 & ~~3 \\
11 &   Polyomino puzzles                   & 12 & ~~0 \\
12 &   Self-reference and other puzzles                     & 12 & ~~8 \\ \hline
 &   Total & 144 & 100 \\ \hline
         
\end{tabular}
\end{table}

One feature of the puzzle dataset is that it contains heterogenous exercises: 
\begin{enumerate}
\item Micro-arithmetic puzzles: in which arithmetical operations interleave with logical operations. 
These 12 puzzles are taken from Brainzilla\footnote{www.brainzilla.com}, Math is Fun\footnote{www.mathisfun.com} and 
Dudeney's collection of ``536 Puzzles and curious problems''~\citep{dudeney2016}. 

\begin{example}[Logic equation]
\textit{"In this $5\times 5$ logic equation you have to find unique integer values for the variables
$A$, $B$, $C$, $D$, $E$ - ranging from 1 to 5 - to make all statements true: (Brainzilla, www.brainzilla.com)}"
\begin{center}
\begin{tabular}{l}
 $C = A + E$\\
 $E = B + 2$\\
 $(B * E + 3 * E) \neq B \rightarrow A * A + D > E$
\end{tabular}
\end{center}
The expectation was that ChatGPT could solve such exercises since 
(1) they interleave basic artihmetic operations with logical ones (e.g. implication) and 
(2) the domain of the variable is small (from 1 to 5 in this example). 
\end{example}

\item Strange numbers: collects 12 puzzles related to unusual numbers from Dudeney~\citep{dudeney2016}, Kordemsky~\citep{kordemsky1992moscow}, Clessa~\citep{clessa1996math}, or from the ``Math is fun'' website
\begin{example}[Multiplication] 
\textit{"How many solutions are for: A B C D E F * 3 = B C D E F A, where each digit is distinct?
(puzzle from Math is fun - www.mathisfun.com~\cite{rod})}
Similarly, the expectation was that LLMs could handle these exercises since it requires basic arithmetic operations.
The puzzles in this chapter are more difficult than micro-arithmetic puzzles, because the domain size is larger.
\end{example}

\item Practical puzzles: this chapter groups twelve puzzles which have some practical touch. 
\begin{example}[Golomb ruler]
\textit{"Define a ruler with $M = 4$ marks (e.g. $a$, $b$, $c$, $d$) so that the distances between any two marks are different. 
Your ruler should be able to measure all the integer distances up to length $L = 6$.
There  should be only one way of measuring an integer distance with your ruler."}

\tikzset{
  ruler from/.initial=0,
  ruler to/.initial=10,
  ruler steps/.initial=10,
  ruler step semi/.initial=5,
  every ruler picture x/.style={x=1cm},
  ruler/.is choice,
  ruler/cm/.style={
    every ruler picture x/.style={x=1cm},
    ruler steps=10,
    ruler step semi=5},
  ruler/in/.style={
    every ruler picture x/.style={x=1in},
    ruler steps=8,
    ruler step semi=4}}
\makeatletter
\newcommand\Ruler[1][]{%
\begin{tikzpicture}[line cap=rect, every ruler picture x, #1]
  \pgfmathtruncatemacro\ruler@steps{\pgfkeysvalueof{/tikz/ruler steps}}
  \pgfmathtruncatemacro\ruler@steps@semi{\pgfkeysvalueof{/tikz/ruler step semi}}
  \pgfmathtruncatemacro\ruler@Start
                                {floor((\pgfkeysvalueof{/tikz/ruler from})*\ruler@steps)}
  \pgfmathtruncatemacro\ruler@End{ceil((\pgfkeysvalueof{/tikz/ruler to})*\ruler@steps)}
  \draw (\ruler@Start/\ruler@steps,0) -- (\ruler@End/\ruler@steps,0);
  \foreach \ruler@Cnt[
    evaluate={\ruler@CntMod=int(Mod(\ruler@Cnt,\ruler@steps))},
    evaluate={\ruler@CntModLength=
      ifthenelse(\ruler@CntMod==0, 9,
        ifthenelse(\ruler@CntMod==\ruler@steps@semi, 6, 3))}
  ] in {\ruler@Start,...,\ruler@End}
    \draw (\ruler@Cnt/\ruler@steps,0) -- ++(up:\ruler@CntModLength pt)
      \ifnum\ruler@CntMod=0
        node[above, text depth=+2pt, inner sep=+0pt]
          {$\pgfmathprint{int(\ruler@Cnt/\ruler@steps)}$}
      \fi;
      \end{tikzpicture}\ignorespaces}

      \begin{center}
 \Ruler[ruler=cm, ruler to=6]\\
 $Marks$:  $a$ \hspace{1cm}  $b$  \hspace{1cm}  $c$        \hspace{1cm}  $d$            
\end{center}
The puzzles in this chapter are of different difficulties, but most of them are available on the internet. 
Apart from the Golomb ruler available on Wikipedia for instance, there are well known crypto-arithmetic problems (e.g. SEND + MORE = MONEY) 
or measuring tasks (e.g. measure 3 liters with two vessels of 2 an 7 liters), whose solutions are publically available.  
\end{example}

\item Ladies and tigers: twelve puzzles appearing in the chapter Ladies or Tigers from~\cite{smullyan2009lady}. 
\begin{example}[Ninth day: three rooms]
\textit{"One room contains a lady and the other two contain tigers.
At most one of the three signs is true. 
The sign on the first room says: "A tiger is in this room".
The sign on the second room says: "A lady is in this room".
The sign on the third room says: "A tiger is in room 2".
Which door to open in order to find the lady?"}
~\cite{smullyan2009lady}
\begin{center}
\includegraphics[width=0.9\textwidth]{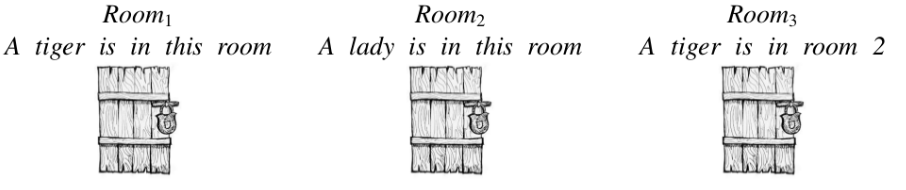} 
\end{center}
These puzzles belong to the category of truth-telling puzzles. 
They can be solved even with propositional logic. 
Theorem provers do not have difficulties to find the solution. 
\end{example}

\item Einstein puzzles: various popular Einstein or zebra puzzles collected from Dudeney~\cite{dudeney2016}, Clessa~\cite{clessa1996math}, Kordemsky~\cite{kordemsky1992moscow}, 
or from the websites Math is fun and Brainzilla.
\begin{example}[Perfect man]
\textit{"Susan's perfect man has black hair, brown eyes, and is tall and slim.
Susan knows 4 men: Arthur, Bill, Charles and Dave. 
Only one of them has all the characteristics that Susan requires.\\
1. Arthur and Bill have the same eye colour.\\
2.  Only one of the men has both black hair and brown eyes.\\
3.  Bill and Charles have the same hair colour.\\
4. ..."}\\
The difficulty of these puzzles relies on the relatevily large number of clues that need to be combined to reduce the interpretation models. 
A famous example from this category is the zebra puzzle, in which the task is to answer ``Who owns the zebra? given a set of clues.  
\end{example}

\item Island of truth: the island discovered by Smullyan~\cite{smullyan1980name} is populated by knights, knaves, or spies. 
The knights are telling the truth, while the knaves are telling lies. 
\begin{example}[We are both knaves]\label{puzzle:k1}
\textit{"On the island of knights and knaves, knights always tell the truth, while knaves always lie. 
You are approached by two people. 
The first one says: ``We are both knaves''. 
What are they actually?"} (\cite{smullyan1980name})
\begin{center}
\includegraphics[width=0.9\textwidth]{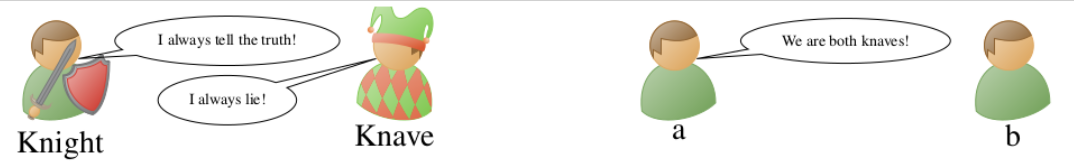} 
\end{center}
These puzzles belong to the same category of truth-telling. 
Plenty of "knight and knaves" puzzles are available in different flavors, including solutions. 
Take for example the 383 such puzzles from \url{https://philosophy.hku.hk/think/logic/knights.php}

\end{example}

\item Love and marriage: 12 puzzles on interwoven family relationships, love and marriage. 

\begin{example}[Two single persons at the end of the row]
\textit{"Four married men and three unmarried men  are seated in a row at random. 
What are the chances that the two men at the ends of the row will be single?"}
(adapted from puzzle 470 from~\cite{dudeney2016})
\begin{center}
\includegraphics[width=0.6\textwidth]{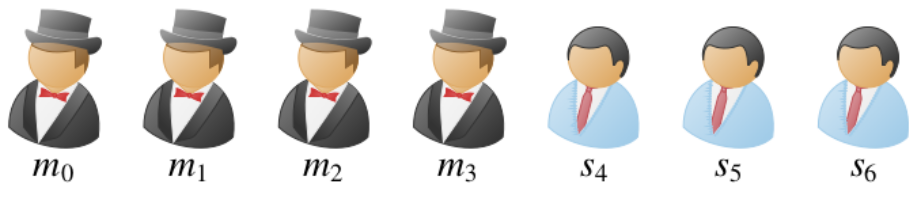} 
\end{center}
 This is an example of a probabilistic puzzle. 
 The expectation was that such puzzles are rather difficult for LLMs. 
 
\end{example}

\item Grid puzzles: 5 puzzles in the world of grids: starting with Latin or Magic squares, and continuing with Fancy queens on a chessboard.
\begin{example}[Grid puzzles]
On the left of the image below, there is the fancy queens puzzle. 
They need not to attack each other, but they are fancy: a queen cannot stay on the two main diagonals. 
I considered that the textual description and the popularity of the queens puzzle are clear enough, for the LLMs to understand the task. 
However, for most of the grid-puzzles, the textual description is not enough to properly describe the accompaning image. 
In the center of the figure below, one has to put a single star in each region so that there is only one star on each line and column. 
As the regions are highly irregular, it is difficult to provide a textual description for each region. 
Similarly, in the right part the request is to put some numbers on specific position of the star. 
Since this is also dificult to be expressed in text, such puzzles were not considered for the experiments. 
Recall in Table~\ref{tab:dataset} that only 5  from 12 puzzles were included.
\end{example}

\begin{center}
\includegraphics[width=0.8\textwidth]{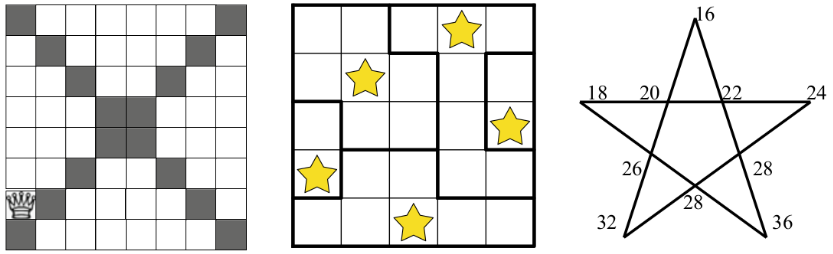} 
\end{center}

\item Japanese puzzles: these are variants of Sudoko-like puzzles. 
\begin{example}[Killer Sudoku, Kakuro]
Killer Sudoku (left image in the figure below) adds an extra constraint to the classical Sudoku: 
the sum in each irregular region should equal the number indicated. For Kakuro ((left image in the figure below) the sum of the values in each cell are indicated on the top and left margins. 
\begin{center}
\includegraphics[width=0.55\textwidth]{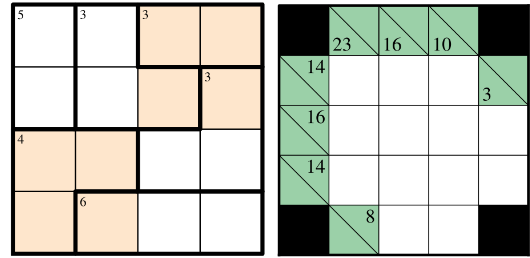} 
\end{center}
\end{example}
Since these constraints are difficult to be described in text, no Japanese puzzle was included in the dataset.

\item Russian puzzles: 3 puzzles from  Kordemsky's "Moscow puzzles"~\cite{kordemsky1992moscow} were included. 
From the available 12 puzzles, only 3 were selected. 
The removed puzzles were either grid-based or the relevant clues were mixed with a too long story. 
Such irrelvant text will introduce further difficulties to the LLM. 
A good puzzle should contain only relevant text: by removing parts of this text the puzzle could not be solved.

\item Polyomino puzzles require to arrange different two-dimensional shapes (i.e. polyomino) in a given grid. 
\begin{example}[A polyomino puzzle]
The puzzle below uses two trominoes (i.e polyomino formed by three squares), 
one domino (i.e polyomino formed by two squares), and 
one monomino (i.e. (i.e polyomino formed by one square). 
They need to be arranged in a $3\times$ 3 grid.
Since describing the shapes in text is difficult, no polyomino puzzle was selected a sinput for ChatGPT or BARD.
\end{example}

\begin{center}
\includegraphics[width=0.8\textwidth]{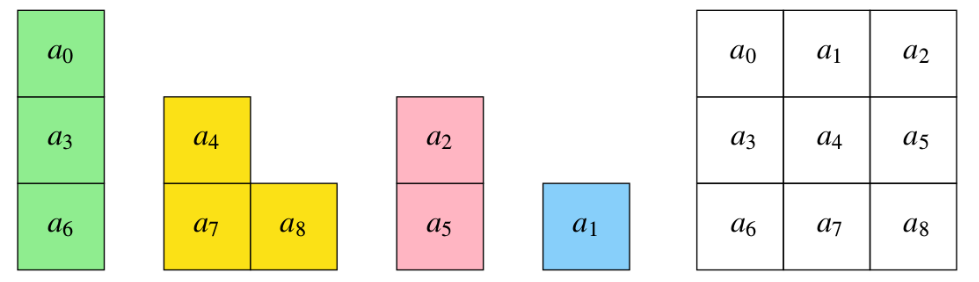} 
\end{center}

\item Self reference: 8 self-referencing brain teasers. 
\begin{example}[Self-counting sentence]\label{puzzle:self}
You have to insert numbers in the blanks to make the following  sentence true: 
{\it In this sentence, the number of occurrences of 0 is \_\_, of 1 is \_\_, of 2 is \_\_, of 3 is
\_\_, of 4 is \_\_, of 5 is \_, of 6 is \_\_, of 7 is \_\_, of 8 is \_\_, and of 9 is \_.}
There are exactly two solutions. (puzzle created by Douglas Hofstadter~\cite{hofstadter1985metamagical})
From such 12 self-referencing puzzles, 8 were included in the dataset 
\end{example}
\end{enumerate}

After this selection process, the becnhmark dataset ends-up with 100 heterogenous puzzles.
Note that most of the puzzles are available on the internet or contained in puzzles books. 
This availability created the expectance that the LLMs will benefit from it.
Recall the current media context claiming the human level capabilities of LLMs to logic and reasoning tasks. 

\section{How much hallucination?}

\subsection{The answering pattern}

\begin{figure}
\centering
\includegraphics[width=1\textwidth]{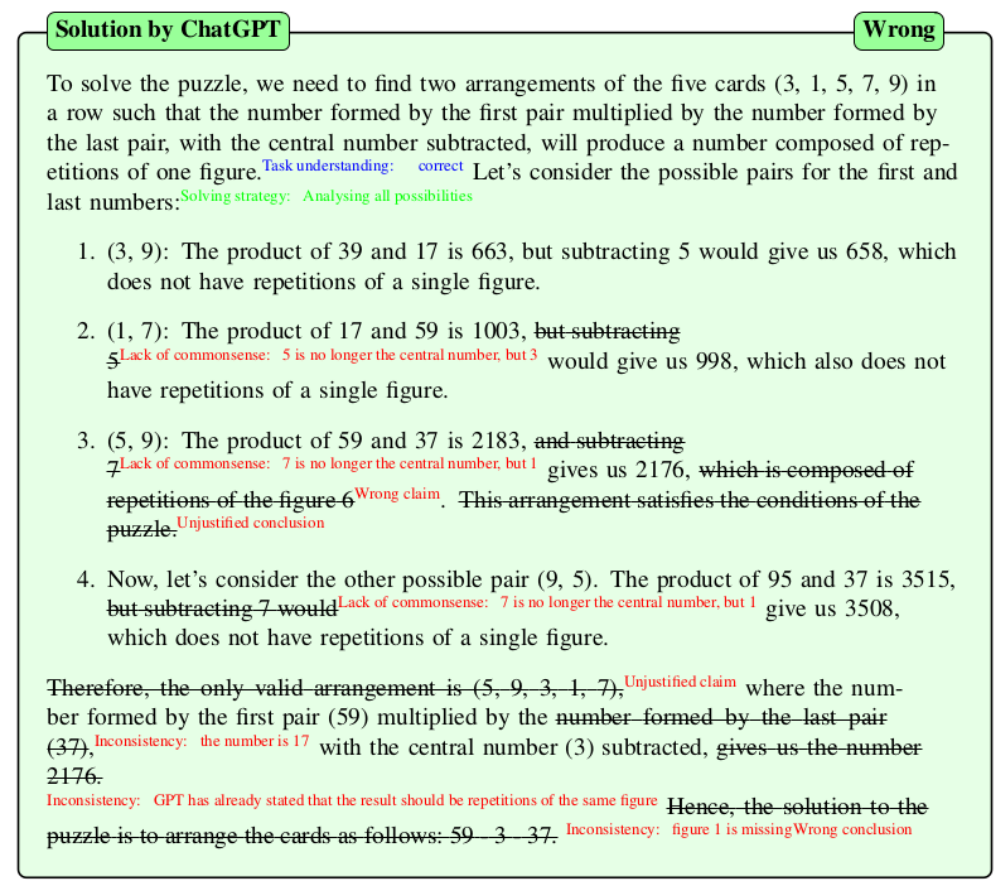} 
\caption{ChatGPT answer has 3 parts: \textit{task understanding}; (2) \textit{solving strategy}; (3) \textit{reasoning}\label{fig:answer}}
\end{figure}

A first observation is that the ChatGPT's answers follow a pattern. 
The pattern has three parts: (1) \textit{task understanding}; (2) \textit{solving strategy}; (3) \textit{reasoning}. 
For each of the 100 answers provided by ChatGPT, I annotated the text corresponding the task understanding (blue in Figure~\ref{fig:answer}, 
solving strategy with green and logical error in the reasoning phase with red. 

In the first part of the answer, i.e. \textit{task understanding}, the LLM mainy reformulates the task or problem to solve. 
Since LLMs are mastering these reformulatations, the generated text is very convincing. 
The human agent is convinced that the LLM had correctly understood the requirments. 
Additionaly to the impressive reformulation, the LLM also tries to formalise or to structure the clues included in the puzzle. 
That is, the first part of the answer creates the impression that the model is on the right truck.

The second part of the answer includes the \textit{solving strategy}. 
That is the method picked by the LLM to handle the task. 
Among the solving strategies conveyed by the model are: 
(i) analysing all possibilities; 
(ii) backward reasoning;
(iii) principle of inclusion-exclusion;
(iv) trial and error;
(v) recursion;
(vi) step by step;
(vii) backtracking;
(viii) euclidian algorithm;
(ix) Heron formula;
(x) assumption based. 
This is an impressive collection of methods, and most of them are adequate to the task. 
The are indeed adequate to the task under the premise that ChatGPT relies on computations capabilities (e.g. arithmetic, exploring the search space). 
But LLMs lack computation capabilities and the generated text is limited in size. 
Take for instance the brute-force strategy \textit{analyse all possibilities}. 
For most of the puzzles, the search space is too large to be analysed by the LLMs. 
Usually, ChatGPT starts with 1-2-3 examples (among the huge number of possibilities), 
and its conclusions are based only on these 1-2-3 random examples. 
Or take for instance the solving strategy \textit{recursion} \textit{backtracking}. 
The ChatGPT 3.5 lacks the capabilities to perform such computations. 
However, since the picked strategy is adequate in most of the cases, 
it increases the expectation that the LLM will solve the task. 

Third, the reasoning step is where ChatGPT fails in so many ways. 
Figure~\ref{fig:answer} shows examples of logical faults: \textit{lack of commonsense}, \textit{wrong claim}, \textit{unjustified conclusion}, \textit{bad arithmetic}, 
\textit{inconsistency}. 
One aim of these study was to built a taxonomy of these errors and also to quantify the generated text that is faulty. 
The following section presents the results of these quantifications.


\subsection{How many puzzles were correctly solve by ChatGPT?}
First, we need to define what ``correctly solved'' means. 
It means ywo aspects: 
(1)  the generated solution is the wright one; and
(2) the justification for the solution is also correct. 
According to these two criteria, ChatGPT correctly solved only 7 from the 100 puzles provided (see line 1 in Table~\ref{tab:answers}). 
The complexity of these correcly solved puzzles is shown by the following threee examples.

\begin{example}[Correctly solved puzzle - logic equation]
 In this $4\times 4$ logic equation you have to find unique integer values for the variables $A$, $B$,
$C$, $D$ - ranging from 1 to 4 - to make all the following statements true: $A + D = B + 4$,
$B + D = A + 2$ (puzzle taken from Brainzilla)
\end{example}

\begin{example}[Correctly solved puzzle - dividing the legacy]
A man left 100 dollars to be divided between his two sons Alfred and Benjamin's. 
If one-
third of Alfred’s legacy was taken from one-fourth of Benjamin’s, the remainder would be
11 dollars. What was the amount of each legacy? (puzzle 15 from~\cite{dudeney2016})
\end{example}

These two examples display arithmetic puzzles - the correct answers generated by ChatGPT can be consulted in the Appendix.

Line 2 in Table~\ref{tab:answers} states that for 2 puzzles the correct solution was provided, but with wrong justification. 
This can easily occur when the puzzle asks for a "yes-no" answer. 
In our case, many truth-telling puzzles ask if a sentence is True or False, 
or to decide if someone is a knight or knave. 
Since there are 50\% chances of guessing the answer, 
the solution itself without right justification should not count as properly solving the puzzle. 
Wrong justification provided by ChatGPT is actually a proof that the solution was only guessed. 
These cases often occur when testing the performance of LLMs, for instance in textual entailment tasks. 
The model is asked to distinguish between two cases: given a statement $H$ and a larger text $T$, is the claim $H$ entailed or in contradiction with the text $T$? 
Sometimes, a third option is allowed: $H$ and $T$ are unrelated. 
For such textual entailment tasks, the solution can be guessed. 

Line 3 signals a case in which the solution was correct, but there were inconsistencies in the justification. 
That is, ChatGPT generates both a statement and its logical contradiction in the same text. 

Line 4 signals that for 6 puzzles the correct solution was provided, but with no justification. 
In my view, this is not enough information to validate that ChatGPT actually solved the puzzle. 
Recall that puzzles are available on the Internet, and in some cases also their solutions. 

Line 5 indicates a puzzle solved only partially. 
Indeed, there are few puzzles that ask for two questions. 
In this case, chatGPT solved only the first one, but not the second one. 

Then, there are 72 wrong solutions supported by wrong justification (line 7). 
In one puzzle, ChatGPT seemed not to understand the task (line 8).
In 2 cases, ChatGPT claimed no solution (line 8), even if the two puzzles do have a solution. 
In 3 cases, ChatGPT claimed not enough information to solve the puzzle (line 9). 
This was not the case, since all the puzzles were completed and also validated with theorem provers. 
In one case, ChatGPT provided a fuzzy and wrong answer instead of a specific solution (line 10). 
For 3 puzzles, ChatGPT provided wrong conclusion and no justification at all (line 11). 

There are two cases in which the LLM did not provide a solution (lines 12 and 13). 
In one case, instead of a specific solution, an algorithm to solve the puzzle was provided. 
The algorithm was correct, but this is not enough to count the puzzle as ``correctly solved''. 
The most interesting answer - in my view - was the \textit{admitting-failure} one. 
ChatGPT simply states that it cannot solve that puzzle. 

Hence, the percent of corectly solved puzzles (meaning both the solution and its justification are correct) is 7\%.
In the following section, examples of answers are detailed.

\begin{table}
\centering
 \caption{Answers provided by ChatGPT\label{tab:answers}}
 \begin{tabular}{lll}\hline
 & Answer & No. of puzzles \\ \hline
1&  Correct & 7 \\ 
2&  Correct (but wrong justification) & 2\\
3&  Correct (with inconsistencies) & 1\\
4&  Correct (but unjustified) & 6\\
5&  Partial correct & 1\\ \hline
6&  Wrong & 72 \\
7&  Wrong (lack of task understanding) & 1\\
8&  Wrong (claiming no solution)& 2\\
9&  Wrong (claiming not enough information)& 3\\
10&  Wrong and fuzzy& 1\\
11&  Wrong (no justification)& 3\\ \hline
12&  No solution only valid but inneficient algorithm& 1\\
13&  Admitting failure& 1\\ \hline
&  Total & 100 \\
 \end{tabular}

\end{table}

\subsection{Quantifying logical faults}

\begin{table}
 \caption{On average, 26.03\% from the generated text is a logical fault\label{tab:faults}}
 \centering
 \begin{tabular}{lll}\hline
  Puzzle & Classified answer by ChatGPT & Amount of faulty text \\ \hline
 
 1 & Wrong & 20.36\% \\
 2 & Wrong & 20.36\% \\
 3 & Wrong (claiming no solution) & 20.36\% \\
 4 & Wrong & 20.36\% \\
 5 & Wrong & 20.36\% \\
 6 & Wrong & 20.36\% \\
 7 & Wrong (did not understand the task) & 20.36\% \\
 8 & Wrong & 20.36\% \\
 9 & Wrong & 20.36\% \\
 .. \\
 14 & Wrong & 47.03\%\\
 15 & Correct & 25.62\%\\
 .. \\ \hline
 & \textbf{Average} & \textbf{26.03\%} \\
\end{tabular}
 \end{table}

One tarket was to quantify how much from the generated text is a logical fault. 
Recall in Figure~\ref{fig:answer}, that in each answer the logical faults were annotated and classified. 

Note that logical faulty claims are different from false claims. 
As an example, consider the following chain of reasoning:
\begin{equation}
a \rightarrow b, \hspace{0.7cm}  b \rightarrow c, \hspace{0.7cm} \color{red} c \rightarrow d \color{black}, \hspace{0.7cm} d \rightarrow e, 
\hspace{0.7cm} c \rightarrow conclusion, 
\hspace{0.7cm} \color{red} conclusion \color{black}   \nonumber
\end{equation}

Assume the implication \color{red}$c \rightarrow d$ \color{black} was identified as faulty (signaled with red here). 
The following implication $d \rightarrow e$ is valid from the logical viewpoint, but it relies on a premise ($d$) supported by a faulty reasoning. 
That is, both the claims $d$ and $e$ are not supported. 
The same for $c\rightarrow conclusion$: it is logically correct, but its antecedent $c$ and its consequent $conclusion$ are false. 

I quantify only the logical fault text and not the false one. 
That is, because the aim is to assess the reasoning capabilities of chatGPT. 
Wrong conclusions are also quantified as faulty. 
For this adapted and formalised example, there are 13 red characters (i.e. logical faults).
The entire answer contains 32 characters. 
The metric quantifying logical faults will be 13/32. 
Note that the false text is larger than the one computed by the logical fault metric.

Each annotated answer was assesed against this logical fault metric. 
Table~\ref{tab:faults} shows a sample of puzzles and their percenteage of fauly text. 
On average, 26.03\% of the generated text is a logical fault. 
Of the 100 answers, there were 698 logical faults. 
That is an average of 7 fallacies for each generated solution.

\section{Towards an ontology of fallacies for LLMs}
Fallacies of human reasoning have been continously analysed in various domains, including fields like argumentation theory or cognitive biases. 
Since humans interact with LLMs, it would be usefull to have similar analyses for the fallacies of LLMs. Do  they have the same logical faults as humans?
Do they tend to fail more often with the same error, e.g.  hallucinations?

\begin{figure}
\centering
\includegraphics[width=0.48\textwidth]{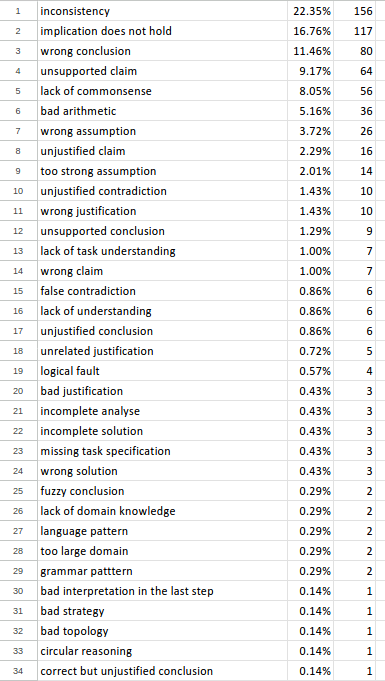} \hfill
\includegraphics[width=0.48\textwidth]{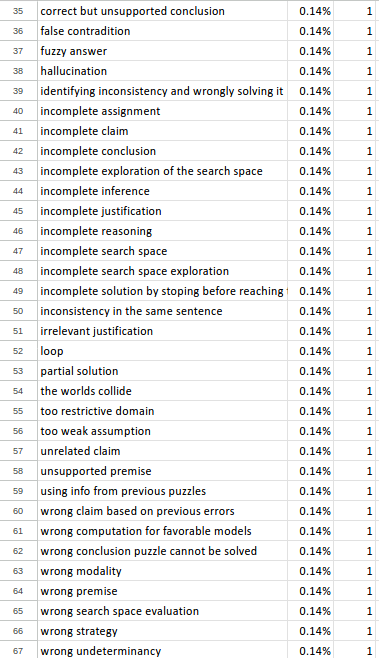} 
 
 \caption{Towards a taxonomy of hallucinations: grouping 698 logigal faulsts into 67 categries. 
 On average, each answer contains 7 logical fallacies\label{fig:tax}}
\end{figure}

Figure~\ref{fig:tax} shows the total number of logical faults identified in the 100 answers: that is 698 fallacies. 

\subsection{Inconsistencies}

Inconsistency means that within the same answer, chatGPT generates claims that contradict each other.

\begin{example}[Inconsistency]
In the crypto-arithmetic puzzle SEND + MOST = MONEY, the task was to indentify the maximum value for MONEY. 
On the one hand, ChatGPT correcly identifies that "As MONEY has five digits, ...". 
On the other hand, ChatGPT computes a solution of 4 digits: "So, we have found the best-case scenario: SEND = 1529, MOST = 6735, MONEY =
8264"
The  complete ChatGPT answer appears in the Appendix.
\end{example}

Inconsistency was most often encountered logical fallacy. 
There are 156 inconsistencies, i.e. 22.35\% from the total number of logical faults. 
In one case, ChatGPT generated an inconsistency in the same sentence. 
\subsection{Lack of commonsense}

Lack of commonsense appears in 56 cases, representing 8.05\% from  the total number of logical faults

\begin{example}[Lack of commonsense]
 Recall the same crypto-arithmetic puzzle SEND+MOST=MONEY, with the to identify the maximum value for MONEY. 
 The ChatGPT's answer contains the following text: 
 "In the worst-case scenario, you will receive \$89,675, and in the best-case scenario, you will
receive \$8,264". 
The lack of commonsense is that "the worst-case scenario" of reciveing \$89,675 is much better than the "best-case scenario" of receiving only \$8,264.
\end{example}

\subsection{No justification}

In several cases, ChatGPT provided only the solution, without any justification.
As an example, recall "Self-counting sentence" puzzle depicted in Example~\ref{puzzle:self}.

\begin{example}[No justification]\label{puzzle:self}
The sentence can be made true in the following two ways:
Solution 1: ”In this sentence, the number of occurrences of 0 is 2, of 1 is 2, of 2 is 1, of 3 is
1, of 4 is 1, of 5 is 1, of 6 is 1, of 7 is 1, of 8 is 1, and of 9 is 1.” 
Solution 2: ”In this sentence, the number of occurrences of 0 is 4, of 1 is 1, of 2 is 1, of 3 is
1, of 4 is 2, of 5 is 1, of 6 is 1, of 7 is 1, of 8 is 0, and of 9 is 1.”
\end{example}

Since, ChatGPT has been instructed from the text of the puzzle that there are two solutions, it computes two such answers.  
However, both Solution 1 and Solution 2 in Example~\ref{puzzle:self} are wrong.
The correct solutions would be $\langle 1, 11, 2, 1, 1, 1, 1, 1, 1, 1\rangle$ and $\langle 1, 7, 3, 2, 1, 1, 1, 2, 1, 1\rangle$.
Note that ChatGPT just generated some values, with no justification of the reasoning steps.

In this example, the fallacy refers to the whole answer. 
However, in many cases, the lack of justification refers only to parts from the answer.
As Figure~\ref{fig:tax} shows, there are logical faults like: \textit{unjustified claim}, \textit{unjustified premise}, \textit{unjustified contradiction}, etc.
Now, there is only a list of these faults, but they can be further grouped in several clusters or taxonomy: e.g. 
\textit{unjustified claim}, where claim can be the whole answer, or part of it like premise, conclusion, contradiction etc. 


\subsection{Traces of the past}
In one case, ChatGPT seemed to use information provided in a previous puzzle. 
Consider the following truth-telling puzzle from the knights and kanves domain:

\begin{example}[Traces form the past]\label{puzzle:knights}
 On the island of knights and knaves, knights always tell the truth, while knaves always lie. 
You are approached by two people. 
The first one says: ``At least one of us is a knave''. 
What are they actually? (puzzle 28 from~\cite{smullyan1980name})
\begin{center}
\includegraphics[width=12cm]{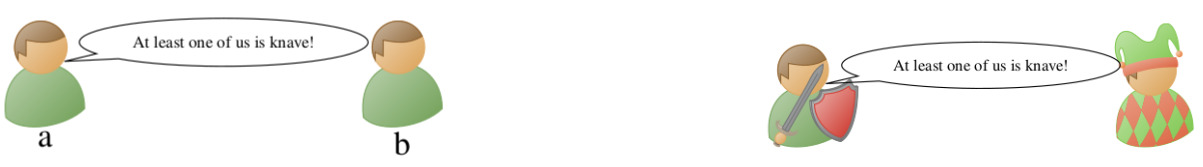} 
\end{center}
For solving this puzzle, ChatGPT has included the additional information that "one agent says - We are both knaves".
But this information does not appear in the text of the Example~\cite{puzzle:knights}. 
ChatGPT might be inventing it, or ChatGPT might take it from the previous puzzle, that is exactly the example~\ref{puzzle:k1}.
Under this assumption, ChatGPT has used information from the previous tasks. 
Hence, it is relevant that the users of LLMs to be aware of the \textit{stateless vs. statefull} modes of running an LLM. 
\end{example}

\subsection{The worlds collide}
The users might have control on the \textit{stateless vs. statefull} modes of an LLMs, but they have more difficulties to control the 
background knowledge that the LLMs are using to generate the answer. 
This is relevant since almost all puzzles have some connection with real world. 
Let the following example:

\begin{example}[The worlds collide]
There are three friends staying on the couch in Central Perk: Rachel, Ross, and
Monica. Monica is looking at Ross. Ross is looking at Rachel. Monica is
married; Rachel is not. Is a married person looking at an unmarried person?
\begin{center}
\includegraphics[width=9cm]{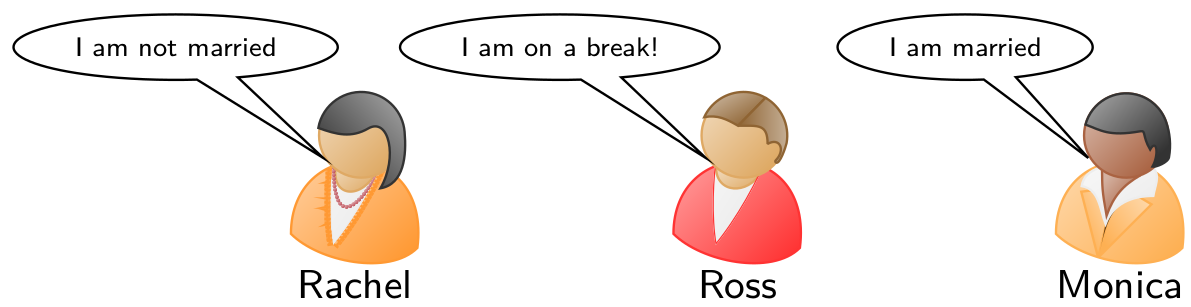} 
\end{center}
To solve this puzzle, ChatGPT includes in its answer the following information "Ross is married to Rachel". 
This clue was not provided by current puzzle, but it was (presumably) learned from the Internet: its languge model has somehow stored this information. 
\end{example}

This is an example in which ChatGPT has used information from the outside world of the puzzle, i.e. from the Friends TV series. 
This information is a kind of background knowledge. In many cases, LLMs do rely on such background knowledge. 
For our task, we need to close the world, to force the LLM to rely only on the facts  given in the task.
This can be tricky, since we also need some of the learned background knowledge by the LLMs. 

This is different when solving reasoning tasks with theorem provers. 
In case of proving, background knolwedge is by default empty. 
The user is responsable to import axiom libries or to formalse the relevant background knowledge. 
In case of LLMs, by defaults all background knowledge (i.e weights of the model) is active.

\subsection{Admitting failure - the most interesting answer}
In one case from 100 puzzles, ChatGPT admitted it is not able to find a solution. 
After some attempts, the model generated the text: "we didn't find a solution for ABCDEF = 123456 or ABCDEF =
142857. 

\begin{example}[Admitting failure]
 The task was to find a six digit number which, when multiplied by an integer between 2 and 9 inclusive,
gives the original six-digit number with its digits reversed. 
ChatGPT used a trail and error solving strategy: it started try with two random numbers. 
The strategy was wrong since random attempts in this large domain have no chances of success.
After the two random attemps, ChatGPT admited failure to solve the task. 

Why is \textit{admitting failure} important? 
If the LLM knows that its generated answer is wrong, the LLM itself can generate a second answer. 
If the LLM knows that the second answer is also wrong, it can continue the generating process. 
Paradoxically, knowing that the answer is wrong may lead theoretically to the correct answer.  
Practical constraints like time or resources may still block reaching the solution.
\end{example}

\section{Data availability}
The 100 puzzles dataset is available at \url{https://users.utcluj.ro/~agroza/puzzles/maloga/100puzzles.txt}.
The formalisation of these puzzles in First Order Logic and their solutions computed by the Prover9 theorem prover and Mace4 models finder
are available at 
\url{https://users.utcluj.ro/~agroza/puzzles/maloga/codes.html}
All the 100 answers provided by ChatGPT and the manual annotation for the logical faults are available at
\url{https://users.utcluj.ro/~agroza/puzzles/maloga/chatGPT_puzzles.pdf}

\section{Conclusions}

This study was a proper scientific exercise: like natural sciences (e.g. physics, chemistry), 
I analysed the behaviour of a creature released in the environment, i.e. ChatGPT. 
This creature belongs to the species LLMs. 
ChatGPT's behaviour was analysed in terms of solving strategy and logical flaws. 
I analysed the most encountered logical flaws on a dataset of 100 logical puzzles. 
The results are:
\begin{enumerate}
 \item 7 puzzles were correctly solved;
 \item on average, 26.03\% from the generated text is a logical fault;
 \item 698 logical faults - on average 7 fallacies/puzzle;
 \item 67 types of logical faults have been identified.
\end{enumerate}
Researchers can test the dataset against other LLMs, or with more crafted prompts.

Based on the amount of logical errors (26.03\% from the generated text was a logical fault) 
and on their types (inconsistencies, lack of commonsense, unjustified claims, the worlds collide), 
the hypothesis is that the current generation of LLMs lack reasoning skills. 

However, LLMs are good at translation. 
Instead of applying them to reasoning task, a better option would be to ask them to translate
from natural language to some logical formalism, e.g. First Order Logic. 
Then, the output of the LLMs should be given to tools specialised in reasoning, like theorem provers.

\section{Acknowledment}
I thank Iulia Cornea for helping with parts of the code.
Presentation of this study has been supported by EUProofNet Cost Action CA20111. 

\bibliography{sn-bibliography}


\begin{thebibliography}{10}
\ifx \bisbn   \undefined \def \bisbn  #1{ISBN #1}\fi
\ifx \binits  \undefined \def \binits#1{#1}\fi
\ifx \bauthor  \undefined \def \bauthor#1{#1}\fi
\ifx \batitle  \undefined \def \batitle#1{#1}\fi
\ifx \bjtitle  \undefined \def \bjtitle#1{#1}\fi
\ifx \bvolume  \undefined \def \bvolume#1{\textbf{#1}}\fi
\ifx \byear  \undefined \def \byear#1{#1}\fi
\ifx \bissue  \undefined \def \bissue#1{#1}\fi
\ifx \bfpage  \undefined \def \bfpage#1{#1}\fi
\ifx \blpage  \undefined \def \blpage #1{#1}\fi
\ifx \burl  \undefined \def \burl#1{\textsf{#1}}\fi
\ifx \doiurl  \undefined \def \doiurl#1{\url{https://doi.org/#1}}\fi
\ifx \betal  \undefined \def \betal{\textit{et al.}}\fi
\ifx \binstitute  \undefined \def \binstitute#1{#1}\fi
\ifx \binstitutionaled  \undefined \def \binstitutionaled#1{#1}\fi
\ifx \bctitle  \undefined \def \bctitle#1{#1}\fi
\ifx \beditor  \undefined \def \beditor#1{#1}\fi
\ifx \bpublisher  \undefined \def \bpublisher#1{#1}\fi
\ifx \bbtitle  \undefined \def \bbtitle#1{#1}\fi
\ifx \bedition  \undefined \def \bedition#1{#1}\fi
\ifx \bseriesno  \undefined \def \bseriesno#1{#1}\fi
\ifx \blocation  \undefined \def \blocation#1{#1}\fi
\ifx \bsertitle  \undefined \def \bsertitle#1{#1}\fi
\ifx \bsnm \undefined \def \bsnm#1{#1}\fi
\ifx \bsuffix \undefined \def \bsuffix#1{#1}\fi
\ifx \bparticle \undefined \def \bparticle#1{#1}\fi
\ifx \barticle \undefined \def \barticle#1{#1}\fi
\bibcommenthead
\ifx \bconfdate \undefined \def \bconfdate #1{#1}\fi
\ifx \botherref \undefined \def \botherref #1{#1}\fi
\ifx \url \undefined \def \url#1{\textsf{#1}}\fi
\ifx \bchapter \undefined \def \bchapter#1{#1}\fi
\ifx \bbook \undefined \def \bbook#1{#1}\fi
\ifx \bcomment \undefined \def \bcomment#1{#1}\fi
\ifx \oauthor \undefined \def \oauthor#1{#1}\fi
\ifx \citeauthoryear \undefined \def \citeauthoryear#1{#1}\fi
\ifx \endbibitem  \undefined \def \endbibitem {}\fi
\ifx \bconflocation  \undefined \def \bconflocation#1{#1}\fi
\ifx \arxivurl  \undefined \def \arxivurl#1{\textsf{#1}}\fi
\csname PreBibitemsHook\endcsname

\bibitem{groza:fol}
\begin{bbook}
\bauthor{\bsnm{Groza}, \binits{A.}}:
\bbtitle{Modelling Puzzles in First Order Logics}.
\bpublisher{Springer}, \blocation{???}
(\byear{2021}).
\doiurl{10.1007/978-3-030-62547-4}
\end{bbook}
\endbibitem

\bibitem{mccune2005release}
\begin{bchapter}
\bauthor{\bsnm{McCune}, \binits{W.}}:
\bctitle{Release of prover9}.
In: \bbtitle{Mile High Conference on Quasigroups, Loops and Nonassociative
  Systems, Denver, Colorado}
(\byear{2005})
\end{bchapter}
\endbibitem

\bibitem{mccune2003mace4}
\begin{botherref}
\oauthor{\bsnm{McCune}, \binits{W.}}:
Mace4 reference manual and guide.
arXiv preprint cs/0310055
(2003)
\end{botherref}
\endbibitem

\bibitem{dudeney2016}
\begin{bbook}
\bauthor{\bsnm{Dudeney}, \binits{H.E.}}:
\bbtitle{536 Puzzles and Curious Problems}.
\bpublisher{Courier Dover Publications}, \blocation{???}
(\byear{2016})
\end{bbook}
\endbibitem

\bibitem{kordemsky1992moscow}
\begin{bbook}
\bauthor{\bsnm{Kordemsky}, \binits{B.A.}}:
\bbtitle{The Moscow Puzzles: 359 Mathematical Recreations}.
\bpublisher{Courier Corporation}, \blocation{???}
(\byear{1992})
\end{bbook}
\endbibitem

\bibitem{clessa1996math}
\begin{bbook}
\bauthor{\bsnm{Clessa}, \binits{J.J.}}:
\bbtitle{Math and Logic Puzzles for PC Enthusiasts}.
\bpublisher{Courier Corporation}, \blocation{???}
(\byear{1996})
\end{bbook}
\endbibitem

\bibitem{rod}
\begin{botherref}
\oauthor{\bsnm{Pierce}, \binits{R.}}:
Math is fun.
Rod Pierce
(2020).
\url{www.mathsisfun.com}
\end{botherref}
\endbibitem

\bibitem{smullyan2009lady}
\begin{bbook}
\bauthor{\bsnm{Smullyan}, \binits{R.M.}}:
\bbtitle{The Lady or the Tiger?: and Other Logic Puzzles}.
\bpublisher{Courier Corporation}, \blocation{???}
(\byear{2009})
\end{bbook}
\endbibitem

\bibitem{smullyan1980name}
\begin{bbook}
\bauthor{\bsnm{Smullyan}, \binits{R.M.}}:
\bbtitle{What Is the Name of this Book? The Riddle of Dracula and Other Logical
  Puzzles}.
\bpublisher{Dover Publications}, \blocation{???}
(\byear{2011})
\end{bbook}
\endbibitem

\bibitem{hofstadter1985metamagical}
\begin{bbook}
\bauthor{\bsnm{Hofstadter}, \binits{D.R.}}:
\bbtitle{Metamagical Themes: Questions for the Es-sence of Mind and Pattern}.
\bpublisher{New York: Bantam Books}, \blocation{???}
(\byear{1985})
\end{bbook}
\endbibitem

\end{thebibliography}

\appendix 



\section*{Supplementary material}

# "Solve the following puzzle"

**by Adrian Groza and ChatGPT**

ii

# Contents







# Chapter 1

# Micro arithmetic puzzles

---

**Puzzle 1. Logic equation**

In this 4×4 logic equation you have to find unique integer values for the variables A, B, C, D - ranging from 1 to 4 - to make the following statements true: $2 \times C = B$ and $4 \times C = D$. (puzzle copyright ©Brainzilla - www.brainzilla.com)

---

**Solution by ChatGPT**                                               **Wrong**

To solve the puzzle, let's analyze the given information:

1. "$2 \times C = B$": We know that the value of $B$ is twice the value of $C$. Since the values range from 1 to 4, the only possible combinations are ($C = 1$, $B = 2$) or ($C = 2$, $B = 4$).

2. "$4 \times C = D$": We know that the value of $D$ is four times the value of $C$. Similarly, the only possible combinations are ($C = 1$, $D = 4$) or ~~($C = 2$, $D = 8$).~~ <sup>Inconsistency: GPT has stated that the values range from 1 o 4</sup>

Now, let's analyze the possible combinations:

1. If $C = 1$, then $B = 2$ (from the first equation) and $D = 4$ (from the second equation). ~~However, this combination violates the condition that all variables ($A$, $B$, $C$, $D$) must be unique.~~ <sup>Unjustified claim</sup>

2. If $C = 2$, then $B = 4$ (from the first equation) and $D = 8$ (from the second equation). This combination satisfies the condition of uniqueness.

Therefore, the solution to the puzzle is:
~~$A =$ Undetermined (can be any unique value from 1 to 4), $B = 4$, $C = 2$, $D = 8$.~~ <sup>Wrong conclusion</sup>

---

**Solution by human**

$A = 3$, $B = 2$, $C = 1$, $D = 4$.

---

**Puzzle 2. Logic equation**

In this 4×4 logic equation you have to find unique integer values for the variables $A$, $B$, $C$, $D$ - ranging from 1 to 4 - to make all the following statements true: $A + D = B + 4$, $B + D = A + 2$ (©Brainzilla - www.brainzilla.com)

**Solution by ChatGPT**                                                          **Correct**

To solve the puzzle, let's analyze the given information:
"$A + D = B + 4$": The sum of variables $A$ and $D$ is equal to the sum of variables $B$ and 4.
"$B + D = A + 2$": The sum of variables $B$ and $D$ is equal to the sum of variables $A$ and 2.
Let's proceed with the solution: First, we can rearrange the equations to make it easier to solve: From equation 1: $A + D - B = 4$. From equation 2: $B + D - A = 2$. By adding both equations together, we get:

$$
\begin{aligned}
(A + D - B) + (B + D - A) &= 4 + 2 \\
A + D + D - A - B + B &= 6 \\
2D &= 6 \\
D &= 3
\end{aligned}
$$

Now that we have found the value of $D$, we can substitute it back into either of the original equations to find the values of the other variables. Using equation 1: $A + 3 = B + 4$.
$A - B = 1$.
Since the variables $A$, $B$, $C$, and $D$ must be unique integers ranging from 1 to 4, the possible combinations for $A$ and $B$ are:
$A = 2$, $B = 1$
$A = 3$, $B = 2$
$A = 4$, $B = 3$
Since $D = 3$, we can substitute the values into equation 1 to find the value of $C$:
$A + D = B + 4$
$A + 3 = B + 4$
$A - B = 1$ The only combination that satisfies this equation is $A = 2$, $B = 1$. Therefore, $C = 4$. The solution to the puzzle is: $A = 2$, $B = 1$, $C = 4$, $D = 3$

**Solution by human**

$A = 2$, $B = 1$, $C = 4$, $D = 3$.

**Puzzle 3. Logic equation 5×5**

In this 5×5 logic equation you have to find unique integer values for the variables
$A$, $B$, $C$, $D$, $E$ - ranging from 1 to 5 - to make all statements true: (©Brainzilla -
www.brainzilla.com)

$$C = A + E$$
$$E = B + 2$$
$$(B * E + 3 * E) \neq B \rightarrow A * A + D > E$$

**Solution by ChatGPT**                                                              **Wrong**

To solve the puzzle, let's analyze the given information:

1. "$C = A + E$": The value of C is equal to the sum of A and E.

2. "$E = B + 2$": The value of E is equal to B plus 2.

3. "$((B * E + 3 * E)$ not equal with $B)$ implies $(A * A + D > E)$": If the expression
   $(B * E + 3 * E)$ is not equal to $B$, then it implies that the square of $A$ plus $D$ is
   greater than $E$."^Task understanding: Seems correct

Let's proceed with the solution:
From equation 2, we can substitute the value of $E$ in terms of $B$ into equation 1:

$$C = A + (B + 2)$$
$$C = A + B + 2$$

Now, let's analyze equation 3: $((B * E + 3 * E)$ not equal with $B)$ implies $(A * A + D > E)$
Substituting $E$ with $B + 2$:

$$(((B * (B + 2)) + (3 * (B + 2))) \neq B) \; implies \; (A * A + D > (B + 2)) \tag{1.1}$$

Simplifying:
$(((B^2 + 2B) + (3B + 6))$ not equal with $B)$ implies $(A * A + D > (B + 2))$
Combining like terms:
$((B^2 + 5B + 6)$ not equal with $B)$ implies $(A * A + D > (B + 2))$
Now, let's analyze the possible combinations for $A$, $B$, $C$, $D$, and $E$ based on their unique
integer values ranging from 1 to 5:^Solving strategy: Brute force

**Solution by ChatGPT**                                                      **Wrong**

If $B = 1$: $(((1^2 + 5 + 6)$ not equal with 1) implies $(A * A + D > (1 + 2))$
(12 not equal with 1) implies $(A * A + D > 3)$
~~True implies True~~ <sup>Unjustified claim</sup>
~~The combination $(A = 2, B = 1, C = 4, D = 5, E = 3)$ satisfies all the conditions.~~ <sup>Wrong claim</sup>
If $B = 2$: $(((2^2 + 5 * 2 + 6)$ not equal with 2) implies $(A * A + D > (2 + 2))$
(20 not equal with 2) implies $(A * A + D > 4)$
~~True implies True~~ <sup>Unjustified claim</sup>
~~The combination ($A$ =Undetermined, $B = 2$, $C = Undetermined$, $D = Undetermined$,~~
~~$E = Undetermined$) satisfies the conditions,~~ <sup>Wrong claim</sup> where $A$, $C$, $D$, and $E$ can take any
unique value from 1 to 5.
If $B = 3$: $(((3^2 + 5 * 3 + 6)$ not equal with 3) implies $(A * A + D > (3 + 2))$
(30 not equal with 3) implies $(A * A + D > 5)$
~~True implies True~~ <sup>Unjustified claim</sup>
~~The combination (A=Undetermined, B=3, C=Undetermined, D=Undetermined,~~
~~E=Undetermined) satisfies the conditions,~~ <sup>Wrong claim</sup> where A, C, D, and E can take any
unique value from 1 to 5.
If $B = 4$:$(((4^2 + 5 * 4 + 6)$ not equal with 4) implies $(A * A + D > (4 + 2))$
(42 not equal with 4) implies $(A * A + D > 6)$
~~True implies True~~ <sup>Unjustified claim</sup>
~~The combination (A=Undetermined, B=4, C=Undetermined, D=Undetermined,~~
~~E=Undetermined) satisfies the conditions~~ <sup>Wrong claim</sup>, where A, C, D, and E can take any
unique value from 1 to 5.
If $B = 5$: $(((5^2 + 5 * 5 + 6) not equal with 5) implies (A * A + D > (5 + 2))$
(60 not equal with 5) implies $(A * A + D > 7)$
~~True implies True~~ <sup>Unjustified claim</sup>
~~The combination (A=Undetermined, B=5, C=Undetermined, D=Undetermined,~~
~~E=Undetermined) satisfies the conditions~~ <sup>Wrong claim</sup>, where A, C, D, and E can take any
unique value from 1 to 5.
~~Therefore, the valid combinations that satisfy all the conditions are: A = 2, B = 1, C = 4, D =~~
~~5, E = 3~~
~~A = Undetermined, B = 2, C = Undetermined, D = Undetermined, E = Undetermined~~
~~A = Undetermined, B = 3, C = Undetermined, D = Undetermined, E = Undetermined~~
~~A = Undetermined, B = 4, C = Undetermined, D = Undetermined, E = Undetermined~~
~~A = Undetermined, B = 5, C = Undetermined, D = Undetermined, E = Undetermined~~
~~In the combinations where B is not equal to 1, the variables A, C, D, and E can take any~~
~~unique value from 1 to 5.~~ <sup>Wrong conclusion</sup>

**Solution by human**

$A = 2$, $B = 1$, $C = 5$, $D = 4$, $E = 3$.

**Puzzle 4. Logic equation 9x9**

Find distinct values for the variables ranging from 1 to 9 to make all statements true.
(puzzle copyright ©Brainzilla - www.brainzilla.com)

$4*F = E$ $B = A + H$ $I = G*H$ $A + G \neq C$
$C + E \neq 12 \rightarrow G = D + E + 2$

**Solution by ChatGPT** **Wrong**

To solve the puzzle, let's analyze the given information:
1. "$4*F = E$": The value of $E$ is four times the value of $F$. 2. "$I = G*H$": The value of $I$ is equal to the product of $G$ and $H$. 3. "$(C + E)$ not equal 12 implies $G = D + E + 2$": If the sum of $C$ and $E$ is not equal to 12, then the value of $G$ is equal to the sum of $D$, $E$, and 2. 4. "$B = A + H$": The value of $B$ is equal to the sum of $A$ and $H$. 5. "$A + G$ not equal to $C$": The sum of $A$ and $G$ is not equal to the value of $C$. <sup>Task understanding:  Seems correct</sup>
Now, let's proceed with the solution: From equation 1, we know that $E$ is divisible by 4. The only possible values for $E$, within the range of 1 to 9, are 4 and 8. Since the values need to be distinct, we can assign $E = 4$ and $F = 1$.
From equation 2, we have $I = G*H$. ~~Since the values need to be distinct and within the range of 1 to 9, we can assign $G = 3$ and $H = 3$.~~<sup>Inconsistency in the same sentence</sup>
From equation 4, we have $B = A + H$. ~~We can assign $A = 2$ to satisfy this equation.~~<sup>Incomplete assignment</sup>
Now, let's analyze equation 3: If $(C + E)$ is not equal to 12, then $G = D + E + 2$. If $C + E = 12$, then $G$ can take any value since this equation doesn't impose any restrictions. If $C + E$ is not equal to 12, we can assign $C = 6$. ~~This means $D = C - E = 6 - 4 = 2$.~~<sup>Unjustified claim</sup>
Therefore, $G = D + E + 2 = 2 + 4 + 2 = 8$.
Finally, for equation 5, ~~we have $A + G$~~<sup>Lack of task understanding</sup>. With the assigned values, $A = 2$ and $G = 8$, so $A + G = 2 + 8 = 10$.
~~The solution to the puzzle is: $A = 2, B = 5, C = 6, D = 2, E = 4, F = 1, G = 8, H = 3, I = 8$, $(A + G, C) = (10, 6)$. Note that the assigned values satisfy all the given conditions and are distinct within the range of 1 to 9.~~<sup>wrong conclusion</sup>

**Solution by human**

$A = 7, \ B = 9, \ C = 8, \ D = 5, \ E = 4, \ F = 1, \ G = 3, \ H = 2, \ I = 6$

**Puzzle 5. Three squares**

Find three distinct squares such that one is the arithmetic mean of the other two. (puzzle taken from www.mathisfun.com Pierce (2020))

**Solution by ChatGPT**                                                      **Wrong**

To solve the puzzle, let's consider three distinct squares: $x^2$, $y^2$, and $z^2$, where $x$, $y$, and $z$ are positive integers.

According to the puzzle statement, one square should be the arithmetic mean of the other two squares. Mathematically, this means that one square is equal to the average of the other two squares: $x^2 = (y^2 + z^2)/2$.

To find a solution, we can try different combinations of squares and check if the equation holds true. <small>Solving strategy: : GPT correctly identifies that the numbers should be squares,</small>

<small>Solving strategy:    but than wrongly it uses the squares instead of the numbers.</small> Let's explore some possibilities:

Let's start with small squares: $x = 1$, $y = 4$, $z = 9$. In this case, $x^2 = 1$, $y^2 = 16$, and $z^2 = 81$. ~~However, 1 is not equal to the average of 16 and 81.~~<small>Lack of task understanding: 1, 4, and 9 are the squares</small> Therefore, this combination does not satisfy the condition.

Let's try another combination: $x = 1$, $y = 9$, $z = 16$. In this case, $x^2 = 1$, $y^2 = 81$, and $z^2 = 256$. ~~Again, 1 is not equal to the average of 81 and 256.~~<small>Lack of task understanding: 1, 9, and 16 are the squares</small> Therefore, this combination also does not satisfy the condition.

Let's try another combination: $x = 4$, $y = 9$, $z = 16$. In this case, $x^2 = 16$, $y^2 = 81$, and $z^2 = 256$. ~~The average of 81 and 256 is (81 + 256) / 2 = 337 / 2 = 168.5.~~<small>Lack of task understanding: 4, 9, and 16 are the squares</small> Since we are looking for integer solutions, this combination does not satisfy the condition.

~~After exploring several combinations, it appears that there are no three distinct squares where one is the arithmetic mean of the other two.~~
<small>Incomplete search space exploration: GPT used 3 examples only to generalise for all cases.</small>

~~Hence, there are no solutions to the puzzle using three distinct squares satisfying the given condition.~~ <small>Wrong conclusion: claiming no solution</small>

**Solution by human**

Mace4 outputs a single solution: $a = 1$, $b = 25$, and $c = 49$.

---

**Puzzle 6. Pocket money**

"When I got to the station this morning," said Harold Tompkins, at his club, "I found I was short of cash. I spent just one-half of what I had on my railway ticket, and then bought a nickel's worth of candy. When I got to the terminus I spent half of what I had left and ten cents for a newspaper. Then I spent half of the remainder on a bus ticket and gave fifteen cents to that old beggar outside the club. Consequently, I arrive here with this single nickel. How much did I start out with?" (puzzle 17 from (Dudeney, 2016))

---

**Solution by ChatGPT**                                    **Wrong**

Let's work backward to determine the amount Harold Tompkins started with. Starting with the single nickel, we'll trace his expenses in reverse order based on the information provided. *Solving strategy: backward reasoning - The succession of events is correctly identifed; only partially applied.*

1. Harold gave fifteen cents to the old beggar outside the club. Therefore, before giving the fifteen cents, he had 20 cents remaining.

2. Harold spent half of the remainder on a bus ticket. ~~Since he had 20 cents before that expense, he must have spent 10 cents on the bus ticket, leaving him with 10 cents.~~ *Lack of commonsense: succession of events*

3. ~~Harold spent half of what he had left on a newspaper. Since he had 10 cents before buying the newspaper, he must have spent 5 cents on the newspaper, leaving him with 5 cents.~~ *Lack of commonsense: succession of events*

4. Harold spent half of what he had left at the terminus. ~~Since he had 5 cents before that expense, he must have spent 2.5 cents at the terminus, leaving him with 2.5 cents.~~ *Lack of commonsense: succession of events*

5. ~~Harold bought a nickel's worth of candy. Since he had 2.5 cents before buying the candy, the remaining 2.5 cents is equivalent to a nickel.~~ *Lack of commonsense: succession of events*

6. Harold spent half of what he had on his railway ticket. Since he had a nickel before that expense, he must have spent 2.5 cents on the ticket, leaving him with 2.5 cents. ~~Based on the information provided, Harold Tompkins started with 2.5 cents or a nickel~~ *Inconsistency: ChatGPT has considered that a nickel is 5 cents* . *Wrong conclusion*

---

**Solution by human**

210 ↝$^{railway}$ 105 ↝$^{candy}$ 100 ↝$^{newspaper}$ 40 ↝$^{ticket}$ 20 ↝$^{beggar}$ 5

**Puzzle 7. Dividing the legacy**

A man left 100 dollars to be divided between his two sons Alfred and Benjamin. If one-third of Alfred's legacy was taken from one-fourth of Benjamin's, the remainder would be 11 dollars. What was the amount of each legacy? (puzzle 15 from (Dudeney, 2016))

**Solution by ChatGPT**                                                            **Correct**

Let's solve the puzzle step by step:
Let's assume the amount of Alfred's legacy as "x" dollars. This means Benjamin's legacy would be 100 - x dollars.
According to the given condition, if one-third of Alfred's legacy (x/3 dollars) was taken from one-fourth of Benjamin's legacy ((100 - x)/4 dollars), the remainder would be 11 dollars. Mathematically, we can represent this as: $(100 - x)/4 - (x/3) = 11$. To solve the equation, we can simplify it:

$$[(100 - x) * 3 - 4x]/12 = 11 \tag{1.2}$$
$$(300 - 3x - 4x)/12 = 11 \tag{1.3}$$
$$300 - 7x = 11 * 12 \tag{1.4}$$
$$300 - 7x = 132 \tag{1.5}$$
$$-7x = 132 - 300 \tag{1.6}$$
$$-7x = -168 \tag{1.7}$$
$$x = -168/-7 \tag{1.8}$$
$$x = 24 \tag{1.9}$$

We have found that Alfred's legacy is 24 dollars. Since the total legacy is 100 dollars, Benjamin's legacy would be 100 - 24 = 76 dollars. Therefore, Alfred's legacy is 24 dollars and Benjamin's legacy is 76 dollars.

**Solution by human**

Alfred's legacy is 24 dollars and Benjamin's legacy is 76 dollars.

## Puzzle 8. Boys and girls

Nine boys and three girls agreed to share equally their pocket money. Every boy gave an equal sum to every girl, and every girl gave another equal sum to every boy. Every child then possessed exactly the same amount. What was the smallest possible amount that each child then possessed? (puzzle 26 from (Dudeney, 2016))

## Solution by ChatGPT                                                     **Wrong**

Let's solve the puzzle step by step: We have nine boys and three girls, making a total of twelve children. Let's assume the amount of money each boy gives to each girl as "x" dollars. Similarly, each girl gives "y" dollars to each boy.

Initially, each boy has their pocket money, and each girl has their pocket money.

After the boys give money to the girls, each girl will receive 9x dollars in total (since there are nine boys). Similarly, after the girls give money to the boys, each boy will receive 3y dollars in total (since there are three girls). According to the puzzle, after the exchange, each child possesses exactly the same amount of money. To find the smallest possible amount that each child possesses, we need to find the common amount that satisfies this condition. Since there are twelve children, the total amount of money possessed by all the children should be divisible by 12.

~~The total amount of money possessed by all the children after the exchange is 9x + 3y (sum of money received by girls) + 3y (sum of money received by boys).~~ <span style="color:red">Inconsistency: GPT has just states that after the exchange each child possesses the same amount</span> ~~We need to find the smallest possible values of x and y such that 9x + 3y + 3y is divisible by 12.~~<span style="color:red">Unjustified claim</span> ~~The smallest values that satisfy this condition are x = 4 and y = 4.~~<span style="color:red">Bad arithmetic: $x = 1$ and $y = 1$ is a solution</span>

Substituting these values back, ~~each boy gives 4 dollars to each girl, and each girl gives 4 dollars to each boy.~~<span style="color:red">Lack of commonsense: arithmetic</span>

~~Therefore, the smallest possible amount that each child possesses is 4 dollars.~~

## Solution by human

Each boy had 7\$ and gave 1\$ to each girl. Each girl had 19\$ and gave 2\$ to every boy. In the end, the boys and girls will have 10\$ each.

**Puzzle 9. Robinson's age**

"How old are you, Robinson?" asked Colonel Crackham one morning. "Well, I forget exactly," was the reply; "but my brother is two years older than me; my sister is four years older than him; my mother was twenty when I was born; and I was told yesterday that the average age of the four of us is thirty-nine years." What was Robinson's age? (puzzle 45 from (Dudeney, 2016))

**Solution by ChatGPT**                                                                                         **Correct**

Let's solve the puzzle step by step: Let's assume Robinson's age as "x" years. According to the given information:

1. Robinson's brother is two years older than him, so his brother's age is (x + 2) years.

2. Robinson's sister is four years older than his brother, so his sister's age is (x + 2 + 4) = (x + 6) years.

3. Robinson's mother was twenty when he was born, so his mother's age is (x + 20) years.

The average age of the four family members (Robinson, brother, sister, and mother) is thirty-nine years: $(x + x + 2 + x + 6 + x + 20)/4 = 39$. Simplifying the equation:

$$
\begin{aligned}
(4x + 28)/4 &= 39 & (1.10)\\
4x + 28 &= 4 * 39 & (1.11)\\
4x + 28 &= 156 & (1.12)\\
4x &= 156 - 28 & (1.13)\\
4x &= 128 & (1.14)\\
x &= 128/4 & (1.15)\\
x &= 32 & (1.16)
\end{aligned}
$$

Therefore, Robinson's age is 32 years.

**Solution by human**

Mace4 finds a single model: Robinson 32, brother 34, sister 38 and mother 52

## Puzzle 10. Five cards

I have five cards bearing the figures 3, 1, 5, 7, and 9. How can I arrange them in a row so that the number formed by the first pair multiplied by the number formed by the last pair, with the central number substracted, will produce a number composed of repetitions of one figure? Thus, in the example I have shown, 31 multiplied by 79 and 5 subtracted will produce 2444, which would have been all right if that 2 had happened to be another 4. Of course, there must be two solutions, for the pairs are clearly interchangeable (puzzle 103 from (Dudeney, 2016)).

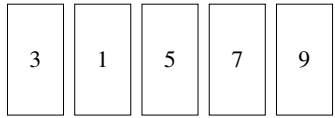

**Solution by ChatGPT**                                                    **Wrong**

To solve the puzzle, we need to find two arrangements of the five cards (3, 1, 5, 7, 9) in a row such that the number formed by the first pair multiplied by the number formed by the last pair, with the central number subtracted, will produce a number composed of repetitions of one figure.[Task understanding:    correct] Let's consider the possible pairs for the first and last numbers:[Solving strategy:  Analysing all possibilities]

1. (3, 9): The product of 39 and 17 is 663, but subtracting 5 would give us 658, which does not have repetitions of a single figure.

2. (1, 7): The product of 17 and 59 is 1003, ~~but subtracting 5~~[Lack of commonsense:  5 is no longer the central number, but 3] would give us 998, which also does not have repetitions of a single figure.

3. (5, 9): The product of 59 and 37 is 2183, ~~and subtracting 7~~[Lack of commonsense:  7 is no longer the central number, but 1] gives us 2176, ~~which is composed of repetitions of the figure 6~~[Wrong claim]. ~~This arrangement satisfies the conditions of the puzzle.~~[Unjustified conclusion]

4. Now, let's consider the other possible pair (9, 5). The product of 95 and 37 is 3515, ~~but subtracting 7 would~~[Lack of commonsense:  7 is no longer the central number, but 1] give us 3508, which does not have repetitions of a single figure.

~~Therefore, the only valid arrangement is (5, 9, 3, 1, 7),~~[Unjustified claim] where the number formed by the first pair (59) multiplied by the ~~number formed by the last pair (37),~~[Inconsistency:  the number is 17] with the central number (3) subtracted, ~~gives us the number 2176.~~

[Inconsistency:  GPT has already stated that the result should be repetitions of the same figure] ~~Hence, the solution to the puzzle is to arrange the cards as follows: 59 - 3 - 37.~~ [Inconsistency:  1 is missing][Wrong conclusion]

**Solution by human**

Mace4 finds two models: 39,157 or 57,139. In either case, the product of the two pairs 39 and 57, minus 1, results in 2,222.

**Puzzle 11. A square family**

A man had nine children, as follows:

1. all born at regular intervals, and

2. the sum of the squares of their ages was equal to the square of his own.

What was the age of each? Every age was an exact number of years. (puzzle 41 from (Dudeney, 2016))

**Solution by ChatGPT**                                              **Wrong**

To solve the puzzle, we need to find the ages of the nine children such that their ages satisfy two conditions: (1) they were all born at regular intervals, and (2) the sum of the squares of their ages is equal to the square of the man's age. Let's assume the ages of the children as consecutive integers starting from 1, and the man's age as "x" years. According to the given conditions, we can form the following equations:

$$1^2 + 2^2 + 3^2 + 4^2 + 5^2 + 6^2 + 7^2 + 8^2 + 9^2 = x^2 \qquad (1.17)$$

Simplifying the equation: $1 + 4 + 9 + 16 + 25 + 36 + 49 + 64 + 81 = x^2$, $285 = x^2$. To find the value of $x$, we take the square root of both sides: $sqrt(285) = 16.88$ ~~Since the man's age needs to be an exact number of years, we round up to the nearest whole number:~~ ~~x = 17.~~ <sup>Lack of task understanding</sup> Now, we have the age of the man as approximately 17 years. ~~To find the ages of the nine children, we can square the consecutive integers starting from 1~~ <sup>Lack of understanding: the ages are the numbers, not their squares:</sup>

$1^2 = 1$ $2^2 = 4$ $3^2 = 9$ $4^2 = 16$ $5^2 = 25$ $6^2 = 36$ $7^2 = 49$ $8^2 = 64$ $9^2 = 81$

Therefore, the ages of the nine children are: 1, 4, 9, 16, 25, 36, 49, 64, and 81. ~~In summary, the man's age is approximately 17 years, and the ages of the nine children are 1, 4, 9, 16, 25, 36, 49, 64, and 81.~~ <sup>Lack of commonsense: a man of 17 has a child of 81.Wrong conclusion</sup>

**Solution by human**

The ages of the nine children were respectively 2, 5, 8, 11, 14, 17, 20, 23, 26, and the age of the father was 48.

**Puzzle 12. Family ages**

A man and his wife had three children, John, Ben and Mary, and the difference between their parents' ages was the same as between John and Ben and between Ben and Mary. The ages of John and Ben, multiplied together, equalled the age of the father, and the ages of Ben and Mary multiplied together equalled the age of the mother. The combined ages of the family amounted to ninety years. What was the age of each person? (puzzle 37 from (Dudeney, 2016))

**Solution by ChatGPT**                                                                **Wrong**

To solve the puzzle, let's assign variables to the ages of the family members: Let's say the father's age is $F$, the mother's age is $M$, John's age is $J$, Ben's age is $B$, and Mary's age is $Y$. According to the given conditions:

1. The difference between their parents ages was the same as between John and Ben and between Ben and Mary. ~~This implies that~~ ~~$F - M = J - B = B - Y$.~~ Implication does not hold:   modulo is needed, since we do not know who is the elder one.

2. The ages of John and Ben, multiplied together, equaled the age of the father. This implies that $J * B = F$.

3. The ages of Ben and Mary, multiplied together, equaled the age of the mother. This implies that $B * Y = M$.

4. The combined ages of the family amounted to ninety years. This gives us the equation: $F + M + J + B + Y = 90$.

We now have a system of equations with four unknowns ($F$, $M$, $J$, $B$) and one additional equation:

$$\cancel{F - M = J - B = B - Y} \tag{1.18}$$
$$J * B = F \tag{1.19}$$
$$B * Y = M \tag{1.20}$$
$$F + M + J + B + Y = 90 \tag{1.21}$$

**Solution by ChatGPT**      **Wrong**

To solve the system of equations, we need to find values that satisfy all the conditions. Let's start by assuming some values and checking if they satisfy the conditions: Solving strategy: Brute force

Assuming $J = 1$, $B = 2$, and $Y = 3$:

From equation 2: $J * B = F \rightarrow 1 * 2 = F \rightarrow F = 2$

From equation 3: $B * Y = M \rightarrow 2 * 3 = M \rightarrow M = 6$

From equation 4: $F + M + J + B + Y = 2 + 6 + 1 + 2 + 3 = 14$, which is not equal to 90.

Since these values do not satisfy the conditions, let's try other values. Assuming $J = 3$, $B = 6$, and $Y = 9$:

From equation 2: $J * B = F \rightarrow 3 * 6 = F \rightarrow F = 18$

From equation 3: $B * Y = M \rightarrow 6 * 9 = M \rightarrow M = 54$

From equation 4: $F + M + J + B + Y = 18 + 54 + 3 + 6 + 9 = 90$.

These values satisfy all the conditions, and the combined ages of the family members sum up to 90 years.

~~Therefore, the age of each person is: Father (F): 18 years Mother (M): 54 years John (J): 3 years Ben (B): 6 years Mary (Y): 9 years.~~ Lack of commonsense: a father of 18 cannot have a child of 9.

Lack of commonsense: a father of 18 and mother of 54 is rather exceptional.

Inconsistency: difference between father and mother (36) is not equal with the difference between john and mary (6) or ben and mary (3)

**Solution by human**

*Mather* and *Father* are both 36 and the three children are 6.

# References

Dudeney, H. E. (2016). *536 Puzzles and curious problems*. Courier Dover Publications.

Pierce, R. (2020). Math is fun.

# Chapter 2

# Strange numbers

---

**Puzzle 13. Adding their cubes**

The numbers 407 and 370 have this peculiarity, that they exactly equal the sum of the cubes of their digits. Thus, the cube of 4 is 64, the cube of 0 is 0, and the cube of 7 is 343. Add together 64, 0 and 343, and you get 407. Again, the cube of 3 (i.e. 27), added to the cube of 7 (i.e. 343), is 370. Can you find a number not containing a zero that will work in the same way? Of course, we bar the absurd case of 1. (puzzle 143 from (Dudeney, 2016))

---

**Solution by ChatGPT**                                                    **Wrong**

To find a number that will work in the same way, where the number exactly equals the sum of the cubes of its digits and does not contain a zero, let's consider the following approach: We know that the number cannot contain a zero, so all digits in the number must be non-zero digits. Also, we need to exclude the case of 1, as stated in the puzzle. Let's consider a three-digit number with digits $x$, $y$, and $z$, where $x$, $y$, and $z$ are non-zero digits. The number can be represented as $100x + 10y + z$. The sum of the cubes of its digits can be represented as $x^3 + y^3 + z^3$. We need to find values for $x$, $y$, and $z$ such that $100x + 10y + z = x^3 + y^3 + z^3$.

To avoid excessive trial and error, let's explore the possible combinations systematically <span style="color:green">Solving strategy: Wrong strategy</span>.

If $x = 1$, then $100 + 10y + z = 1 + y^3 + z^3$. This means $99 + 10y = y^3 + z^3$. ~~Since $y$ is a non-zero digit, $y^3$ can only be 1, 8, or 27.~~<span style="color:red">Too strong assumption</span> However, none of these values satisfy the equation.

If $x = 2$, then $200 + 10y + z = 8 + y^3 + z^3$. This means $192 + 10y = y^3 + z^3$. ~~Since $y$ is a non-zero digit, $y^3$ can only be 8 or 27.~~<span style="color:red">Too strong assumption</span> However, none of these values satisfy the equation.

If $x = 3$, then $300 + 10y + z = 27 + y^3 + z^3$. This means $273 + 10y = y^3 + z^3$. ~~Since $y$ is a non-zero digit, $y^3$ can only be 8 or 27.~~<span style="color:red">Too strong assumption</span> ~~If $y^3 = 27$, then $z^3 = 273$, which is not possible as $z$ cannot be 0.~~<span style="color:red">Wrong justification: $z \neq 0$ does not imply that $z^3$ cannot be 336.</span> If $y^3 = 8$, then $z^3 = 273 + 10y - 8 = 265 + 10y$. ~~However, no combination of $y$ and $z$ (both being non-zero digits) satisfies this condition.~~ <span style="color:red">Bad arithmetic: GPT fails to infer that $y = 2$ based on its assumption $y^3 = 8$</span>

---

---

**Solution by ChatGPT**                                                          **Wrong**

If $x = 4$, then $400 + 10y + z = 64 + y^3 + z^3$. This means $336 + 10y = y^3 + z^3$. ~~Since $y$ is a non-zero digit, $y^3$ can only be 8 or 27.~~[Too strong assumption] ~~If $y^3 = 27$, then $z^3 = 336$, which is not possible as $z$ cannot be 0.~~[Wrong justification:  $z \neq 0$ does not imply that $z^3$ cannot be 336.] If $y^3 = 8$, then $z^3 = 336 + 10y - 8 = 328 + 10y$. ~~However, no combination of $y$ and $z$ (both being non-zero digits) satisfies this condition.~~[Bad arithmetic:  GPT fails to infer that $y = 2$ based on its assumption $y^3 = 8$] If $x = 5$, then $500 + 10y + z = 125 + y^3 + z^3$. This means $375 + 10y = y^3 + z^3$. ~~Since $y$ is a non-zero digit, $y^3$ can only be 8 or 27.~~[Too strong assumption] ~~If $y^3 = 27$, then $z^3 = 375$~~[Bad arithmetic:  if $y^3 = 27$, then $z^3 = 378$], ~~which is not possible as $z$ cannot be 0.~~[Bad justification] If $y^3 = 8$, then $z^3 = 375 + 10y - 8 = 367 + 10y$. ~~However, no combination of $y$ and $z$ (both being non-zero digits) satisfies this condition.~~[Bad arithmetic:  GPT fails to infer that $y = 2$ based on its assumption $y^3 = 8$]

If $x = 6$, then $600 + 10y + z = 216 + y^3 + z^3$. This means $384 + 10y = y^3 + z^3$. ~~Since $y$ is a non-zero digit, $y^3$ can only be 8 or 27.~~[Too strong assumption] ~~If $y^3 = 27$, then $z^3 = 384$~~[Bad arithmetic:  if $y^3 = 27$, then $z^3 = 387$], ~~which is not possible as $z$ cannot be 0.~~[Wrong justification] If $y^3 = 8$, then $z^3 = 384 + 10y - 8 = 376 + 10y$. ~~However, no combination of $y$ and $z$ (both being non-zero digits) satisfies this condition.~~[Bad arithmetic:  GPT fails to infer that $y = 2$ based on its assumption $y^3 = 8$] If $x = 7$, then $700 + 10y + z = 343 + y^3 + z^3$. This means $357 + 10y = y^3 + z^3$. ~~Since $y$ is a non-zero digit, $y^3$ can only be 8 or 27.~~[Too strong assumption] ~~If $y^3 = 27$, then $z^3 = 357$~~[Bad arithmetic:  if $y^3 = 27$, then $z^3 = 360$], ~~which is not possible as $z$ cannot be 0.~~[Bad justification] If $y^3 = 8$, then $z^3 = 357 + 10y - 8 = 349 + 10y$. ~~However, no combination of $y$ and $z$b(both being non-zero digits) satisfies this condition.~~[Bad arithmetic:  GPT fails to infer that $y = 2$ based on its assumption $y^3 = 8$] If $x = 8$, then $800 + 10y + z = 512 + y^3 + z^3$. This means $288 + 10y = y^3 + z^3$. ~~Since $y$ is a non-zero digit, $y^3$ can only be 8 or 27.~~[Too strong assumption] ~~If $y^3 = 27$, then $z^3 = 288$~~[Bad arithmetic:  if $y^3 = 27$, then $z^3 = 291$], ~~which is not possible as $z$ cannot be 0.~~[Bad justification] If $y^3 = 8$, then $z^3 = 288 + 10y - 8 = 280 + 10y$. ~~However, no combination of $y$ and $z$ (both being non-zero digits) satisfies this condition.~~[Bad arithmetic:  GPT fails to infer that $y = 2$ based on its assumption $y^3 = 8$] If $x = 9$, then $900 + 10y + z = 729 + y^3 + z^3$. This means $171 + 10y = y^3 + z^3$. Since $y$ is a non-zero digit, $y^3$ can only be 8 or 27. If $y^3 = 27$, then $z^3 = 171$, which is not possible as $z$ cannot be 0. If $y^3 = 8$, then $z^3 = 171 + 10y - 8 = 163 + 10y$. However, no combination of $y$ and $z$ (both being non-zero digits) satisfies this condition. ~~After exploring all possible combinations,~~[Wrong search space evaluation] we find that there is no number, not containing a zero, that works in the same way as 407 and 370. ~~Therefore, there is no solution to the puzzle with the given conditions.~~[Wrong conclusion]

---

**Solution by human**

We assume numbers of three digits, hence we limit the domain size to 1000. The number $X$ is formalised with $N = 100 * x + 10 * y + z$. The constraint of the puzzle says also that $N = x^3 + y^3 + z^3$ (line 7). Mace4 finds two solutions:

| Model | $x$ | $y$ | $z$ | $N$ |
|-------|-----|-----|-----|-----|
| 1     | 3   | 7   | 1   | 371 |
| 2     | 1   | 5   | 3   | 153 |

**Puzzle 14. Multiplication**

How many solutions are for:  A B C D E F * 3 = B C D E F A, where each digit is distinct? (puzzle from Math is fun - www.mathisfun.com Pierce (2020))

---

**Solution by ChatGPT**                                                                    **Wrong**

To solve the puzzle, let's consider the following steps: We have a six-digit number *ABCDEF* multiplied by 3, which gives another six-digit number *BCDEFA*. ~~Since the multiplication by 3 gives a six-digit number, we can deduce that *A* must be 1 (if *A* was greater than 1, the product would have seven digits).~~ <span style="color:red">Too restrictive domain: for variable *A*</span> Now, the equation becomes: 1*BCDEF* * 3  =  *BCDEF*1 Now, let's analyze the units digit: The units digit of 3 times *F* must be 1, which implies *F* = 7 (since 3 * 7 = 21). Now, we need to carry-over a value from the product of *E* * 3 to the tens place. ~~The possible values for E are 2, 4, and 6 (since *F* = 7 and the tens digit cannot be 0 or 1).~~

<span style="color:red">Too large domain: for variable *E* - 3E + 2 should end in 7 (i.e. *F*). Hence 3E ends in 5.</span> Let's try each possibility: If *E* = 2: 3 * 2 + carry-over = 1 ~~This implies carry-over = 0~~ <span style="color:red">Implication does not hold</span>, which means there is no carry-over. ~~The tens digit of 3 times *D* plus the carry-over must be 7~~

<span style="color:red">Bad arithmetic: 3D = 2, since E = 2,</span> ~~which implies *D* = 2~~<span style="color:red">Implication does not hold</span> (since 3 * 2 = 6 and there is no carry-over).

Now, we need to carry-over a value from the product of *C* * 3 to the hundreds place. ~~The possible values for *C* are 3, 4, 5, and 6 (since *D* = 2)~~

<span style="color:red">Too large domain: for the variable *C*. *C* can be only 4 since 3*C* ends in 2.</span> Let's try each possibility: If *C* = 3: 3 * 3 + carry-over = 2 ~~This implies carry-over = 1.~~<span style="color:red">Implication does not hold</span> ~~The hundreds digit of 3 times *B* plus the carry-over must be 2~~ <span style="color:red">Bad arithmetic: 3B ends in C, which is 2,</span> ~~which implies *B* = 5 (since 3 * 5 + carry-over = 2, and carry-over = 1).~~ <span style="color:red">Implication does not hold</span>

Finally, *A*  =  1 (already given). The solution is: A = 1, ~~B = 5, C = 3, D = 2, E = 2,~~ F = 7.<span style="color:red">Wrong solution: 2 digits are correct *A* and *F*</span>. ~~The product  1 5 3 2 2 7 * 3 = 4 6 5 6 8 1,~~ ~~which satisfies the equation  1 5 3 2 2 7 * 3 = 4 6 5 6 8 1.~~<span style="color:red">Bad arithmetic: equation is not satisfied</span> ~~Therefore, there is one solution to the puzzle with each digit being distinct.~~<span style="color:red">Wrong solution: actually there are 2 solutions, different from the wrong one computed by GPT</span>

> **Solution by human**
>
> The two solutions are: 142,857 * 3 = 428,571 and 285,714 * 3 = 857,142.

---

**Puzzle 15. The arithmetical cabby**

The driver of the taxi was wanting in civility, so Mr. Wilkins asked him for his number. "You want my number, do you?" said the driver. "Well, work it out for yourself. If you divide my number by 2, 3, 4, 5, or 6 you will find there is always 1 over; but if you divide it by 11 there is no remainder. What's more, there is no other driver with a lower number who can say the same." What was the fellow's number? (puzzle 223 from (Dudeney, 2016))

---

**Solution by ChatGPT**　　　　　　　　　　　　　　　　　　　　　　　**Wrong**

Let's solve the puzzle step by step: Let's assume the driver's number is $x$. According to the given information: When you divide the driver's number by 2, 3, 4, 5, or 6, there is always a remainder of 1. This can be represented as: $x \bmod 2 = 1$, $x \bmod 3 = 1$, $x \bmod 4 = 1$, $x \bmod 5 = 1$, $x \bmod 6 = 1$.

When you divide the driver's number by 11, there is no remainder. This can be represented as: $x \bmod 11 = 0$.

Let's find the smallest possible number that satisfies these conditions:

The number that is divisible by 2, 3, 4, 5, and 6 with a remainder of 1 can be represented as "$k$": $k = lcm(2, 3, 4, 5, 6) + 1 = 60 + 1 = 61$.

Now, let's find the smallest multiple of 11 that is greater than or equal to "$k$":

~~The smallest multiple of 11 that is greater than or equal to 61 is 66.~~
<span style="color:red">Wrong strategy: the same value 61 should be divisible by 11</span>

~~Therefore, the smallest possible number "x" that satisfies the conditions is 66.~~ <span style="color:red">wrong conclusion</span>

Let's verify: ~~66 divided by 2, 3, 4, 5, or 6 gives a remainder of 0, 0, 2, 1, and 0, respectively (not 1).~~ <span style="color:red">Inconsistency: when divided by 2,3,4,5 or 6, the remainder should be 1</span>

However, 66 divided by 11 gives no remainder, which satisfies the condition.

~~Therefore, the driver's number is 66.~~<span style="color:red">Wrong conclusion</span>

---

**Solution by human**

The first solution is 121 The next number having the same properties would be 1,331.

**Puzzle 16. The ten barrels**

A merchant had ten barrels of sugar, which he placed in the form of a pyramid. On the first row of the pyramid, there are four barrels with values 3, 5, 2, 6. On the second row of the pyramid, there are three barrels with values 4, 7, 0. On the third row of the pyramid, there are two barrerls with values 8 and 9. On the fourth row of the pyramid there is one barrel with value 1. Every barrel bore a different number from 0 to 9. It will be seen that he had accidentally arranged them so that the numbers in the three sides added up alike, that is, to 16. Can you rearrange them so that the three sides shall sum-up to the smallest number possible? Of course the central barrel (which happens to be 7 in the illustration) does not come into the count. (puzzle 462 from (Dudeney, 2016))

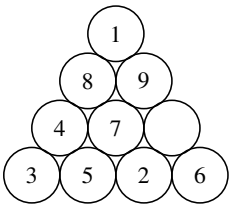    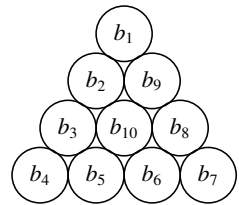

**Solution by ChatGPT**                                                              **Wrong**

To rearrange the barrels in the pyramid so that the three sides sum up to the smallest number possible, we need to find a configuration that minimizes the sum of the three sides while ensuring that each barrel bears a different number from 0 to 9. Let's start by analyzing the given configuration: 1st row: 3, 5, 2, 6 2nd row: 4, 7, 0 3rd row: 8, 9 4th row: 1 [Task understanding:  GPT fails to undrstand the task.   One difficulty comes from the textual description of a pyramid.]

To minimize the sum of the three sides, we should try to distribute the higher values among the sides with fewer barrels and the lower values among the sides with more barrels. [Solving strategy:  Wrong:  to minimise the sum the largest value (i.e.  9) should be in the middle.]   Additionally, since we have limited options for the numbers (0 to 9), we need to carefully arrange the values to find the smallest possible sum. Let's rearrange the barrels in the following way:

1st row: 0, 3, 4, 9 (Sum = 16) 2nd row: 1, 2, 7 (Sum = 10) 3rd row: 5, 6 (Sum = 11) Now, ~~we can see that the sum of the three sides is minimized~~[Lack of commonsense:  the row of a pyramid is not the side of the pyramyd], and each barrel bears a different number from 0 to 9 (excluding the central barrel).

1st row (0 + 3 + 4 + 9) = 16 2nd row (1 + 2 + 7) = 10 3rd row (5 + 6) = 11 ~~The total sum of all the barrels (excluding the central barrel)~~[Lack of commonsense:  the central barrel is not the top barrel] in the rearranged pyramid is

16 + 10 + 11 = 37, ~~which is the smallest possible sum we can achieve.~~

[Wrong conclusion:  even the wrong assumption that the task is to add the sum on each row, the minimum value is not 37 but 1+2+3+4+5+6+7+8=36]

Therefore, by rearranging the barrels as described above, ~~we achieve the smallest possible sum of the three sides in the pyramid.~~[Wrong conclusion]

**Solution by human**

Two such models are:

| Model | $b_1$ | $b_2$ | $b_3$ | $b_4$ | $b_5$ | $b_6$ | $b_7$ | $b_8$ | $b_9$ | $b_{10}$ | $p$ |
|-------|-------|-------|-------|-------|-------|-------|-------|-------|-------|----------|-----|
| 1 | 2 | 8 | 3 | 0 | 5 | 7 | 1 | 6 | 4 | 9 | 13 |
| 2 | 0 | 3 | 8 | 2 | 4 | 6 | 1 | 5 | 7 | 9 | 13 |

These two models are also depicted in Figure 2.1. For this value $p = 13$, Mace4 finds 96 models.

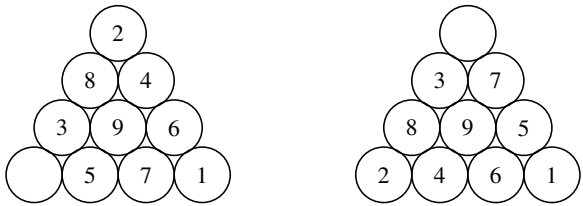

Figure 2.1: Two of the 96 solutions found by Mace4 for the minimum sum on each side $p = 13$

**Puzzle 17. Find the triangle**

The sides and height of a triangle are four consecutive whole numbers. What is the area of the triangle? (puzzle 230 from (Dudeney, 2016))

**Solution by ChatGPT**                                                    **Wrong**

Let's assume the consecutive whole numbers representing the sides and height of the triangle are $x$, $x + 1$, $x + 2$, and $x + 3$. To find the area of the triangle, we can use Heron's formula, which allows us to calculate the area of a triangle when we know the lengths of its three sides. Heron's formula is given as: <span style="color:teal">Solving strategy: Correct</span>
$Area = \sqrt{s * (s - a) * (s - b) * (s - c)}$ where $s$ is the semi-perimeter of the triangle, and $a$, $b$, and $c$ are the lengths of its sides. The semi-perimeter ($s$) is calculated as the sum of the three sides divided by 2:
~~$s = (x + x + 1 + x + 2 + x + 3)/2$~~<span style="color:red">Inconsistency: GPT assumed that one value is the height of the triangle</span>
<span style="color:red">Lack of domain knowledge: a triangle has three sides</span> $s = (4x + 6)/2$  $s = 2x + 3$
Now, we can find the area using Heron's formula:
~~$Area = \sqrt{((2x + 3) * ((2x + 3) - x) * ((2x + 3) - (x + 1)) * ((2x + 3) - (x + 2)))}$~~
<span style="color:red">Wrong assumption: we do not know which among the four consecutive values is the height</span>  Simplifying further:
$Area = \sqrt{((2x + 3) * (2x + 3 - x) * (2x + 3 - x - 1) * (2x + 3 - x - 2))}$
$Area = \sqrt{((2x + 3) * (x + 3) * (x + 2) * (x + 1))}$
Now, we can find the value of $x$. Since the sides and height of the triangle are four consecutive whole numbers, we have the following equation:
$x + (x + 1) + (x + 2) + (x + 3) = 4x + 6$<span style="color:red">Lack of domain knowledge: the hight shouldn't be added to the semipermiter</span>
Simplifying: $4x + 6 = 4x + 6$ Since this equation is true for all values of $x$, it implies that any four consecutive whole numbers can represent the sides and height of the triangle. Therefore, the area of the triangle is: $Area = \sqrt{((2x + 3) * (x + 3) * (x + 2) * (x + 1))}$
<span style="color:red">Implication does not hold</span> Remember that the actual value of the area will depend on the specific consecutive whole numbers chosen for $x$, but the formula above will give you the area for any valid set of four consecutive whole numbers.

**Solution by human**

Let $a$, $b$, $c$ the sides and $h$ its height. Mace4 finds: $a = 13$, $b = 14$, $c = 15$, $p = 21$.

**Puzzle 18. A digital difficulty**

Arrange the ten digits, 1 2 3 4 5 6 7 8 9 0, in such order that they shall form a number that may be divided by every number from 2 to 18 without in any case a remainder. As an example, if I arrange them thus, 1,274,953,680, this number can be divided by 2, 3, 4, 5, and so on up to 16, without any remainder, but it breaks down at 17. (puzzle 118 from (Dudeney, 2016))

**Solution by ChatGPT**                                                    **wrong**

To find the arrangement of the ten digits (1 2 3 4 5 6 7 8 9 0) to form a number that may be divided by every number from 2 to 18 without any remainder, ~~we need to find the smallest such number~~[Too strong assumption].

To achieve this, we can start by considering the divisibility rules for some of the smaller numbers:

Divisibility by 2: The number must end with an even digit, ~~which means it must end with 0~~[Bad arithmetic]

Divisibility by 3: The sum of the digits must be divisible by 3.

Divisibility by 4: The last two digits must form a number divisible by 4. Since the number must end with 0, the second last digit must be even (2, 4, 6, or 8).

Divisibility by 5: The number must end with 0 or 5.

Divisibility by 6: The number must satisfy both the rules of divisibility by 2 and 3.

By considering these rules, we can try to find the arrangement that satisfies all the divisibility conditions from 2 to 18.

Let's try to arrange the digits accordingly:

---

**Solution by ChatGPT**                                                    **Wrong**

~~The number must end with 0, so it will be the last digit~~[Too strong assumption].
The sum of the digits must be divisible by 3. The digits 1 to 9 sum up to 45, ~~so we need
to add one more digit from 1 to 9 to make the sum divisible by 3~~[Bad arithmetic:  45 is divisible by 3].  ~~To
minimize the number, we add 9 to the sum~~[Bad strategy:  9 is the largest number; it maximises the sum].  ~~So the
first nine digits will be: 9, 1, 2, 3, 4, 5, 6, 7, 8~~[Unsupported claim].
~~The number must end with 0 or 5~~[Inconsistency:  the current (wrong) assumption was that the number should end in 0]
[Inconsistency:  the current asumption was that the last digit should be even].  The number 5 cannot be at the end, as
it will make the number divisible by 5, but not necessarily by 10. So, the final arrange-
ment is: 9 1 2 3 4 5 6 7 8 0. Now, let's check the divisibility by other numbers:
Divisibility by 4: The last two digits are 0 and 8, ~~which form 08~~[Bad arithmetic:  the last digits are 80],
divisible by 4.
Divisibility by 7: ~~The number 9123456780 is divisible by 7~~[Bad arithmetic], as 9123456 - 2 * 780
= 9123456 - 1560 = 9121896 [Unrelated justification], which is divisible by 7.
Divisibility by 8: The last three digits are 780, ~~which is divisible by 8~~[Bad arithmetic].
Divisibility by 9: ~~The sum of the digits is 45 + 9 = 54~~[Bad arithmetic:  sum of the digits is 45], which is
divisible by 9.
Divisibility by 11: The number 9123456780 is not divisible by 11.
~~Therefore, the number 9123456780 can be divided by every number from 2 to 18 without any
remainder~~[Unsupported conclusion] except for 11.
~~So, the smallest arrangement of the digits that forms a number divisible by every number from
2 to 18 (except for 11) is 9123456780.~~[Wrong conclusion]

---

**Solution by human**

We model the number $N$ with ten digits $c_0$, $c_1$, $c_2$, $c_3$, $c_4$, $c_5$, $c_6$, $c_7$, $c_8$, $c_9$ as a sum:

$$N = c_9 + c_8 * 10 + c_7 * 10^2 + c_6 * 10^3 + c_5 * 10^4 + c_4 * 10^5 + c_3 * 10^6 + c_2 * 10^7 + c_1 * 10^8 + c_0 * 10^9 \quad (2.1)$$

Mace4 computes four solutions:

| Model | $c_0$ | $c_1$ | $c_2$ | $c_3$ | $c_4$ | $c_5$ | $c_6$ | $c_7$ | $c_8$ | $c_9$ |
|-------|-------|-------|-------|-------|-------|-------|-------|-------|-------|-------|
| 1     | 2     | 4     | 3     | 8     | 1     | 9     | 5     | 7     | 6     | 0     |
| 2     | 3     | 7     | 8     | 5     | 9     | 4     | 2     | 1     | 6     | 0     |
| 3     | 4     | 7     | 5     | 3     | 8     | 6     | 9     | 1     | 2     | 0     |
| 4     | 4     | 8     | 7     | 6     | 3     | 9     | 1     | 5     | 2     | 0     |

**Puzzle 19. The archery match**

On a target on which the scoring was 40 for the bull's-eye, and 39, 24, 23, 17, and 16 respectively for the rings from the center outwards, three players had a match with six arrows each. The result was: Miss Dora Talbot, 120 points; Reggie Watson, 110 points; Mrs. Finch, 100 points. Every arrow scored, and the bull's eye was only once hit. Can you, from these facts, determine the exact six hits made by each competitor? (puzzle 467 from (Dudeney, 2016))

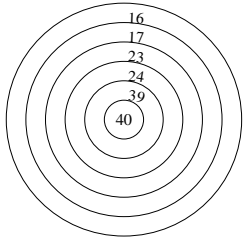

**Solution by ChatGPT**                                **Wrong**

To determine the exact six hits made by each competitor, we need to find a combination of scores that adds up to the total points for each player and satisfies the given conditions.[Task understanding: seems correct]

Let's represent the hits made by Miss Dora Talbot, Reggie Watson, and Mrs. Finch as follows: $D$ = Score of Miss Dora Talbot, $R$ = Score of Reggie Watson, $F$ = Score of Mrs. Finch. We know the following information:

The scoring on the target: 40 for the bull's-eye, and 39, 24, 23, 17, and 16 respectively for the rings from the center outwards.

The total points for each player: $D$ = 120 points, $R$ = 110 points, $F$ = 100 points.

Every arrow scored, and the bull's-eye was hit only once.

Based on this information, we can write the equations for each player's score:

$D = 40 + a + b + c + d + e$ [Too strong assumption: we do not know that Dora hit the bull's eye], where $a$, $b$, $c$, $d$, $e$ are the scores for the five rings from the center outwards.

$R = 40 + f + g + h + i + j$ [Too strong assumption: we do not know that Reggie hit the bull's eye],
[Inconsistency: GPT already learned that bull's eye was hit only once.]

~~$F = 40 + k + l + m + n + o$~~ [Too strong assumption: we do not know that Dora hit the bull's eye],
[Inconsistency: GPT already learned that bull's eye was hit only once.]

Also, ~~we know that $a$, $b$, $c$, $d$, $e$, $f$, $g$, $h$, $i$, $j$, $k$, $l$, $m$, $n$, $o$ are distinct~~ [Too strong assumption: a player could hit the same circle several times] positive integers ~~between 16 and 39~~ [Too weak assumption: the values are exactly 39, 24, 23, 17, and 16]
[Inconsistency: GPT have already learned that the values are 39, 24, 23, 17, 16 or 40] (inclusive).

Now, we can try different combinations of scores for $a$, $b$, $c$, $d$, $e$, $f$, $g$, $h$, $i$, $j$, $k$, $l$, $m$, $n$, $o$ to find the one that satisfies the given conditions.

**Solution by ChatGPT**                                                      **Wrong**

By ^(Solving strategy:   trial and error), we can find the following solution:

~~D = 120 : 40 + 23 + 24 + 17 + 16 + 0~~ ^(Inconsistency:   GPT (wrongly) assumed that the values are between 16 and 40.   Value 0 should not appe)

~~R = 110 : 40 + 23 + 24 + 17 + 0 + 6~~ ^(Inconsistency:   GPT (wrongly) assumed that the values are between 16 and 40.   Values 0 and 6 should not)

~~F = 100 : 40 + 24 + 17 + 0 + 6 + 13~~ ^(Inconsistency:   GPT (wrongly) assumed that the values are between 16 and 40.   Values 0, 6 and 13 should)

So, ~~the exact six hits made by each competitor~~^(Lack of commonsense:   0 is not a hit) are as follows:

~~Miss   Dora   Talbot:   23,   24,   17,   16,   Bull's-eye,~~
~~Miss~~^(Inconsistency:   Miss appears even if GPT learned that every arrow scored)

~~Reggie   Watson:   23,   24,   17,   Bull's-eye,   Miss,~~
~~6~~^(Inconsistency:   Miss appears even if GPT learned that every arrow scored)

~~Mrs. Finch: 24, 17, Bull's-eye, Miss, 6, 13~~^(Inconsistency:   Miss appears even if GPT learned that every arrow scored)
^(Wrong conclusion)

**Solution by human**

Let $p_{xy}$ represent the number of hits on target $y$ by player $x$. Mace4 finds a single solution:

| $p_{10}$ | $p_{11}$ | $p_{12}$ | $p_{13}$ | $p_{14}$ | $p_{15}$ | $p_{20}$ | $p_{21}$ | $p_{22}$ | $p_{23}$ | $p_{24}$ | $p_{25}$ | $p_{30}$ | $p_{31}$ | $p_{32}$ | $p_{33}$ | $p_{34}$ | $p_{35}$ |
|---|---|---|---|---|---|---|---|---|---|---|---|---|---|---|---|---|---|
| 1 | 0 | 0 | 0 | 0 | 5 | 0 | 0 | 0 | 2 | 0 | 4 | 0 | 0 | 0 | 0 | 4 | 2 |

**Puzzle 20. Target practice**

Colonel Crackham paid a visit one afternoon by invitation to the Slocomb-on-Sea Toxophilite Club, where he picked up the following little poser. Three men in a competition had each six shots at a target. They all hit the target every time. The bull's-eye scores 50, the next ring 25, the next 20, the next 10, the next 5, the next 3, the next 2, and the outside ring scores only 1. It will be seen that the hits on the target are one bull's-eye, two 25's, three 20's, three 10's, three 1 's, and two hits in every other ring. Now the three men tied with an equal score. Next morning the Colonel asked his family to show the exact scoring of each man. Will it take the reader many minutes to find the correct answer? (puzzle 468 from (Dudeney, 2016))

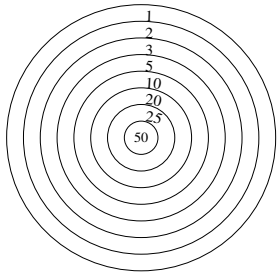

**Solution by ChatGPT**                                              **wrong**

Let's assume the three men are *A*, *B*, and *C*, and they each had six shots at the target. We are given the following information: Scoring on the target: Bull's-eye (50), 25 ring (25), 20 ring (20), 10 ring (10), 5 ring (5), 3 ring (3), 2 ring (2), and outside ring (1). Hits on the target for the three men: one bull's-eye, two 25's, three 20's, three 10's, three 1's, and two hits in every other ring.

To find the correct answer, we need to determine how many points each man scored with each type of shot (bull's-eye, 25's, 20's, etc.) and then calculate the total score for each man. <sup>Task understanding: Seems right</sup>

Let's represent the scores for the three men as follows: *A* = Score of man *A*, *B* = Score of man *B*, *C* = Score of man *C*. We know the following:

One bull's-eye: ~~Each man scored one bull's-eye, so we add 50 points to each of their scores.~~<sup>Inconsistency: there is only one bull's-eye</sup> Two 25's: ~~Each man scored two 25's, so we add 2 * 25 = 50 points to each of their scores.~~<sup>Inconsistency: there are two 25's in total</sup> Three 20's: ~~Each man scored three 20's, so we add 3 * 20 = 60 points to each of their scores.~~<sup>Inconsistency: there are three 20's in total</sup> Three 10's: ~~Each man scored three 10's, so we add 3 * 10 = 30 points to each of their scores.~~<sup>Inconsistency: there are three 10's in total</sup> Three 1's: ~~Each man scored three 1's, so we add 3 * 1 = 3 points to each of their scores.~~<sup>Inconsistency: there are three 10's in total</sup>

**Solution by ChatGPT**                                                                    **Wrong**

Now, let's consider the ~~two hits in every other ring (5, 3, and 2 rings)~~ <sup>Wrong assumption: none says there are two hits in these rings</sup> ~~We have a total of 6 shots in these rings for each man.~~<sup>Wrong assumption: there 6 shots in all rings, not only in these three rings</sup> Since these rings have values of 5, 3, and 2, we need to distribute these 6 shots in such a way that the total score is equal for all three men. One possible distribution is:

Man A: Two 5's, two 3's, and two 2's (2 * 5 + 2 * 3 + 2 * 2 = 14 points) Man B: Two 5's, two 3's, and two 2's (2 * 5 + 2 * 3 + 2 * 2 = 14 points) Man C: Two 5's, two 3's, and two 2's (2 * 5 + 2 * 3 + 2 * 2 = 14 points)

Now we can calculate the total scores for each man:

~~A = 50 (bull's-eye) + 50 (25's) + 60 (20's) + 30 (10's) + 3 (1's) + 14 (5's, 3's, and 2's) = 207 points~~ <sup>Wrong claim based on previous errors</sup> ~~B = 50 (bull's-eye) + 50 (25's) + 60 (20's) + 30 (10's) + 3 (1's) + 14 (5's, 3's, and 2's) = 207 points C = 50 (bull's-eye) + 50 (25's) + 60 (20's) + 30 (10's) + 3 (1's) + 14 (5's, 3's, and 2's) = 207 points~~

~~All three men scored 207 points~~, so they tied with an equal score.

Therefore, the exact scoring for each man is as follows:

~~Man A: one bull's-eye, two 25's, three 20's, three 10's, three 1's, and two hits in every other ring for a total of 207 points. Man B: one bull's-eye, two 25's, three 20's, three 10's, three 1's, and two hits in every other ring for a total of 207 points. Man C: one bull's-eye, two 25's, three 20's, three 10's, three 1's, and two hits in every other ring for a total of 207 points.~~<sup>Wrong conclusion</sup>

**Solution by human**

One solution computed by Mace4 is:

| Player | 50 | 25 | 20 | 10 | 5 | 3 | 2 | 1 |
|--------|----|----|----|----|---|---|---|---|
| 0      | 1  | 0  | 0  | 1  | 1 | 1 | 1 | 1 |
| 1      | 0  | 1  | 2  | 0  | 0 | 1 | 1 | 1 |
| 2      | 0  | 1  | 1  | 2  | 1 | 0 | 0 | 1 |

This solution represents the first 21 values from the function $p(x, y)$ Since we can permute each player, there are 3!=6 isomorphic solutions.

**Puzzle 21. The nine barrels**

In how many different ways may these nine barrels be arranged in three tiers of three so that no barrel shall have a smaller number than its own below it or to the right of it? The first correct arrangement that will occur to you is 1 2 3 at the top, then 4 5 6 in the second row, and 7 8 9 at the bottom. The second correct arrangement is 1 3 6 at the top, then 2 5 7 in the second row, and 4 8 9 at the bottom. How many are there altogether? (puzzle 138 from (Dudeney, 2016))

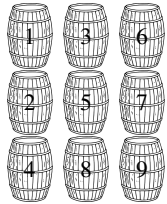     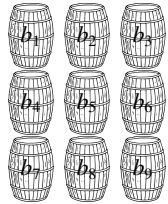

**Solution by ChatGPT**                                               **Wrong**

To find all the different ways to arrange nine barrels in three tiers of three such that no barrel has a smaller number than its own below it or to the right of it, we can use a recursive approach.<sup>Solving strategy: Not wrong</sup> Let's represent the nine barrels as follows:

1 2 3
4 5 6
7 8 9

We will start by placing the smallest number (1) in the top-left corner. Now, we need to place the remaining eight barrels in the remaining positions while satisfying the given condition. To do this, we can recursively place the next number in the top row, then the next number in the middle row, and finally the last number in the bottom row.

Since there are two valid arrangements given in the puzzle (1 2 3 at the top and 1 3 6 at the top), ~~we can start with each of these arrangements and recursively find all the possible arrangements.~~<sup>Implication does not hold</sup>

Starting with 1 2 3 at the top: We place 2 in the top-middle position (2 options: 1 2 3 and ~~1 3 2~~)<sup>Bad topology: 2 is not in the top middle position</sup>. Then, we place 4 in the middle-left position (2 options: 4 5 6 and 4 6 5). Next, we place 7 in the bottom-left position (1 option: 7 8 9). ~~The remaining numbers are 5, 6, 8, and 9, which are already placed following the given condition.~~<sup>Inconsistency: 4 6 5 does not comply with the given condition</sup>

So, we have two valid arrangements:

1 2 3               1 2 3
4 5 6               4 ~~6 5~~
7 8 9               7 8 9

**Solution by ChatGPT**                                                      **Wrong**

Starting with 1 3 6 at the top: We place 2 in the top-middle position (~~1  option:  1  3  6~~)Inconsistency:  2 is not in the top middle position. Then, we place 4 in the middle-left position (2 options:  4 5 6 and 4 6 5). Next, we place 7 in the bottom-left position (1 option: 7 8 9). ~~The  remaining  numbers  are  5,  8,  and  9,  which  are  already  placed  following  the  given condition.~~Inconsistency:  6 appears twice, 2 is missing

So, we have one valid arrangement:  1 3 6 4 5 ~~6~~Inconsistency:  6 appears twice, 2 is missing 7 8 9

In total, ~~there  are  three  different  ways  to  arrange  the  nine  barrels  in  three  tiers  of  three, satisfying  the  given  condition~~ Wrong conclusion:  there are 42 ways

1 2 3 4 5 6 7 8 9
1 2 3 4 6 ~~5~~ 7 8 9
1 3 6 4 5 ~~6~~ 7 8 9
~~Therefore,    there    are    three    different    arrangements    in    total.~~
Hallucination:  Even 2 correct examples were given, GPT prints 3 final solutions in which two of them are wrong

Solving strategy:  Since two examples were given, this can be considered a two-shot learning for GPT

---

**Solution by human**

Let the nine barrels noted with $b_i$, $i \in [1..9]$ as in the illustration above. Mace4 finds 42 models. Six of them are:

| Model | $b_1$ | $b_2$ | $b_3$ | $b_4$ | $b_5$ | $b_6$ | $b_7$ | $b_8$ | $b_9$ |
|-------|-------|-------|-------|-------|-------|-------|-------|-------|-------|
| 1     | 1     | 2     | 3     | 4     | 5     | 6     | 7     | 8     | 9     |
| 2     | 1     | 2     | 4     | 3     | 5     | 7     | 6     | 8     | 9     |
| 3     | 1     | 2     | 5     | 3     | 4     | 8     | 6     | 7     | 9     |
| 4     | 1     | 3     | 4     | 2     | 5     | 8     | 6     | 7     | 9     |
| 5     | 1     | 3     | 5     | 2     | 4     | 8     | 6     | 7     | 9     |
| 6     | 1     | 3     | 6     | 2     | 5     | 7     | 4     | 8     | 9     |

Figure 2.2 illustrates model 6 (left), model 3 (center), and model 2 (right).

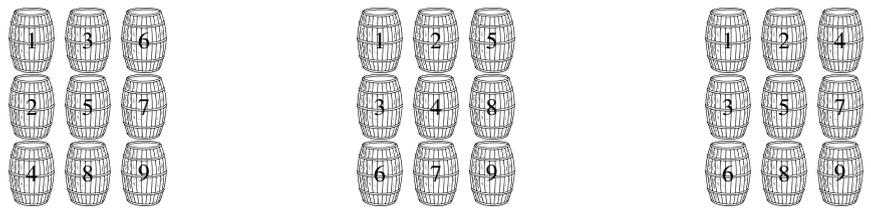

Figure 2.2: Three solutions among 42 found by Mace4

**Puzzle 22. Upside-down year**

An upside down year is a year which reads the same when the 'year number' is turned upside down. That is when the digits which form the year appear the same when rotated 180 degrees. How many upside down years have been from 0BC until now? When was the latest year that is the same upside down? Which is the next upside down year? (puzzle 39 from (Kordemsky, 1992)).

**Solution by ChatGPT**                                    **Partial Correct**

To find upside-down years from 0 BC until now, we need to consider the digits that can be rotated 180 degrees and still remain the same <span style="color:blue">Task understanding: It seems so, but in the follwoing GPT forgets that the rotates year should remain the same</span>. These digits are 0, 1, 6, 8, and 9<span style="color:orange">Solving strategy: Good strategy</span>. Now, we'll check the numbers in each century and count the upside-down years: From 0 to 99:

0 can be rotated to 0, so it's an upside-down year.

1 can be rotated to 1, so it's an upside-down year.

~~6 can be rotated to 9, so it's an upside-down year.~~<span style="color:red">Wrong assumption: the upside down year should be equal with the initial year</span>

8 can be rotated to 8, so it's an upside-down year.

~~9 can be rotated to 6, so it's an upside-down year.~~<span style="color:red">Wrong assumption: the upside down year should be equal with the initial year</span>

~~Total upside-down years in the first century: 5~~<span style="color:red">Incomplete solution: what about years of 2 digits, e.g. 11</span>

From 100 to 199: ~~1 can be rotated to 1, so it's an upside-down year.~~<span style="color:red">Inconsistency: 1 is not between 100 and 199</span>

~~Total upside-down years in the second century: 1~~<span style="color:red">Implication does not hold</span>

From 200 to 299: None of the numbers can be rotated to form another valid number, so there are no upside-down years in this century.

From 300 to 399: None of the numbers can be rotated to form another valid number, so there are no upside-down years in this century.

From 400 to 499: None of the numbers can be rotated to form another valid number, so there are no upside-down years in this century.

From 500 to 599: None of the numbers can be rotated to form another valid number, so there are no upside-down years in this century.

**Solution by ChatGPT**                                                    **Wrong**

From 600 to 699: ~~None of the numbers can be rotated to form another valid number, so there are no upside-down years in this century.~~ Wrong assumption: 609, 619 and 689 are three up-side years in this interval

From 700 to 799: None of the numbers can be rotated to form another valid number, so there are no upside-down years in this century.

From 800 to 899: ~~None of the numbers can be rotated to form another valid number, so there are no upside-down years in this century.~~ Wrong assumption: 808, 818 and 888 are three up-side years in this interval

From 900 to 999: ~~9 can be rotated to 6, so it's an upside-down year.~~ Inconsistency: the year does not belong to the current interval

~~Total upside-down years in the tenth century: 1~~ Wrong conclusion: 906, 916 and 986 are three upside years in this interval

~~Now, we can count the total number of upside-down years from 0 BC until now.~~ Incomplete reasoning: years after 1000 are missing

Total upside-down years = Total upside-down years in each century = ~~5 (1st century) + 1 (2nd century) + 1 (10th century) = 7~~ Wrong conclusion

The latest upside-down year that is the same upside down is 1961 ~~(reading it upside-down: 1691)~~. Wrong assumption: the upside down year is equal to the initial year

The next upside-down year after 1961 is 6009 ~~(reading it upside-down: 9006)~~
Wrong assumption: the upside down year is equal to the initial year

So, ~~there have been 7 upside-down years from 0 BC until now~~
Wrong conclusion: there are 24 upside down years, and the latest one was 1961, with the next one being 6009.

**Solution by human**

There are 24 solutions in interval [0..2020], among which 3 solutions for $n = 1$, 4 solutions for $n = 2$, 12 solutions for $n = 3$, and 5 solutions for $n = 4$:

| Models | $a$ | $b$ | $c$ | $d$ | $n$ |
|---|---|---|---|---|---|
| 1 | 0 | 0 | 0 | 0 | 1 |
| 2 | 0 | 0 | 0 | 1 | 1 |
| 3 | 0 | 0 | 0 | 8 | 1 |
| 4 | 0 | 0 | 1 | 1 | 2 |
| 5 | 0 | 0 | 6 | 9 | 2 |
| 6 | 0 | 0 | 8 | 8 | 2 |
| 7 | 0 | 0 | 9 | 6 | 2 |
| 8 | 0 | 1 | 0 | 1 | 3 |
| 9 | 0 | 1 | 1 | 1 | 3 |
| 10 | 0 | 1 | 8 | 1 | 3 |
| 11 | 0 | 6 | 0 | 9 | 3 |
| 12 | 0 | 6 | 1 | 9 | 3 |

| Models | $a$ | $b$ | $c$ | $d$ | $n$ |
|---|---|---|---|---|---|
| 13 | 0 | 6 | 8 | 9 | 3 |
| 14 | 0 | 8 | 0 | 8 | 3 |
| 15 | 0 | 8 | 1 | 8 | 3 |
| 16 | 0 | 8 | 8 | 8 | 3 |
| 17 | 0 | 9 | 0 | 6 | 3 |
| 18 | 0 | 9 | 1 | 6 | 3 |
| 19 | 0 | 9 | 8 | 6 | 3 |
| 20 | 1 | 0 | 0 | 1 | 4 |
| 21 | 1 | 1 | 1 | 1 | 4 |
| 22 | 1 | 6 | 9 | 1 | 4 |
| 23 | 1 | 8 | 8 | 1 | 4 |
| 24 | 1 | 9 | 6 | 1 | 4 |

The latest upside-down year was 1961 (model 24). Mace4 quickly finds that the next upside-down year is 6009.

## Puzzle 23. Rotate digits

In a $3 \times 3$ grid, the digits from 1 to 9 are arranged so that the first row (containing digits 5, 8 and 3) added to the second row (containing digits 1, 4 and 6) equals the bottom row (containing digits 7, 2, and 9) (i.e. $583 + 146 = 729$). Now if the grid is rotated clockwise at 90 degrees, you will see that the first two rows still add up to the last row (i.e. $715 + 248 = 963$). Here the rotated first row contains the digits 7, 1, and 5, while the rotated second row contains the digits 2, 4, and 8. The rotated third row contains their sum: the digits 9, 6 and 3. Can you find another combination of the digits 1-9 which has the same property? (adapted from puzzle 60 from (Clessa, 1996))

| 5 | 8 | 3 |
|---|---|---|
| 1 | 4 | 6 |
| 7 | 2 | 9 |

| 7 | 1 | 5 |
|---|---|---|
| 2 | 4 | 8 |
| 9 | 6 | 3 |

## Solution by ChatGPT                                    Wrong

Let's consider the $3 \times 3$ grid as follows:

5 8 3
1 4 6
7 2 9

When the grid is rotated clockwise at 90 degrees, it becomes:

7 1 5
2 4 8
9 6 3

Now, we need to find another combination of the digits 1 to 9 that has the same property: the sum of the first two rows equals the third row both in the original and rotated grids. Task understanding: It seems so

To do this, we need to try different combinations of the digits Solving strategy: Trial and error:brute force to find a suitable arrangement. We can start with the top row and try different digits for each position. Since the bottom row will be the sum of the first two rows, there will be fewer possibilities to try. Let's try the following arrangement:

8 5 6
1 9 4
7 2 3

**Solution by ChatGPT**                                                    **Wrong**

Now, ~~when we add the first row (8 + 5 + 6 = 19) to the second row (1 + 9 + 4 = 14), we get the third row (7 + 2 + 3 = 12).~~[Bad arithmetic] Now, let's rotate the grid clockwise:

7 1 8
2 9 5
3 4 6

~~When we add the first row (7 + 1 + 8 = 16) to the second row (2 + 9 + 5 = 16), we get the third row (3 + 4 + 6 = 13).~~[Bad arithmetic] So, in this arrangement, the sum of the first two rows (19 + 14) equals the sum of the third row (12) in the original grid, and when rotated, the sum of the first two rows (16 + 16) equals the sum of the third row (13) in the rotated grid. Therefore, ~~the arrangement that satisfies the property is:~~[Lack of understanding: of the expression adding the first row]

8 5 6 1 9 4 7 2 3
and its rotated version:
7 1 8 2 9 5 3 4 6
~~Both arrangements have the same property described in the puzzle.~~[Wrong conclusion]

**Solution by human**

Mace4 finds two models: model 1 was given as example, and model 2 appears in Fig. 2.3.

| Model | $a_1$ | $a_2$ | $a_3$ | $a_4$ | $a_5$ | $a_6$ | $a_7$ | $a_8$ | $a_9$ |
|-------|-------|-------|-------|-------|-------|-------|-------|-------|-------|
| 2     | 5     | 8     | 3     | 1     | 4     | 6     | 7     | 2     | 9     |
| 1     | 4     | 8     | 2     | 1     | 5     | 7     | 6     | 3     | 9     |

| 4 | 8 | 2 |
|---|---|---|
| 1 | 5 | 7 |
| 6 | 3 | 9 |

| 6 | 1 | 4 |
|---|---|---|
| 3 | 5 | 8 |
| 9 | 7 | 2 |

Figure 2.3: The second solution for the rotated digits

**Puzzle 24. An unusual number**

Find a six-digit number which, when multiplied by an integer between 2 and 9 inclusive, gives the original six-digit number with its digits reversed. Thus, if the original number was 123,456, and the chosen integer is 8, then $123,456 \times 8$ should equal 654,321, which, of course, it doesn't. However, it is possible to find more than one solution to this problem, but I'll accept anyone that meets the required condition (puzzle 34 from (Clessa, 1996))

**Solution by ChatGPT**                                      **Admitting failure**

To find a six-digit number that, when multiplied by an integer between 2 and 9 inclusive, gives the original number with its digits reversed, we can use trial and error.<sup>Solving strategy: Trial and error: for GPT this is always incomplete.</sup>

<sup>Solving strategy: Wrong since it has to try all numbers of 6 digits.</sup> to check different numbers. We need to find a number that, when multiplied by any integer between 2 and 9, gives a result that has the same digits as the original number but in reverse order.<sup>Task understanding: It seems so</sup>

Let's start with a systematic approach: Start with a six-digit number in the form of ABCDEF. Check all possible multipliers from 2 to 9 (inclusive) and calculate the product. Check if the product has the same digits as the original number ABCDEF but in reverse order.<sup>Solving strategy: Brute force: it seems like an algorithm</sup>

Let's illustrate this process with an example: Original number: ABCDEF Let's assume ABCDEF = 123456 as an example. Now, we'll check all possible multipliers from 2 to 9.

When multiplied by 2: 123456 * 2 = 246912 (not a solution).
When multiplied by 3: 123456 * 3 = 370368 (not a solution).
When multiplied by 4: 123456 * 4 = 493824 (not a solution).
When multiplied by 5: 123456 * 5 = 617280 (not a solution).
When multiplied by 6: 123456 * 6 = 740736 (not a solution).
When multiplied by 7: 123456 * 7 = 864192 (not a solution).
When multiplied by 8: 123456 * 8 = 987648 (not a solution).
When multiplied by 9: 123456 * 9 = 1111104 (not a solution).

None of the products have the same digits as the original number in reverse order for the example ABCDEF = 123456.

**Solution by ChatGPT**                                                      **Wrong**

Let's try another number: Original number: ABCDEF Let's assume ABCDEF = 142857
as another example. Now, we'll check all possible multipliers from 2 to 9.
When multiplied by 2: 142857 * 2 = 285714 (not a solution).
When multiplied by 3: 142857 * 3 = 428571 (not a solution).
When multiplied by 4: 142857 * 4 = 571428 (not a solution).
When multiplied by 5: 142857 * 5 = 714285 (not a solution).
When multiplied by 6: 142857 * 6 = 857142 (not a solution).
When multiplied by 7: 142857 * 7 = 999999 (not a solution).
When multiplied by 8: 142857 * 8 = 1142856 (not a solution).
When multiplied by 9: 142857 * 9 = 1285713 (not a solution).
None of the products have the same digits as the original number in reverse order for the
example ABCDEF = 142857.
It is important to note that there are six-digit numbers that satisfy this
condition,$^{\text{Solving strategy: Admitting failure may be trigger by the input sentence: It is possible to find more than one solution"}}$
but finding them requires a more systematic approach or potentially using a
computer program to check all possible combinations. In this particular ex-
ample, we didn't find a solution for ABCDEF = 123456 or ABCDEF =
142857.$^{\text{Solving strategy: Interesting formulation: we didn't find a solution.}}$

**Solution by human**

Let the number composed of six digits $\overline{abcdef}$. Mace4 finds 2 models, one for $n = 4$ and
the other one for $n = 9$:

| Model | $a$ | $b$ | $c$ | $d$ | $e$ | $f$ | $n$ |
|-------|-----|-----|-----|-----|-----|-----|-----|
| 1 | 2 | 1 | 9 | 9 | 7 | 8 | 4 |
| 2 | 1 | 0 | 9 | 9 | 8 | 9 | 9 |

The two solutions are: 219,978 * 4 = 879,912 and 109,989 * 9 = 989,901.

# References

Clessa, J. (1996). *Math and Logic Puzzles for PC Enthusiasts*. Courier Corporation.

Dudeney, H. E. (2016). *536 Puzzles and curious problems*. Courier Dover Publications.

Kordemsky, B. A. (1992). *The Moscow puzzles: 359 mathematical recreations*. Courier Corporation.

Pierce, R. (2020). Math is fun.



# Chapter 3

# Practical puzzles

---

**Puzzle 25. Letter to parents**

You are sending the following crypto-arithmetic message asking money from your parents: SEND + MOST = MONEY. Each letter is unique. How much money you will receive in the worst case and in the best case?

---

**Solution by ChatGPT**        **Wrong**

To solve the crypto-arithmetic puzzle "SEND + MOST = MONEY," we need to find unique digits for each letter (S, E, N, D, M, O, T, Y) so that the addition holds true. Each letter represents a different digit from 0 to 9.

Let's try to find the possible solutions step by step:<span style="color:green">Solving strategy: New strategy: step by step</span> Observe that since "SEND + MOST = MONEY," the length of the sum (MONEY) must be equal to or greater than the lengths of the addends (SEND and MOST). ~~As MONEY has five digits, the two addends (SEND and MOST) must have at most four digits.~~<span style="color:red">Bad arithmetic</span>

Considering the carryovers, ~~it's clear that *S* and *M* must be 1 since there are no numbers that add up to 10 or more from only two digits~~<span style="color:red">Inconsistency: GPT already learned that the letters are distinct</span>.<span style="color:red">Bad arithmetic</span>

Now, we can set *S* = 1 and *M* = 1. The equation becomes: 1END + 1OST = 1ONEY

To ensure the largest possible sum, we should maximize the value of *N*, which means *N* = 9. ~~Now the equation is: 1END + 1OST = 19OEY.~~<span style="color:red">Bad arithmetic: letter *O* is replaced with 9 instead of *N*</span>

For the next digit, we need to find values for *E*, *O*, and *Y*. We cannot set *Y* = 1 since both *S* and *M* are already 1. Let's set *Y* = 0. The equation becomes: 1END + 1OST = 190E0

Now, we need to find values for *E* and *O*. Since *E* + *O* ends in 0, the only possible values for *E* and *O* are 5 and 5 (5 + 5 = 10)<span style="color:red">Bad arithmetic: carry might occur</span>. However, we cannot have duplicate digits, so this is not a valid solution.

~~So, we have found the best-case scenario: SEND = 1529, MOST = 6735, MONEY = 8264~~<span style="color:red">Implication does not hold Inconsistency: GPT already learned that MONEY has five digits</span>.

---

---

**Solution by ChatGPT**                                                    **Wrong**

To find the worst-case scenario, ~~we need to maximize the sum~~[Lack of commonsense: we need to minimise the sum] while keeping all letters unique. Let's set $S = 9$ and $M = 8$ to maximize the sum. The equation becomes: $9END + 8OST = 89OEY$
To maximize the sum, we need to set $E$, $O$, and $Y$ as large as possible while keeping them unique. Let's set $E = 7$, $O = 6$, and $Y = 5$. The equation becomes: $9N7D + 86ST = 89675$
Now, we need to find values for $N$, $D$, and $T$. To maximize the sum, we set $N = 4$, $D = 3$, and $T = 2$. The equation becomes: ~~$9473 + 8622 = 89675$~~[Bad arithmetic: the sum does not hold]
~~So, the worst-case scenario is: SEND = 9473, MOST = 8622, MONEY = 89675~~[Wrong conclusion]
~~In the worst-case scenario, you will receive \$89,675, and in the best-case scenario, you will receive \$8,264.~~[Lack of commonsense]

---

**Solution by human**

There are 16 models satisfying the sum:

| Models | D | E | M | N | O | S | T | Y |
|--------|---|---|---|---|---|---|---|---|
| 1  | 2 | 3 | 1 | 4 | 0 | 9 | 5 | 7 |
| 2  | 2 | 3 | 1 | 4 | 0 | 9 | 6 | 8 |
| 3  | 2 | 4 | 1 | 5 | 0 | 9 | 6 | 8 |
| 4  | 2 | 6 | 1 | 7 | 0 | 9 | 3 | 5 |
| 5  | 2 | 7 | 1 | 8 | 0 | 9 | 3 | 5 |
| 6  | 2 | 7 | 1 | 8 | 0 | 9 | 4 | 6 |
| 7  | 3 | 5 | 1 | 6 | 0 | 9 | 4 | 7 |
| 8  | 3 | 6 | 1 | 7 | 0 | 9 | 2 | 5 |
| 9  | 3 | 6 | 1 | 7 | 0 | 9 | 5 | 8 |
| 10 | 3 | 7 | 1 | 8 | 0 | 9 | 2 | 5 |
| 11 | 4 | 5 | 1 | 6 | 0 | 9 | 3 | 7 |
| 12 | 4 | 7 | 1 | 8 | 0 | 9 | 2 | 6 |
| 13 | 5 | 3 | 1 | 4 | 0 | 9 | 2 | 7 |
| 14 | 5 | 6 | 1 | 7 | 0 | 9 | 3 | 8 |
| 15 | 6 | 3 | 1 | 4 | 0 | 9 | 2 | 8 |
| 16 | 6 | 4 | 1 | 5 | 0 | 9 | 2 | 8 |

Three such models (1, 2 and 3 from the table above) are:

```
  9 3 4 2  +          9 4 5 6  +          9 5 6 4  +
  1 0 9 5              1 0 9 2              1 0 9 3
-----------         -----------         -----------
1 0 4 3 7           1 0 5 4 8           1 0 6 5 7
```

The minimum value for MONEY is: 10,437 (model 1).
The maximum value for MONEY is: 10,876 (model 6).

## Puzzle 26. Ambiguous dates

In this country a date such as July 4, 1971, is often written 7/4/71, but in other countries the month is given second and the same date is written 4/7/71. If you do not know which system is being used, how many dates in a year are ambiguous in this two-slash notation? Gardner (2005)

## Solution by ChatGPT — **Wrong**

To find the number of dates in a year that are ambiguous in the two-slash notation, we need to consider both date formats (month/day/year and day/month/year) and determine the overlapping dates. In the month/day/year format (e.g., 7/4/71), the range of possible dates for the month is from 1 to 12, and the range for the day is from 1 to 31. In this format, the year can be represented by the last two digits (e.g., 71 for 1971). In the day/month/year format (e.g., 4/7/71), the range of possible dates for the day is from 1 to 31, and the range for the month is also from 1 to 12. In this format, the year can be represented by the last two digits (e.g., 71 for 1971). *Task understanding: It seems so*

Now, let's determine the overlapping dates: *Solving strategy: No apparent strategy* For dates in which the month and day have the same numeric representation, they can be ambiguous in both formats (e.g., 1/1/71 or 11/11/71). For dates in which the day is greater than 12, there is no ambiguity because it can only be interpreted as the day/month/year format (e.g., 15/4/71). For dates in which the month is greater than 12, there is no ambiguity because it can only be interpreted as the month/day/year format (e.g., 4/15/71). For dates in which the month and day are less than or equal to 12, there can be ambiguity. For example, 2/3/71 could be interpreted as either February 3, 1971 (month/day/year) or March 2, 1971 (day/month/year). *Task understanding: The problem is well analysed*

To find the total number of ambiguous dates in a year, we need to count the overlapping dates for months and days less than or equal to 12:

*Solving strategy: This is the solving strategy and it is the right one*

---

**Solution by ChatGPT**                                                                    **Wrong**

There are 12 months (1 to 12) and 12 days (1 to 12) that are the same numeric representation. So, there are 12 ambiguous dates for these cases.

For months and days between 1 and 12 (excluding the dates with the same numeric representation), we have 12 months and 12 days, resulting in 12 * 12 = 144 ambiguous dates.

~~The total number of ambiguous dates in a year is 12 (for the dates with the same numeric representation) + 144 (for the other overlapping dates) = 156.~~
Bad interpretation in the last step:   correct is 144-12=132

~~Therefore, there are 156 ambiguous dates in a year in the two-slash notation.~~
Wrong conclusion:  but supported by faulty chain of reasoning

---

**Solution by human**

Mace4 computes 132 models. Each month has 11 ambiguous dates, making 132 in all.
For December (*month* = 12 in line 9), Mace4 outputs the following 11 ambiguous dates:

| Model | 1 | 2 | 3 | 4 | 5 | 6 | 7 | 8 | 9 | 10 | 11 |
|-------|----|----|----|----|----|----|----|----|----|----|----|
| *day* | 1 | 2 | 3 | 4 | 5 | 6 | 7 | 8 | 9 | 10 | 11 |
| *month* | 12 | 12 | 12 | 12 | 12 | 12 | 12 | 12 | 12 | 12 | 12 |

**Puzzle 27. One landlord and 100 servants**

A certain head of a household had 100 servants. He ordered that they be given 100 modia of corn as follows. The men should receive three modia; the women, two; and the children, half a modium. Thus how many men, women, and children were there? Can you find all seven solutions? (puzzle composed by the medieval scholar Alcuin of York in *Propositiones ad acuendos juvenes* (puzzle 34 there, and translated by Hadley and Singmaster (1992)) Danesi (2018)).

**Solution by ChatGPT**          **No solution, only valid but inefficient algorithm**

To solve this puzzle, we need to find the number of men, women, and children among the 100 servants based on the given distribution of corn: Men: 3 modia, Women: 2 modia, Children: 0.5 modium (0.5 modium = 1/2 modium) Let's assume there are $x$ men, $y$ women, and $z$ children. Now, we can write the equation for the distribution of corn:
$3x + 2y + 0.5z = 100$
We also know that the total number of servants is 100: $x + y + z = 100$
Now, we have a system of two equations with three variables.<sup>Task understanding: right understanding and formalisation</sup>

To find the solutions, we need to find all possible integer values of $x$, $y$, and $z$ that satisfy both equations. Let's find the solutions step by step:<sup>Task understanding: a favorite strategy for GPT</sup>

Start with $x = 0$ (assuming there are no men). Then the second equation becomes $y + z = 100$. We can find solutions for $y$ and $z$ that satisfy this equation, and then check if the first equation ($3x + 2y + 0.5z = 100$) is satisfied.

For example, when $y = 50$, $z = 50$, we have:
$3x + 2 * 50 + 0.5 * 50 = 100$
$3x + 100 + 25 = 100$
$3x = -25$
$x = -25/3$ (Not a valid solution as we need integer values for $x$, $y$, and $z$.)

Continue this process for other values of $y$ and $z$.

Next, start with $x = 1$ (assuming there is 1 man). Then the second equation becomes $1 + y + z = 100$. Again, find solutions for $y$ and $z$ that satisfy this equation and also check the first equation ($3x + 2y + 0.5z = 100$).

Continue this process for $x = 2$, $x = 3$, and so on, up to $x = 100$.

By systematically trying different values for $x$, $y$, and $z$, you can find all seven solutions that satisfy the given conditions.

**Solution by human**

There are 100 man ($m$), women ($f$) and children ($c$): $m + f + c = 100$. The 100 modia is split with: $3 * m + 2 * f + c/2 = 100$. Mace4 finds 6 models satisfying the constraints:

| Model | $c$ | $f$ | $m$ | $n$ |
|-------|-----|-----|-----|-----|
| 1 | 68 | 30 | 2 | 34 |
| 2 | 70 | 25 | 5 | 35 |
| 3 | 72 | 20 | 8 | 36 |
| 4 | 74 | 15 | 11 | 37 |
| 5 | 76 | 10 | 14 | 38 |
| 6 | 78 | 5 | 17 | 39 |

Without the assumption $m \neq 0$, $f \neq 0$, and $c \neq 0$, there would be a seventh model: 20 man, no women and 80 children.

**Puzzle 28. No consecutive numbers in adjacent nodes**

Place numbers 1 through 8 on eight nodes $a$, $b$, $c$, $d$, $e$, $f$, $g$, $h$ so that:

1. Each number appears exactly once;

2. No connected nodes have consecutive numbers

Node $a$ is connected with $b$, $g$ and $f$. Node $b$ is connected with $a$, $g$, $h$ and $c$. Node $c$ is connected with $b$, $g$, $h$ and $d$. Node $d$ is connected with $c$, $h$ and $f$. Node $e$ is connected with $d$, $g$, $h$ and $f$. Node $g$ is connected with $a$, $b$, $c$, $h$, $f$, $e$ and $d$. Node $h$ is connected with $d$, $b$, $c$, $g$, $f$, $e$ and $d$. How many solutions exist?

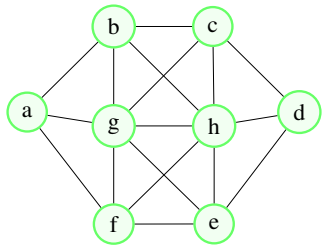

**Solution by ChatGPT**                                                    **Wrong**

To solve this puzzle, we need to place the numbers 1 through 8 on the eight nodes ($a$, $b$, $c$, $d$, $e$, $f$, $g$, $h$) such that each number appears exactly once, and no connected nodes have consecutive numbers. Let's consider the connectivity of the nodes: Node $a$ is connected with $b$, $g$ and $f$. Node $b$ is connected with $a$, $g$, $h$ and $c$. Node $c$ is connected with $b$, $g$, $h$ and $d$. Node $d$ is connected with $c$, $h$ and $f$. Node $e$ is connected with $d$, $g$, $h$ and $f$. Node $g$ is connected with $a$, $b$, $c$, $h$, $f$, $e$ and $d$. Node $h$ is connected with $d$, $b$, $c$, $g$, $f$, $e$ and $d$.

To find the number of solutions, we can use a backtracking algorithm to explore all possible combinations of numbers on the nodes while ensuring that the conditions are satisfied. <span style="color:green">Solving strategy: Backtracking</span>.

We start by placing the number 1 on node a, as it doesn't have any adjacent nodes that could violate the condition. Then, we proceed to place the other numbers on the remaining nodes while ensuring that no adjacent nodes have consecutive numbers.

Here is one possible solution: ~~a (1) – b (4) – c (7) – d (2) – e (6) – f (8) – g (3) – h (5)~~ <span style="color:red">Logical fault: $g = 3$ conflicts with $b = 4$, h=5 conflicts with $b = 4$</span>

To find all possible solutions, we can use a backtracking algorithm that explores all combinations of numbers on the nodes, making sure to check the conditions at each step and backtrack when a condition is violated. This will lead to finding all valid solutions.

Using the backtracking approach, it is possible to find all solutions. ~~The total number of solutions for this puzzle is 15.~~ <span style="color:red">Wrong conclusionUnjustified conclusion</span>

**Solution by human**

Mace4 finds 4 solutions:

| Model | $a$ | $b$ | $c$ | $d$ | $e$ | $f$ | $g$ | $h$ |
|-------|-----|-----|-----|-----|-----|-----|-----|-----|
| 1 | 2 | 5 | 3 | 7 | 4 | 6 | 8 | 1 |
| 2 | 7 | 4 | 6 | 2 | 5 | 3 | 1 | 8 |
| 3 | 2 | 6 | 4 | 7 | 3 | 5 | 8 | 1 |
| 4 | 7 | 3 | 5 | 2 | 6 | 4 | 1 | 8 |

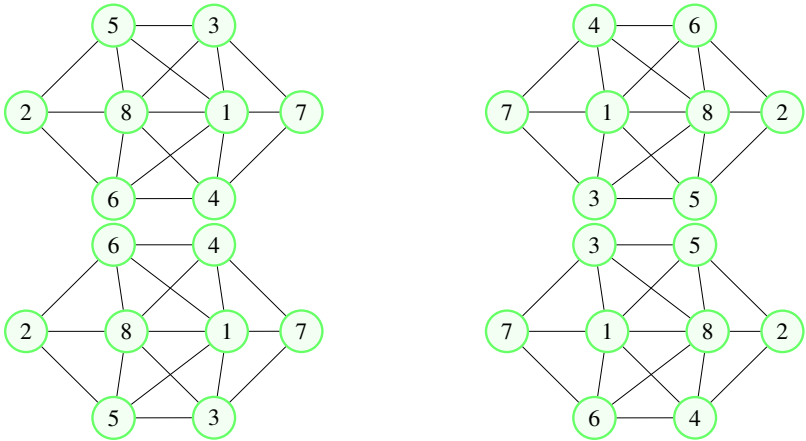

Figure 3.1: The four solutions found by Mace4

**Puzzle 29. Colouring Dracula's land in red**

The map of Romania contains the following regions: Transylvania, Maramures, Bucovina, Moldova, Muntenia, Oltenia, Banat, Crisana and Dobrogea. Transylvania is connected with all other regions excepting Dobrogea. Dobrogea is connected only with Moldova and Munteania. The connection relation is symmetric. Can the following map of Romania be coloured with only 3 distinct colours? What about 4 colours? Find a solution in which Transylvania, the birth place of Dracula, is coloured red. How many solutions exist?

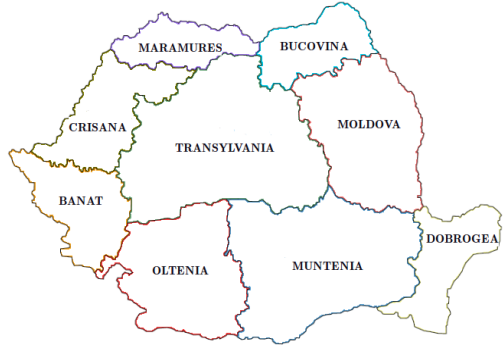

**Solution by ChatGPT**                    **Wrong and ambigous**

To determine if the map of Romania can be colored with only 3 distinct colors (3-colorable) or with 4 distinct colors (4-colorable), we need to apply the concept of the Four Color Theorem.

The Four Color Theorem states that any map on a plane can be colored with only 4 distinct colors in such a way that no two adjacent regions (sharing a common border) have the same color. This theorem has been proven mathematically and applies to any planar map.

Since the map of Romania is a planar map (regions are on a 2D surface without any overlapping edges), we can conclude that it is 4-colorable. This means that it can be colored with 4 distinct colors in such a way that no two adjacent regions share the same color.

**Solution by ChatGPT**                                                            **Wrong**

To find a solution where Transylvania is colored red, we can assign colors to the regions while ensuring that adjacent regions have different colors:

Transylvania (red) - Banat (green) - Crisana (blue) - Maramures (yellow)

- Oltenia - Muntenia (green) - Bucovina (blue) ~~Moldova (yellow) - Dobrogea (yellow)~~ Inconsistency: Moldova and Dobrogea have the same color even if they are connected

In this solution, Transylvania is colored red, and no two adjacent regions share the same color, fulfilling the 4-colorable condition.

Regarding the number of solutions, if we consider color swapping (e.g., switching the colors of two regions while keeping the overall map structure unchanged), there can be multiple valid solutions with the same coloring pattern. However, the overall structure and the number of colors used will remain the same. For a map of Romania with the regions specified and the connectivity given, there can be multiple colorings that follow the 4-colorable rule Fuzzy conclusion: multiple coloring, instead of specific answer.

**Solution by human**

Assume there are only 3 colors available. In this case Mace4 cannot find any model. By introducing the fourth color, Mace4 outputs 252 models (most of them isomorphic). One such solution is depicted below. The regions are assigned to the following values from the domain:

| Banat | Crisana | Bucovina | T | Maramures | Moldova | Muntenia | Oltenia | Dobrogea |
|-------|---------|----------|---|-----------|---------|----------|---------|----------|
| 7 | 8 | 2 | 0 | 1 | 3 | 5 | 6 | 4 |

Note that the Transylvania has value 0 which corresponds to the first position in the vector of colors. This first position contains the value 0, which corresponds to the red colour

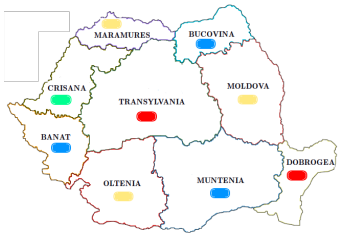

Figure 3.2: Colored map of Romania with Mace4

**Puzzle 30. Golomb ruler**

Define a ruler with $M = 4$ marks (e.g. $a$, $b$, $c$, $d$) so that the distances between any two marks are different. Your ruler should be able to measure all the integer distances up to length $L = 6$. There should be only one way of measuring an integer distance with your ruler.

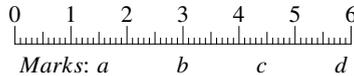

*Marks*: $a$     $b$     $c$     $d$

**Solution by ChatGPT**     **Wrong**

To construct a ruler with $M = 4$ marks ($a$, $b$, $c$, $d$) that can measure all integer distances up to length $L = 6$, with only one way of measuring each distance, we can use the following configuration: $a - b - c - d$. To measure the distances, we have the following markings on the ruler:

$a - b$: Distance 1

$a - c$: Distance 2

$a - d$: Distance 3

~~$b - c$: Distance 1 + 2 = 3~~ Inconsistency: GPT uses the configuration $a - b - c - d$.

~~$b - d$: Distance 2 + 1 = 3~~ Inconsistency: GPT uses the configuration $a - b - c - d$.

~~$c - d$: Distance 3 + 1 = 4.~~ Inconsistency: GPT uses the configuration $a - b - c - d$.

~~So far, we have covered distances 1, 2, 3, and 4.~~ Implication does not hold . Now, to cover distance 5, ~~we can use the distance between the midpoint of $a - c$ and $d$.~~ Inconsistency: This is a new mark Let's call the midpoint of $a - c$ as $e$: $a - e - c - d$

~~$a - e$: Distance 2 + 2 = 4~~ Logical fault

~~$e - c$: Distance 2~~ Logical fault

~~$e - d$: Distance 4 + 1 = 5~~ Logical fault

~~Now, we have covered distances 1, 2, 3, 4, and 5.~~ Wrong conclusion To cover distance 6, ~~we can use the distance between the midpoint of $a - d$ and $c$. Let's call the midpoint of $a - d$ as $f$~~ Inconsistency: this is a new mark $a - e - f - c - d$.

$e - f$: Distance 4 - 2 = 2

$f - c$: Distance 2

$f - d$: Distance 2 + 1 = 3

Now, we have covered distance 6. ~~So, this ruler configuration with marks $a$, $b$, $c$, and $d$ satisfies all the requirements: It can measure all integer distances up to length 6.~~ Wrong conclusion ~~There is only one way of measuring each distance.~~ Unsupported claim

~~Note that this configuration is unique, as there is only one way to measure each distance using this ruler.~~ Wrong conclusion: they are 2 solutions

> **Solution by human**
>
> Let $\mathcal{M} = \{a, b, c, d\}$ the set containing the four marks on the ruler. These marks should be distinct. The ruler is able to measure integers up to 6, hence we set the domain size to 7. To remove some isomorphic models, we set an order relation between the values $a$, $b$, $c$, and $d$, e.g. $a < b < c < d$. We help Mace4 by stating that the ruler starts with zero (i.e. $a = 0$). The ruler should measure all the integers up to 6. To be able to measure the value 1 we need:
>
> $$|a - b| = 1 \vee |b - c| = 1 \vee |c - d| = 1 \vee |a - c| = 1 \vee |a - d| = 1 \vee |b - d| = 1. \qquad (3.1)$$
>
> Similarly, to be able to measure the value 2, we need:
>
> $$|a - b| = 2 \vee |b - c| = 2 \vee |c - d| = 2 \vee |a - c| = 2 \vee |a - d| = 2 \vee |b - d| = 2. \qquad (3.2)$$
>
> Similar equations are needed for all the values up to 6 (lines 15-25). Note that this is a particular implementation for $M = 4$ and $L = 6$. A different approach is required to formalise a general solution for any $M$ and $L$. Mace4 computes the following two Golomb rulers:
>
> | Model | $a$ | $b$ | $c$ | $d$ |
> |-------|-----|-----|-----|-----|
> | 1     | 0   | 1   | 4   | 6   |
> | 2     | 0   | 2   | 5   | 6   |
>
> These two models are depicted in Figure 3.3, where we can see that all values up to 6 can be measured.



Figure 3.3: Two Golomb rulers for $M = 4$ marks and length $L = 6$

**Puzzle 31. Two cube calendar**

In Grand Central Terminal in New York I saw in a store window an unusual desk calendar. The day was indicated simply by arranging the two cubes so that their front faces gave the date. The face of each cube bore a single digit, 0 through 9, and one could arrange the cubes so that their front faces indicated any date from 01, 02, 03, . . . , to 31. On the left cube two values are visible: 1 and 2. On the right cube three values are visibile: 3, 4, and 5. What are the four digits that cannot be seen on the left cube and the three that cannot be seen on the right cube? It is a bit trickier than one might expect. (puzzle 1.4 from Gardner (2005))

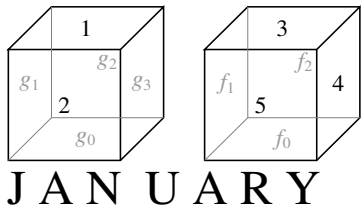

J A N U A R Y

**Solution by ChatGPT**                                              **Wrong**

To solve this puzzle, let's consider the visible digits on each cube. On the left cube (tens place of the date), two values are visible: 1 and 2. On the right cube (ones place of the date), three values are visible: 3, 4, and 5. Now, let's determine the digits that cannot be seen on each cube.

For the left cube, the visible values are 1 and 2. The valid two-digit numbers range from 01 to 31. ~~So, the digits that cannot be seen on the left cube are the ones that cannot form valid two-digit numbers when combined with 1 and 2.~~ <sub>Unrelated claim</sub> ~~The valid two-digit numbers with 1 in the tens place are 10, 11, 12, 13, 14, 15, 16, 17, 18, 19, 21, 31.~~ <sub>Bad arithmetic: 21 and 31 do not have 1 in the tens place</sub> ~~So, the digits that cannot be seen on the left cube when combined with 1 are: 0, 3, 4, 5, 6, 7, 8, and 9.~~ <sub>Inconsistency: based on the above two statements</sub>

**Solution by ChatGPT** **Wrong**

The valid two-digit numbers with 2 in the tens place are 20, 21, 22, 23, 24, 25, 26, 27, 28, 29. ~~So, the digit that cannot be seen on the left cube when combined with 2 is: 0.~~ [red: Inconsistency: 20 is a valid date]

~~Therefore, the four digits that cannot be seen on the left cube are: 0, 3, 4, and 5.~~ [red: Implication does not hold]

For the right cube, the visible values are 3, 4, and 5. The valid one-digit numbers range from 0 to 9. ~~So, the digits that cannot be seen on the right cube are the ones that cannot form valid one-digit numbers when combined with 3, 4, and 5.~~ [red: Unsupported claim]

~~The valid one-digit numbers with 3 in the ones place are 03, 13, 23, 30, 31.~~ [red: Inconsistency: 30 and 31 do not have 3 in the ones place] [red: Inconsistency: one-digit number doe snot have two digits] ~~So, the digits that cannot be seen on the right cube when combined with 3 are: 1, 4, 5, 6, 7, 8, and 9.~~ [red: Unsupported claim]

The valid ~~one-digit~~ [red: Inconsistency: one-digit number does not have two digits] numbers with 4 in the ones place are 04, 14, 24. ~~So, the digits that cannot be seen on the right cube when combined with 4 are: 0, 1, 2, 5, 6, 7, 8, and 9.~~ [red: Unsupported claim]

The valid ~~one-digit~~ [red: Inconsistency: one-digit number does not have two digits] numbers with 5 in the ones place are 05, 15, 25, ~~15~~. [red: Bad arithmetic: 15 appears twice] ~~So, the digit that cannot be seen on the right cube when combined with 5 is: 0.~~ [red: Unsupported claim]

~~Therefore, the three digits that cannot be seen on the right cube are: 0, 1, and 6.~~ [red: Unsupported claim]

To summarize: ~~Digits not visible on the left cube: 0, 3, 4, 5~~ [red: Wrong conclusionUnsupported claim] ~~Digits not visible on the right cube: 0, 1, 6~~ [red: Wrong conclusionUnsupported claim]

**Solution by human**

Mace4 finds a single model (see Figure 3.4):

| Model | $f_0$ | $f_1$ | $f_2$ | $g_0$ | $g_1$ | $g_2$ | $g_3$ |
|-------|-------|-------|-------|-------|-------|-------|-------|
| 1     | 0     | 1     | 2     | 0     | 6     | 7     | 8     |

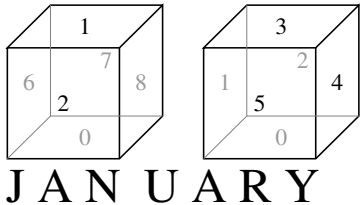

JANUARY

Figure 3.4: The solution for the two cube calendar

## Puzzle 32. How many routes?

In our Mathematics Circle we diagrammed 25 blocks of our city. They form a grid of $5 \times 5$. How many different routes can we draw from the bottom-left corner to the top-right corner moving only upward and to the right? Different routes may, of course, have portions that coincide. (puzzle 50 from Kordemsky (1992))

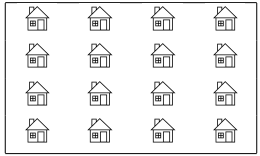

## Solution by ChatGPT                                    Correct

To solve this puzzle, we can use the concept of combinatorics and count the number of different routes from the bottom-left corner to the top-right corner of the $5 \times 5$ grid.

When moving from the bottom-left corner to the top-right corner, we need to take a total of 4 steps upward and 4 steps to the right. The order in which we take these steps will determine the different routes.

Since there are 4 steps upward and 4 steps to the right, the total number of different routes can be calculated using the binomial coefficient formula:

$C(n, k) = n!/(k! * (n - k)!)$

where $n$ is the total number of steps ($4 + 4 = 8$ in this case) and $k$ is the number of steps in one direction (4 upward steps or 4 steps to the right).

For our case, we need to calculate $C(8, 4)$:

$C(8, 4) = 8!/(4! * (8 - 4)!) = 70$

So, there are 70 different routes that can be drawn from the bottom-left corner to the top-right corner of the $5 \times 5$ grid by moving only upward and to the right.

## Solution by human

Observe that any route contains exactly 8 moves: 4 upward moves and 4 right moves. We code the upward move with 1 and the right move with 0. Hence the domain size is 2. As there are 4 right moves and 4 left moves, the sum of the moves $m_i$ should be exactly four (see Listing **??**)

$$m_1 + m_2 + m_3 + m_4 + m_5 + m_6 + m_7 + m_8 = 4 \tag{3.3}$$

Mace4 outputs 70 models for this equation.

**Puzzle 33. Going to church**

A man living in the house wants to know what is the greatest number of different routes by which he can go to the church. The house and the church are in a 5× 5 grid. The house is on the bottom left (or SW) position (1,1). The church is on the top right (or NE) position (5,5) The possible roads are indicated by the lines, and he always walks either to the N, to the E, or NE; that is, he goes so that every step brings him nearer to the church. Can you count the total number of different routes from which he may select? (puzzle 417 from Dudeney (2016))

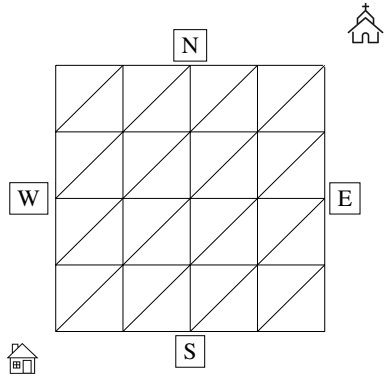

**Solution by ChatGPT** **Wrong**

To count the total number of different routes from the bottom-left position (1, 1) to the top-right position (5, 5) in a 5×5 grid, following the rules that he can only move to the North (N), to the East (E), or diagonally Northeast (NE), we can use dynamic programming.

Let's create a $5 \times 5$ grid to keep track of the number of routes from each position to the church (top-right corner). Here's how the grid looks initially:

$$? \quad ? \quad ? \quad ? \quad ?$$
$$? \quad ? \quad ? \quad ? \quad ?$$
$$? \quad ? \quad ? \quad ? \quad ?$$
$$? \quad ? \quad ? \quad ? \quad ?$$
$$? \quad ? \quad ? \quad ? \quad ?$$

We start from the church position (5, 5) and work our way backward to the house position (1, 1). The church position (5, 5) has only one route (it's the destination). The positions in the rightmost column (excluding the church) and the topmost row (excluding the church) ~~can only move directly towards the church (either N or E), so they all have one route each.~~ Incomplete search space: actually there are 3 moves (N, E, NE), instead of 2 For the other positions, the number of routes from a position is the sum of the number of routes from the positions to the right (E), above (N), and diagonally above-right (NE). After filling in all the routes, the completed grid will look like this:

| 70 | 35 | 15 | 5 | 1 |
|----|----|----|---|---|
| 35 | 20 | 10 | 4 | 1 |
| 15 | 10 | 6  | 3 | 1 |
| 5  | 4  | 3  | 2 | 1 |
| 1  | 1  | 1  | 1 | 1 |

The number at the bottom-left position (1, 1) represents the total number of different routes from the house to the church, following the given rules. In this case, there are 70 different routes.

**Solution by human**

There are 321 ways

**Puzzle 34. Buying wine**

A wine seller has two wine jugs, o small one of 4 liters capacity, and a larger one of 9 liters. There is no measuring label mentioned on either of these two jugs i.e. he cannot know the exact amount filled in the jug. Can we measure all values from 1 to 9 using these unmarked jugs? (a generalisation of classical water jugs problem)

**Solution by ChatGPT**                                                                 **Wrong**

Yes, it is possible to measure all values from 1 to 9 using the unmarked jugs of 4 liters and 9 liters. This problem can be solved using the concept of "Euclidean algorithm."^Solving strategy: Euclidean algorithm. Here's how it can be done:

1. Fill the 9-liter jug completely.

2. Pour the 9 liters from the large jug into the 4-liter jug. Now, the large jug has 9 - 4 = 5 liters of wine, and the small jug has 4 liters.

3. Empty the small jug.

4. Pour the 5 liters of wine from the large jug into the small jug. ~~Now, the large jug has 5 liters of wine, and the small jug has 4 liters.~~ ^Bad arithmetic: the large jug will have 5 liters, and the small 4

5. Fill the 9-liter jug again.

6. Carefully pour the wine from the 9-liter jug into the small jug until the small jug is full (which is 4 liters). After this step, there will be 5 liters of wine remaining in the large jug (9 - 4 = 5 liters).

7. Now, the small jug contains 4 liters of wine, and the large jug contains 5 liters of wine.

8. Empty the small jug.

9. Pour the remaining 5 liters of wine from the large jug into the small jug.

10. Fill the large jug again.

11. Carefully pour the wine from the 9-liter jug into the small jug until the small jug is full (which is 4 liters). ^Loop: This is the third repetition of the same sequence of moves ~~After this step, there will be 6 liters of wine remaining in the large jug (9 - 4 = 5 liters)~~^Bad arithmetic Inconsistency: 9 − 4 = 5 ≠ 4.

12. ~~Now, the small jug contains 4 liters of wine, and the large jug contains 6 liters of wine.~~^Inconsistency

13. Empty the small jug.

14. Pour the remaining 6 liters of wine from the large jug into the small jug.

15. Fill the large jug again.

16. Carefully pour the wine from the 9-liter jug into the small jug until the small jug is full (which is 4 liters). ~~After this step, there will be 2 liters of wine remaining in the large jug (9 - 4 = 5 liters)~~ ^Bad arithmetic Inconsistency: 9 − 4 = 5 ≠ 4.

17. Now, the small jug contains 4 liters of wine, and the large jug contains 2 liters of wine.

18. Empty the small jug.

19. Pour the remaining 2 liters of wine from the large jug into the small jug.

20. Fill the 9-liter jug again.

21. Pour the 9 liters from the large jug into the small jug. ~~Now, the small jug has 2 + 9 = 11 liters of wine~~^Lack of commonsense, which is more than its capacity. However, since the small jug can only hold 4 liters, it will be filled to its capacity, and ~~there will be 7 liters of wine remaining in the large jug (11 - 4 = 7 liters)~~.^Lack of commonsense

22. Now, the small jug contains 4 liters of wine, and the large jug contains 7 liters of wine.

23. Empty the small jug.

24. ~~Pour the remaining 7 liters of wine from the large jug into the small jug.~~^Lack of commonsense

25. Fill the 9-liter jug again.

26. Pour the 9 liters from the large jug into the small jug. Now, ~~the small jug has 7 + 9 = 16 liters of wine~~^Lack of commonsense  ^Inconsistency: 9 liters has the larger jar, not the smaller, which is more than its capacity. However, since the small jug can only hold 4 liters, it will be filled to its capacity, and ~~there will be 12 liters of wine remaining in the large jug (16 - 4 = 12 liters)~~^Lack of commonsense.

**Solution by ChatGPT**                                                                 **Wrong**

27. Now, the small jug contains 4 liters of wine, and the large jug contains 12 liters of wine.

28. Empty the small jug.

29. Pour the remaining 12 liters of wine from the large jug into the small jug.

30. Fill the 9-liter jug again.

31. Pour the 9 liters from the large jug into the small jug. ~~Now, the small jug has $12 + 9 = 21$ liters of wine~~[Lack of commonsense], which is more than its capacity. However, since the small jug can only hold 4 liters, it will be filled to its capacity, and ~~there will be 17 liters of wine remaining in the large jug ($21 - 4 = 17$ liters).~~[Lack of commonsense]

32. Now, the small jug contains 4 liters of wine, and the large jug contains 17 liters of wine.

33. Empty the small jug.

34. ~~Pour the remaining 17 liters of wine from the large jug into the small jug.~~[Lack of commonsense]

35. Fill the 9-liter jug again.

36. Pour the 9 liters from the large jug into the small jug. ~~Now, the small jug has $17 + 9 = 26$ liters of wine~~[Lack of commonsense], which is more than its capacity. However, since the small jug can only hold 4 liters, it will be filled to its capacity, and ~~there will be 22 liters of wine remaining in the large jug ($26 - 4 = 22$ liters)~~[Lack of commonsense].

37. Now, the small jug contains 4 liters of wine, and the large jug contains 22 liters of wine.

38. Empty the small jug.

39. ~~Pour the remaining 22 liters of wine from the large jug into the small jug.~~[Lack of commonsense]

40. Fill the 9-liter jug again.

41. Pour the 9 liters from the large jug into the small jug. Now, ~~the small jug has $22 + 9 = 31$ liters of wine~~[Lack of commonsense], ~~which is exactly its capacity~~.
[Inconsistency:  GPT learned that the capacity is 9] There will be no wine remaining in the large jug.

42. Now, the small jug contains 4 liters of wine, and the large jug is empty.

~~By following the above steps, the wine seller can measure all values from 1 to 9 using the unmarked jugs of 4 liters and 9 liters.~~[Wrong conclusion  Unsupported claim]

**Solution by human**

Let $j_1 = 4$ be the capacity of the small jug and $j_2 = 9$ the capacity of the large jug. We use a *state space search*, where each state is represented with $J(x, y)$: $x$ is the current amount of wine in jug $j_1$ and $y$ the amount in jug $j_2$. The initial state is $J(0, 0)$. For each measured value, Prover9 computes the corresponding proof. The theorem asks to find a state in which either the first jug or the second one contains only 1 liter. For $\exists x\, J(x, 1) \lor \exists y\, J(1, y)$, the $Proof_1$ has 5 steps.

| $Proof_1$ | #answer(Action) | #answer(State) |
|---|---|---|
| 1 | "init state" | $J(0, 0)$ |
| 2 | "fill the big jug" | $J(0, j_2)$ |
| 3 | "big into small, until full" | $J(j_1, 9 - (j_1 - 0))$ |
| 4 | "empty the small jug" | $J(0, 5)$ |
| 5 | "big into small, until full" | $J(j_1, 5 - (j_1 - 0))$ |

For measuring two liters, we need $\exists x\, J(x, 2) \lor \exists y\, J(2, y)$. Prover9 finds the $Proof(2)$ in 11 steps:

| $Proof_2$ | #answer(Action) | #answer(State) |
|---|---|---|
| 1 | "init state" | $J(0, 0)$ |
| 2 | "fill the big jug" | $J(0, j_2)$ |
| 3 | "big into small, until full" | $J(j_1, 9 - (j_1 - 0))$ |
| 4 | "empty the small jug" | $J(0, 5)$ |
| 5 | "big into small, until full" | $J(j_1, 5 - (j_1 - 0))$ |
| 6 | "empty the small jug" | $J(0, 1)$ |
| 7 | "empty the big jug into the small jug" | $J(0 + 1, 0)$ |
| 8 | "fill the big jug" | $J(1, j_2)$ |
| 9 | "big into small, until full" | $J(j_1, 9 - (j_1 - 1))$ |
| 10 | "empty the small jug" | $J(0, 6)$ |
| 11 | "big into small, until full" | $J(j_1, 6 - (j_1 - 0)))$ |

Note that the first five steps in $Proof_2$ are similar with $Proof_1$. The value 2 is obtained in the big jug. For measuring three liters with $\exists x\, J(x, 3) \lor \exists y\, J(3, y)$, the $Proof_3$ has 7 steps. The value 3 is obtained in the small jug.

| $Proof_3$ | #answer(Action) | #answer(State) |
|---|---|---|
| 1 | "init state" | $J(0, 0)$ |
| 2 | "fill the small jug" | $J(j_1, 0)$ |
| 3 | "empty the small jug into the big jug" | $J(0, 4 + 0)$ |
| 4 | "fill the small jug" | $J(j_1, 4)$ |
| 5 | "empty the small jug into the big jug" | $J(0, 4 + 4)$ |
| 6 | "fill the small jug" | $J(j_1, 8)$ |
| 7 | "small into big, until full" | $J(4 - (j_2 - 8), j_2)$ |

$\exists x\, J(x, 5) \lor \exists y\, J(5, y)$, the $Proof_5$ has only 3 steps. The value 5 is obtained in the large jug.

| $Proof_5$ | #answer(Action) | #answer(State) |
|---|---|---|
| 1 | "init state" | $J(0, 0)$ |
| 2 | "fill the big jug" | $J(0, j_2)$ |
| 3 | "big into small, until full" | $J(j_1, 9 - (j_1 - 0))$ |

**Solution by human**

For $\exists x\, J(x, 6) \vee \exists y\, J(6, y)$, the $Proof_6$ has 9 steps. The steps in $Proof_6$ represent the first 9 steps in $Proof_2$.

| $Proof_6$ | #answer(Action) | #answer(State) |
|---|---|---|
| 1 | "init state" | $J(0, 0)$ |
| 2 | "fill the big jug" | $J(0, j_2)$ |
| 3 | "big into small, until full" | $J(j_1, 9 - (j_1 - 0))$ |
| 4 | "empty the small jug" | $J(0, 5)$ |
| 5 | "big into small, until full" | $J(j_1, 5 - (j_1 - 0))$ |
| 6 | "empty the small jug" | $J(0, 1)$ |
| 7 | "empty the big jug into the small jug" | $J(0 + 1, 0)$ |
| 8 | "fill the big jug" | $J(1, j_2)$ |
| 9 | big into small, until full" | $J(j_1, 9 - (j_1 - 1))$ |

$Proof_7$ continues $Proof_3$ with the following four steps:

| $Proof_7$ | #answer(Action) | #answer(State) |
|---|---|---|
| 8 | "empty the big jug" | $J(3, 0)$ |
| 9 | "empty the small jug into the big jug" | $J(0, 3 + 0)$ |
| 10 | "fill the small jug" | $J(j_1, 3)$ |
| 11 | "empty the small jug into the big jug" | $J(0, 4 + 3)$ |

Finally, $Proof_8$ is the same with the first 5 steps in $Proof_3$.
The following relations between proofs are identified:

$$\begin{aligned} Proof_1 + [6, 7, 8, 9, 10, 11] &= Proof_2 \\ Proof_8 + [6, 7] &= Proof_3 \\ Proof_3 + [8, 9, 10, 11] &= Proof_7 \\ Proof_6 + [10, 11] &= Proof_2 \\ Proof_5 + [4, 5] &= Proof_1 \end{aligned}$$

A lattice can be built with the following subsumption relations between proofs:

$$Proof_8 \sqsubseteq Proof_3$$
$$Proof_5 \sqsubseteq Proof_1 \sqsubseteq Proof_2$$
$$Proof_6 \sqsubseteq Proof_2$$

The graph below represents a simplified railway system, and we want to know how many different ways there are of going from A to E, if we never go twice along the same line in any journey. From A to B there is route r1. From B to C there are three routes: denoted with r3, r5, and r7. From B to D there are 2 routes: r2 and r9. From C to D there are three routes: denoted with r4, r6, and r8. From D to E there is a single route: r10. Of course, if there is a route from x to y, then there is a route from y to x. This is a very simple proposition, but practically impossible to solve until you have hit on some method of recording the routes. You see there are many ways of going, from the short route ABDE, taking one of the large loops, up to the long route ABCDBCDBCDE, which takes you over every line on the system and can itself be varied in order in many ways. How many different ways of going are there? (puzzle 424 from Dudeney (2016))

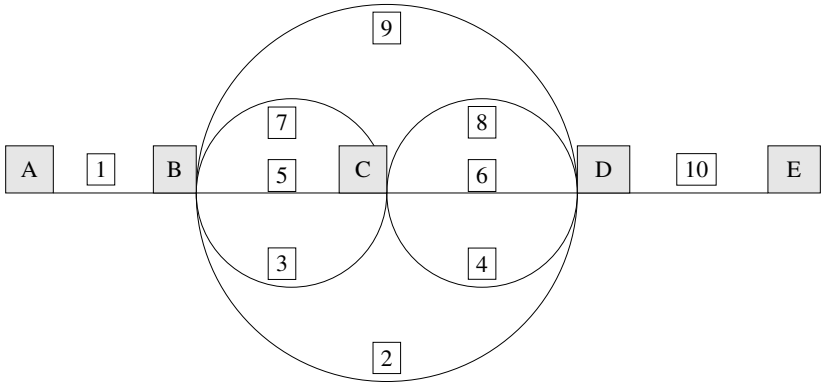

Solution by ChatGPT                                                                                    wrong

To find the total number of different ways of going from A to E in the given railway system, we can use the concept of graph theory and apply depth-first search (DFS) or breadth-first search (BFS) algorithms to explore all possible routes without repeating any line. ~~Let's represent the graph using the information given:~~ Inconsistency: the graph does not correspond with the description

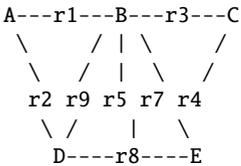

**Solution by ChatGPT**                                                      **Wrong**

Starting from A, we explore all possible paths to reach E while making sure that no line is repeated in any journey. Here's the list of possible routes from A to E:

1. A - r1 - B - r3 - C - r4 - D - r10 - E
2. ~~A - r1 - B - r3 - C - r4 - D - r8 - E~~ [Inconsistency:  r8 does not connect D with E]
3. A - r1 - B - r3 - C - r8 - D - r10 - E
4. A - r1 - B - r5 - C - r4 - D - r10 - E
5. ~~A - r1 - B - r5 - C - r4 - D - r8 - E~~ [Inconsistency:  r8 does not connect D with E]
6. A - r1 - B - r5 - C - r8 - D - r10 - E
7. A - r1 - B - r7 - C - r4 - D - r10 - E
8. ~~A - r1 - B - r7 - C - r4 - D - r8 - E~~ [Inconsistency:  r8 does not connect D with E]
9. A - r1 - B - r7 - C - r8 - D - r10 - E
10. ~~A - r2 - D - r4 - C - r3 - B - r1 - E~~ [Inconsistency:  r2 does not connect A with D]
11. ~~A  -  r2  -  D  -  r4  -  C  -  r3  -  B  -  r5  -
E~~ [Inconsistency:  r2 does not connect A with D][Inconsistency:  r5 does not connect B with E]
12. ~~A  -  r2  -  D  -  r4  -  C  -  r5  -  B  -  r1  -
E~~ [Inconsistency:  r2 does not connect A with D][Inconsistency:  r1 does not connect B with E]
13. ~~A  -  r2  -  D  -  r4  -  C  -  r5  -  B  -  r3  -
E~~ [Inconsistency:  r2 does not connect A with D][Inconsistency:  r3 does not connect B with E]
14. ~~A  -  r2  -  D  -  r4  -  C  -  r7  -  B  -  r1  -
E~~ [Inconsistency:  r2 does not connect A with D][Inconsistency:  r1 does not connect B with E]
15. ~~A  -  r2  -  D  -  r4  -  C  -  r7  -  B  -  r3  -
E~~ [Inconsistency:  r2 does not connect A with D][Inconsistency:  r3 does not connect B with E]
16. ~~A  -  r2  -  D  -  r4  -  C  -  r8  -  B  -  r1  -
E~~ [Inconsistency:  r2 does not connect A with D][Inconsistency:  r1 does not connect B with E  Inconsistency:  r8 does not connect C with B]
17. ~~A  -  r2  -  D  -  r4  -  C  -  r8  -  B  -  r3  -
E~~ [Inconsistency:  r2 does not connect A with D][Inconsistency:  r3 does not connect B with E  Inconsistency:  r8 does not connect C with B]
18. ~~A  -  r9  -  D  -  r4  -  C  -  r3  -  B  -  r1  -
E~~ [Inconsistency:  r9 does not connect A with D][Inconsistency:  r1 does not connect B with E]
19. ~~A  -  r9  -  D  -  r4  -  C  -  r3  -  B  -  r5  -
E~~ [Inconsistency:  r9 does not connect A with D][Inconsistency:  r5 does not connect B with E]
20. ~~A  -  r9  -  D  -  r4  -  C  -  r5  -  B  -  r1  -
E~~ [Inconsistency:  r9 does not connect A with D][Inconsistency:  r1 does not connect B with E]
21. ~~A  -  r9  -  D  -  r4  -  C  -  r5  -  B  -  r3  -
E~~ [Inconsistency:  r9 does not connect A with D][Inconsistency:  r3 does not connect B with E]
22. ~~A  -  r9  -  D  -  r4  -  C  -  r7  -  B  -  r1  -
E~~ [Inconsistency:  r9 does not connect A with D][Inconsistency:  r1 does not connect B with E]
23. ~~A  -  r9  -  D  -  r4  -  C  -  r7  -  B  -  r3  -
E~~ [Inconsistency:  r9 does not connect A with D][Inconsistency:  r3 does not connect B with E]
24. ~~A  -  r9  -  D  -  r4  -  C  -  r8  -  B  -  r1  -
E~~ [Inconsistency:  r9 does not connect A with D][Inconsistency:  r1 does not connect B with E  Inconsistency:  r8 does not connect C with B]
25. ~~A  -  r9  -  D  -  r4  -  C  -  r8  -  B  -  r3  -
E~~ [Inconsistency:  r9 does not connect A with D][Inconsistency:  r3 does not connect B with E]
[Inconsistency:  r8 does not connect C with B][Inconsistency:  r8 does not connect C with B]

~~There are a total of 25 different routes from A to E without repeating any line in any journey.~~ [Wrong conclusion]

---

**Solution by human**

There are 2,501 ways of going from *A* to *E*. Since there is only one route from *A* to *B* and also one route from *D* to *E*, this value of 2,501 is the same number as going from *B* to *D*. Mace4 finds all 2501 solutions, the first one being:

| | | | |
|---|---|---|---|
| $function(A,$ | $[0]),$ | function(B, | $[1]),$ |
| $function(C,$ | $[2]),$ | function(D, | $[3]),$ |
| $function(E,$ | $[4]),$ | function(c1, | $[1]),$ |
| $function(f(\_),$ | $[0,1,2,4,3,5,6,8,7,9,10]),$ | $function(g(\_),$ | $[1,2,4,3,5,6,8,7,9,10,0]),$ |
| $function(p(\_),$ | $[0,0,1,2,3,4,5,6,7,8,9]),$ | $function(s(\_),$ | $[1,2,3,4,5,6,7,8,9,10,0])$ |
| $function(n(\_),$ | $[0,1,3,2,1,2,3,2,1,3,4])$ | | |

The route is stored by the *f* function. It represents the long route:

$$A \xrightarrow{1} B \xrightarrow{2} D \xrightarrow{4} C \xrightarrow{3} B \xrightarrow{5} C \xrightarrow{6} D \xrightarrow{8} C \xrightarrow{7} B \xrightarrow{9} D \xrightarrow{10} E$$

In the model 2501, $function(f(\_), [0, 0, 0, 0, 0, 0, 0, 0, 1, 9, 10])$ stores the short route:

$$A \xrightarrow{1} B \xrightarrow{9} D \xrightarrow{10} E$$

**Puzzle 36. A stolen balsam**

Three men robbed a gentleman of a vase containing 24 ounces of balsam. While running away, they met in a forest a glass seller, from whom, in a great hurry, they purchased three vessels. On reaching a place of safety they wished to divide the booty, but they found that their vessels contained 5, 11, and 13 ounces respectively. How could they divide the balsam into equal portions? (puzzle 409 from Dudeney (2016))

**Solution by ChatGPT**                                                                           **Wrong**

To divide the 24 ounces of balsam into equal portions among the three men using the vessels of 5, 11, and 13 ounces, follow these steps:

1. Pour 13 ounces from the 24-ounce vase into the 13-ounce vessel. The 13-ounce vessel now contains 13 ounces of balsam, and the 24-ounce vase has 24 - 13 = 11 ounces of balsam remaining.

2. Pour 11 ounces from the 24-ounce vase into the 11-ounce vessel. The 11-ounce vessel now contains 11 ounces of balsam, and ~~the 24-ounce vase has 24 - 11 = 13 ounces of balsam remaining.~~ <sup style="color:red">Inconsistency: the 11-ounce vase is now empty</sup>

3. Pour the entire contents of the 13-ounce vessel (which is 13 ounces) into the 5-ounce vessel. ~~The 5-ounce vessel now contains 5 + 13 = 18 ounces of balsam~~<sup style="color:red">Lack of commonsense</sup>, and the 13-ounce vessel is empty.

4. Pour the entire contents of the 11-ounce vessel (which is 11 ounces) into the 13-ounce vessel. The 13-ounce vessel now contains 11 ounces of balsam, and ~~the 11-ounce vessel is empty.~~<sup style="color:red">Bad arithmetic</sup>

5. Pour the remaining 11 ounces from the 13-ounce vessel into the 11-ounce vessel. ~~The 11-ounce vessel now contains 11 + 11 = 22 ounces of balsam~~<sup style="color:red">Lack of commonsense</sup>, and the 13-ounce vessel is empty.

┌─ **Solution by ChatGPT** ──────────────────────────── **Wrong** ─┐

6. Pour the entire contents of the 11-ounce vessel (which is 22 ounces) into the 5-ounce vessel. ~~The 5-ounce vessel now contains 18 + 22 = 40 ounces of balsam.~~<sup>Lack of commonsense</sup>
<span style="color:red">Unsupported claim: there are no 18 ounces in the smallest vase</span>

7. Finally, ~~pour 4 ounces of balsam from the 5-ounce vessel back into the 24-ounce vase~~<sup>Lack of commonsense</sup>. ~~The 5-ounce vessel now contains 40 - 4 = 36 ounces of balsam~~<sup>Lack of commonsense</sup>, and ~~the 24-ounce vase contains 11 + 4 = 15 ounces of balsam.~~
<span style="color:red">Inconsistency: the 24 ounce vase does not contain 11 ounces</span>

~~Now, each vessel contains exactly 12 ounces of balsam, and the three men have divided the booty equally.~~ <span style="color:red">Inconsistency: GPT has just computes that the 24-ounce case contains 11 ounces.</span>
<span style="color:red">Bad arithmetic: 12*4=48 ounces instead of 24. Lack of commonsense: vases cannot acomodate 12 ounces, for they are smaller.</span>

<span style="color:red">Wrong conclusion</span>

└──────────────────────────────────────────────────────┘

┌─ **Solution by human** ─┐

Let the state $J(x, y, z, t)$ meaning that the first vessel contains $x = 24$ ounces of balsam, the second one can accommodate $y \leq 11$ ounces, the third one $z \leq 13$ and the $4^{th}$ vessel $t \leq 5$ ounces: $J(24, 11, 13, 5)$. The init state is $J(24, 0, 0, 0)$. Since the last vessel cannot accommodate 8 ounces, there is only one possible goal state to equally split the total amount of 24 ounces: $J(8, 8, 8, 0)$. Prover9 found a five-steps solution:

| Step  | 0             | 1             | 2             | 3             | 4            | 5            |
|-------|---------------|---------------|---------------|---------------|--------------|--------------|
| State | $J(24,0,0,0)$ | $J(19,0,0,5)$ | $J(8,11,0,5)$ | $J(8,11,5,0)$ | $J(8,3,8,5)$ | $J(8,8,8,0)$ |

└──────────────────────────────────────────────────────┘

# References

Brooke, M. (1963). *150 Puzzles in Crypt-arithmetic*, volume 1039. Dover Publications.

Danesi, M. (2018). *Ahmes' Legacy Puzzles and the Mathematical Mind*. Springer.

Darling, D. (2004). *The Universal Book of Mathematics from Abracadabra to Zeno's Paradoxes*. John Wiley & Sons, Inc.

Dudeney, H. E. (2016). *536 Puzzles and curious problems*. Courier Dover Publications.

Gardner, M. (2005). *The Colossal Book of Short Puzzles and Problems*. W.W. Norton.

Gelfond, M. and Przymusinska, H. (1986). Negation as failure: Careful closure procedure. *Artificial Intelligence*, 30(3):273–287.

Hadley, J. and Singmaster, D. (1992). Problems to sharpen the young. *The Mathematical Gazette*, 76(475):102–126.

Kordemsky, B. A. (1992). *The Moscow puzzles: 359 mathematical recreations*. Courier Corporation.

Memarsadeghi, N. (2016). NASA computational case study: Golomb rulers and their applications. *Computing in Science & Engineering*, 18(6):58–62.

Polash, M. A., Newton, M. H., and Sattar, A. (2017). Constraint-based search for optimal golomb rulers. *Journal of Heuristics*, 23(6):501–532.

Reger, G., Suda, M., and Voronkov, A. (2016). Finding finite models in multi-sorted first-order logic. In *International Conference on Theory and Applications of Satisfiability Testing*, pages 323–341. Springer.

# Chapter 4

# Lady and tigers

**Puzzle 37. Ladies and tigers in jail**

Each of the two rooms contains either a lady or a tiger. There could be tigers in both rooms or ladies in both rooms. There is a sign on each room. The sign on the first room says: In this room there is a lady and in the other room there is a tiger. The sign on the second room says: In one of these rooms there is a lady and in one of these rooms there is a tiger. One of the signs is true and the other is false. Which door to open in order to find the lady? (adapted from Smullyan (2009))

*Room₁*

*In this room there is a lady and in the other room there is a tiger*

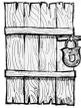

*Room₂*

*In one of these rooms there is a lady and in one of these rooms there is a tiger*

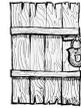

**Solution by ChatGPT**                    **Correct (but unjustified)**

Let's consider the two possibilities:

If the sign on the first room is true, then there is a lady in the first room and a tiger in the second room. This means the sign on the second room is false, so there is no lady in the second room. ~~Therefore, the lady must be in the first room.~~ <sup>Unsupported claim: there can be two tigers</sup>

~~If the sign on the first room is false, then there is a tiger in the first room and a lady in the second room.~~ <sup>Unsupported claim: there can be two tigers.</sup> This means the sign on the second room is true, so there is a lady in one of the rooms (the second room). ~~Therefore, the lady must also be in the second room.~~<sup>Inconsistency: GPT learns from the first posibility that lady is in the first room</sup> ~~In both cases, the lady is in the second room.~~<sup>Inconsistency</sup>. ~~So~~<sup>Unsupported claim: even if the conclusion is correct</sup>, you should open the second door to find the lady.

69

**Solution by human**

Let the formalisation: lady is in room 1 ($L_1$), lady is in room 2 ($L_2$), tiger is in room 1 ($T_1$), tiger is in room 2 ($T_2$), message in room 1 is true ($R_1$), and message in room 2 is true ($R_2$). Mace4 outputs a single model, in which the message $R_1$ is false and $R_2$ is true:

| Model | $L_1$ | $L_2$ | $T_1$ | $T_2$ | $R_1$ | $R_2$ |
|-------|-------|-------|-------|-------|-------|-------|
| 1     | 0     | 1     | 1     | 0     | 0     | 1     |

*Room₁*
*In this room there is a lady and*
*in the other room there is a tiger*

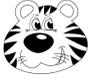

*Room₂*
*In one of these rooms there is a lady and*
*in one of these rooms there is a tiger*

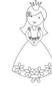

> **Puzzle 38. The trials of the second day**
>
> Each of the two rooms contains either a lady or a tiger. There could be tigers in both rooms or ladies in both rooms. There is a sign on each room. The sign on the first room says: "At least one of these rooms contains a lady". The sign on the second room says: "A tiger is in the other room". The signs are either both true or both false. Which door to open in order to find the lady? (adapted from Smullyan (2009))

<div align="center">

*Room₁*
*At least one of these rooms contains a lady*

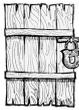

</div>

<div align="center">

*Room₂*
*A tiger is in the other room*

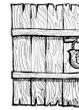

</div>

> **Solution by ChatGPT**      **Correct (but unjustified)**
>
> Let's consider the two possibilities:
> If the sign on the first room is true, ~~then there is at least one lady in the first room.~~ <sup>Unsupported claim</sup> This means the sign on the second room is also true, so there is a tiger in the first room (since a tiger is in the other room). Therefore, there is a lady in the second room.
> If the sign on the first room is false, then there are no ladies in the first room. This means the sign on the second room is also false, ~~so there are no tigers in the second room~~<sup>Unsupported claim</sup> (since a tiger is not in the other room). ~~Therefore, there are ladies in both rooms.~~<sup>Inconsistency: GPT already knows there is a tiger in room 1.</sup> In both cases, the lady is in the second room. So, you should open the second door to find the lady.

> **Solution by human**
>
> Mace4 computes a single model, in which both messages $R_1$ and $R_2$ are true:
>
> | Model | $L_1$ | $L_2$ | $T_1$ | $T_2$ | $R_1$ | $R_2$ |
> |-------|-------|-------|-------|-------|-------|-------|
> | 1     | 0     | 1     | 1     | 0     | 1     | 1     |

<div align="center">

*Room₁*
*At least one of these rooms contains a lady*

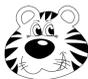

</div>

<div align="center">

*Room₂*
*A tiger is in the other room*

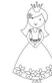

</div>

**Puzzle 39. Ladies only**

Each of the two rooms contains either a lady or a tiger. There could be tigers in both rooms or ladies in both rooms. There is a sign on each room. The sign on the first room says: "Either a tiger is in this room or a lady is in the other room". The sign on the second room says: "A lady is in the other room". The signs are either both true of both false. Which door to open in order to find the lady??  Smullyan (2009)

*Room₁*
*Either a tiger is in this room*
*or a lady is in the other room*

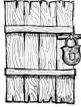

*Room₂*
*A lady is in the other room*

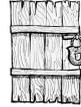

**Solution by ChatGPT**                                                                                      **Wrong**

Let's consider the two possibilities:

If the sign on the first room is true, then either a tiger is in the first room or a lady is in the second room. This means the sign on the second room is also true, so there is a lady in the first room (since a lady is in the other room). Therefore, there is a lady in the first room.

If the sign on the first room is false, then neither a tiger is in the first room nor a lady is in the second room. This means the sign on the second room is also false, so there is a tiger in the first room (since a lady is not in the other room). Therefore, there is a tiger in the first room.

~~In both cases, the lady is in the first room.~~ Inconsistency:  in the second case, GPT learnt there is a tiger in the first room
~~So, you should open the first door to find the lady.~~ Wrong conclusion:  you can open any door

**Solution by human**

Mace4 finds a single model in which both messages are true:

| Model | $L_1$ | $L_2$ | $T_1$ | $T_2$ | $R_1$ | $R_2$ |
|-------|-------|-------|-------|-------|-------|-------|
| 1     | 1     | 1     | 0     | 0     | 1     | 1     |

*Room₁*
*Either a tiger is in this room*
*or a lady is in the other room*

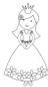

*Room₂*
*A lady is in the other room*

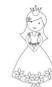

**Puzzle 40. Fourth day: Ladies are honest but tigers are liars**

Each of the two rooms contains either a lady or a tiger. There could be tigers in both rooms or ladies in both rooms. There is a sign on each room. The sign on the first room says: "Both rooms contain ladies". The sign on the second room says: "Both rooms contain ladies". If a lady is in room 1, then the sign on the door is true. If a tiger is in room 1, then the sign on the door is false. For room 2, the situation is opposite. If a lady is in room 2, then the sign on the door is false. If a tiger is in room 2, then the sign on the door is true. The signs are either both true of both false. Which door to open in order to find the lady? Smullyan (2009)

*Room₁*
*Both rooms contain ladies*

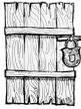

*Room₂*
*Both rooms contain ladies*

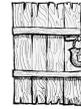

**Solution by ChatGPT**          **Wrong (claiming there is no solution)**

Let's analyze the possibilities:
If both signs are true, then both rooms must contain ladies. However, this creates a contradiction because one sign says that if a lady is in room 1, then the sign on the door is true, ~~but the other sign says that if a lady is in room 1~~ [Inconsistency: room 2 instead of room 1] [Lack of understanding: the other sign doesn't state that ladies in room 2 are liers, but the assumptions in the puzzle], then the sign on the door is false. This means that both signs cannot be true.
If both signs are false, then neither room contains a lady. This also creates a contradiction because ~~one sign~~ [Lack of understanding: the sign doesn't state this] says that if a lady is in room 2, then the sign on the door is false, but ~~the other sign says~~ [Lack of understanding: the sign doesn't state this] that if a lady is in room 2 [Inconsistency: small but relevant], then the sign on the door is true. ~~So, both signs cannot be false either.~~ [Unsupported claim]
Since both possibilities lead to contradictions, there must be an error in the problem statement or signs. ~~The puzzle cannot be solved with the given information.~~ [Wrong conclusion puzzle cannot be solved]

**Solution by human**

Mace4 finds a single model in which both messages $R_1$ and $R_2$ are false:

| Model | $L_1$ | $L_2$ | $T_1$ | $T_2$ | $R_1$ | $R_2$ |
|-------|-------|-------|-------|-------|-------|-------|
| 1     | 0     | 1     | 1     | 0     | 0     | 0     |

*Room₁*
*Both rooms contain ladies*

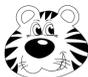

*Room₂*
*Both rooms contain ladies*

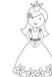

**Puzzle 41. Fifth day**

Each of the two rooms contains either a lady or a tiger. There could be tigers in both rooms or ladies in both rooms. There is a sign on each room. The sign on the first room says: "At least one room contains a lady". The sign on the second room says: "The other room contains a lady". If a lady is in room 1, then the sign on the door is true. If a tiger is in room 1, then the sign on the door is false. For room 2, the situation is opposite. If a lady is in room 2, then the sign on the door is false. If a tiger is in room 2, then the sign on the door is true. The signs are either both true of both false. Which door to open in order to find the lady?  Smullyan (2009)

$Room_1$
*At least one room contains a lady*

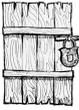

$Room_2$
*The other room contains a lady*

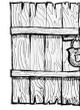

**Solution by ChatGPT**                                                                                        **Wrong**

Let's consider the two possibilities:
If the sign on the first room is true, then at least one room contains a lady. This means the sign on the second room is also true, so the other room contains a lady ~~as well~~
Unsupported claim:  we do not know there is a lady in the second room. ~~Therefore, there is a lady in both rooms~~ Unsupported claim  Wrong claim

If the sign on the first room is false, then neither room contains a lady. This means the sign on the second room is also false, so the other room does not contain a lady either. Therefore, there are tigers in both rooms.
~~In both cases, there is at least one lady.~~ Inconsistency:  GPT has just learned there are tigers in both rooms ~~So, you should open either door to find a lady.~~ Wrong conclusion

**Solution by human**

Mace4 finds a single model in which both messages $R_1$ and $R_2$ are true.

| Model | $L_1$ | $L_2$ | $T_1$ | $T_2$ | $R_1$ | $R_2$ |
|-------|-------|-------|-------|-------|-------|-------|
| 1     | 1     | 0     | 0     | 1     | 1     | 1     |

$Room_1$
*At least one room contains a lady*

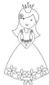

$Room_2$
*The other room contains a lady*

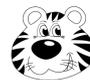

**Puzzle 42. Sixth day**

Each of the two rooms contains either a lady or a tiger. There could be tigers in both rooms or ladies in both rooms. There is a sign on each room. The sign on the first room says: "It makes no difference which room you pick". The sign on the second room says: "The other room contains a lady". If a lady is in room 1, then the sign on the door is true. If a tiger is in room 1, then the sign on the door is false. For room 2, the situation is opposite. If a lady is in room 2, then the sign on the door is false. If a tiger is in room 2, then the sign on the door is true. The signs are either both true of both false. Which door to open in order to find the lady?   Smullyan (2009)

*Room₁*
*It makes no difference which room you pick*

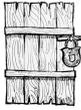

*Room₂*
*The other room contains a lady*

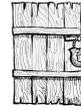

**Solution by ChatGPT**          **Wrong (claiming not enough information)**

Let's analyze the possibilities:
If both signs are true, then it makes no difference which room you pick, and the other room contains a lady. This means that both rooms must contain ladies, which creates a contradiction because ~~the puzzle allows for the possibility of both rooms containing tigers.~~<sup>Wrong justification</sup>
If both signs are false, ~~then it makes no difference which room you pick~~<sup>Unsupported claim</sup>, and the other room does not contain a lady. ~~This means that both rooms must contain tigers~~<sup>Unsupported claim</sup>, which also creates a contradiction.
Since both possibilities lead to contradictions, there must be an error in the problem statement or signs. ~~The puzzle cannot be solved with the given information.~~<sup>Wrong conclusion</sup>

**Solution by human**

Mace4 finds a single model in which both messages are false.

| Model | $L_1$ | $L_2$ | $T_1$ | $T_2$ | $R_1$ | $R_2$ |
|-------|-------|-------|-------|-------|-------|-------|
| 1     | 0     | 1     | 1     | 0     | 0     | 0     |

*Room₁*
*It makes no difference which room you pick*

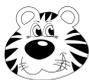

*Room₂*
*The other room contains a lady*

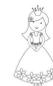

**Puzzle 43. Seventh day**

Each of the two rooms contains either a lady or a tiger. There could be tigers in both rooms or ladies in both rooms. There is a sign on each room. The sign on the first room says: "It makes a difference which room you pick". The sign on the second room says: "You are better off choosing the other room". If a lady is in room 1, then the sign on the door is true. If a tiger is in room 1, then the sign on the door is false. For room 2, the situation is opposite. If a lady is in room 2, then the sign on the door is false. If a tiger is in room 2, then the sign on the door is true. The signs are either both true of both false. Which door to open in order to find the lady?   Smullyan (2009)

*Room$_1$*
*It makes a difference which room you pick*

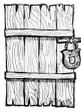

*Room$_2$*
*You are better off choosing the other room*

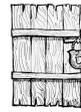

**Solution by ChatGPT**                    **Wrong (claiming not enough information)**

Let's analyze the possibilities:
If both signs are true, then it makes a difference which room you pick, and you are better off choosing the other room. ~~This means that both rooms must contain ladies~~Unsupported claim, which creates a contradiction because ~~the puzzle allows for the possibility of both rooms containing tigers.~~Wrong justification

If both signs are false, then it makes no difference which room you pick, and you are not better off choosing the other room. ~~This means that both rooms must contain tigers~~Unsupported claim, ~~which also creates a contradiction~~Unsupported claim

Since both possibilities lead to contradictions, there must be an error in the problem statement or signs. ~~The puzzle cannot be solved with the given information.~~Wrong conclusion

**Solution by human**

Mace4 finds a single model in which both messages are true.

| Models | $L_1$ | $L_2$ | $T_1$ | $T_2$ | $R_1$ | $R_2$ |
|--------|-------|-------|-------|-------|-------|-------|
| 1      | 1     | 0     | 0     | 1     | 1     | 1     |

*Room$_1$*
*It makes a difference which room you pick*

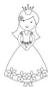

*Room$_2$*
*You are better of choosing the other room*

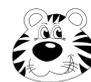

**Puzzle 44. Eighth day**

Each of the two rooms contains either a lady or a tiger. There could be tigers in both rooms or ladies in both rooms. There is a sign on each room. If a lady is in room 1, then the sign on the door is true. If a tiger is in room 1, then the sign on the door is false. For room 2, the situation is opposite. If a lady is in room 2, then the sign on the door is false. If a tiger is in room 2, then the sign on the door is true. Now, the problem is that signs are not already put on the doors. That is, the two messages can appear either on the first door or on the second one. One sign says: "This room contains a tiger". The other sign says: "Both rooms contain tigers". The signs are either both true of both false. Which door to open in order to find the lady?   Smullyan (2009)

*Room₁* or *Room₂*?
*This room contains a tiger*

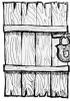

*Room₁* or *Room₂*?
*Both rooms contain tigers*

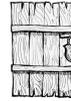

**Solution by ChatGPT**     **Wrong**

Let's analyze the possibilities: If both signs are true, then one room contains a tiger, and the other room also contains a tiger. ~~This creates a contradiction because the puzzle allows for the possibility of both rooms containing ladies.~~ᵂʳᵒⁿᵍ ʲᵘˢᵗⁱᶠⁱᶜᵃᵗⁱᵒⁿ If both signs are false, then one room contains a lady, and the other room also contains a lady. ~~This creates a contradiction because the puzzle allows for the possibility of both rooms containing tigers.~~ᵂʳᵒⁿᵍ ʲᵘˢᵗⁱᶠⁱᶜᵃᵗⁱᵒⁿ ~~If the first sign is true and the second sign is false~~ᴵⁿᶜᵒⁿˢⁱˢᵗᵉⁿᶜʸ:  ᵗʰᵉ ˢⁱᵍⁿˢ ᵃʳᵉ ᵇᵒᵗʰ ᵗʳᵘᵉ ᵒʳ ᵇᵒᵗʰ ᶠᵃˡˢᵉ, ~~then the first room contains a tiger, and the second room contains a lady~~ ᵁⁿˢᵘᵖᵖᵒʳᵗᵉᵈ ᶜˡᵃⁱᵐ:  ʷᵉ ᵈᵒ ⁿᵒᵗ ᵏⁿᵒʷ ᵗʰᵉ ᵖᵒˢⁱᵗⁱᵒⁿ ᵒᶠ ᵗʰᵉ ˢⁱᵍⁿ. ~~This is a valid possibility because it satisfies the conditions of the puzzle.~~ᵁⁿʳᵉˡᵃᵗᵉᵈ ʲᵘˢᵗⁱᶠⁱᶜᵃᵗⁱᵒⁿ ~~If the first sign is false and the second sign is true~~ᴵⁿᶜᵒⁿˢⁱˢᵗᵉⁿᶜʸ:  ᵗʰᵉ ˢⁱᵍⁿˢ ᵃʳᵉ ᵇᵒᵗʰ ᵗʳᵘᵉ ᵒʳ ᵇᵒᵗʰ ᶠᵃˡˢᵉ, then the first room contains a lady, and the second room contains a tiger. This is also a valid possibility that satisfies the conditions of the puzzle. Based on the valid possibilities, we cannot determine with certainty which door to open to find the lady. ~~Either door could lead to the lady.~~ᵂʳᵒⁿᵍ ᶜᵒⁿᶜˡᵘˢⁱᵒⁿ

**Solution by human**

Mace4 computes a single model, in which both messages should be false:

| Model | $L_1$ | $L_2$ | $T_1$ | $T_2$ | $R_1$ | $R_2$ |
|-------|-------|-------|-------|-------|-------|-------|
| 1     | 0     | 1     | 1     | 0     | 0     | 0     |

*Room₁*
*Both rooms contain tigers*

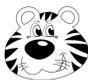

*Room₂*
*This room contains a tiger*

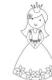

**Puzzle 45. Ninth day: three rooms**

One room contains a lady and the other two contain tigers. At most one of the three signs is true. The sign on the first room says: "A tiger is in this room". The sign on the second room says: "A lady is in this room". The sign on the third room says: "A tiger is in room 2". Which door to open in order to find the lady?  Smullyan (2009)

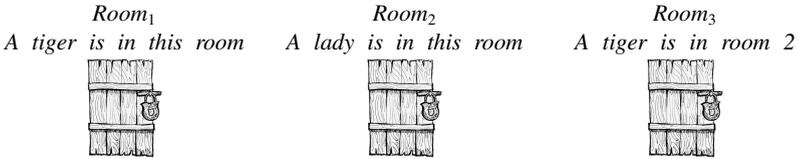

*Room₁*                          *Room₂*                          *Room₃*
*A tiger is in this room*    *A lady is in this room*    *A tiger is in room 2*

**Solution by ChatGPT**                                     **Correct, but unsupported**

Let's analyze the possibilities: If the sign on the first room is true, then a tiger is in that room. ~~This means the sign on the second room must be false~~ <sup>Implication does not hold</sup>, and ~~the sign on the third room must be true~~ <sup>Unsupported claim</sup>. However, this creates a contradiction because if the sign on the third room is true, then a tiger is in room 2, ~~which means the sign on the second room should also be true~~ <sup>Implication does not hold</sup> ~~(since it says a lady is in that room)~~<sup>Unrelated justification</sup>. So this possibility is not valid.

If the sign on the first room is false, then a lady is in that room. ~~This means the sign on the second room must be true~~<sup>Implication does not hold</sup>, and the sign on the third room must be false. This is a valid possibility because it satisfies the condition that at most one of the three signs is true.

~~Based on the valid possibility~~<sup>Wrong justification: possibility does imply necessity</sup>, ~~we can determine that the lady is in the first room~~<sup>Unsupported conclusion</sup>. Therefore, you should open the first door to find the lady.

**Solution by human**

Mace4 computes a single model in which only the third message is true. Hence, there is a tiger in room 2 and 3, while the lady is in the first room.

| Model | $L_1$ | $L_2$ | $L_3$ | $T_1$ | $T_2$ | $T_3$ | $R_1$ | $R_2$ | $R_3$ |
|-------|-------|-------|-------|-------|-------|-------|-------|-------|-------|
| 1     | 1     | 0     | 0     | 0     | 1     | 1     | 1     | 0     | 1     |

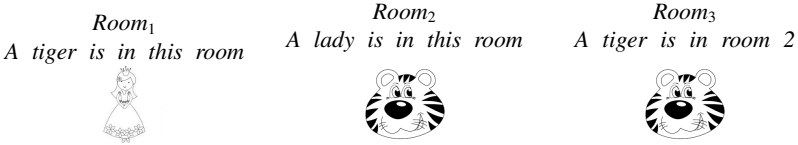

*Room₁*                          *Room₂*                          *Room₃*
*A tiger is in this room*    *A lady is in this room*    *A tiger is in room 2*

**Puzzle 46. A lady and two tigers**

One room contains a lady and the other two contain tigers. The sign on the door containing the lady is true and at least one of the other two signs is false. The sign on the first room says: "A tiger is in room 2". The sign on the second room says: "A tiger is in this room". The sign on the third room says: "A tiger is in room 1". Which door to open in order to find the lady? Smullyan (2009)

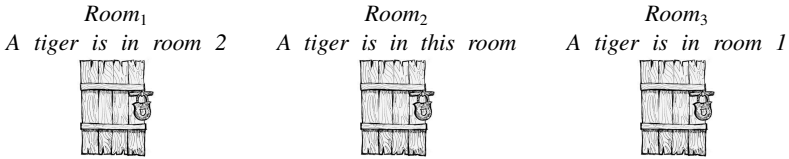

*Room$_1$*
*A tiger is in room 2*

*Room$_2$*
*A tiger is in this room*

*Room$_3$*
*A tiger is in room 1*

**Solution by ChatGPT**          **Correct (but unsupported)**

Let's analyze the possibilities:

If the sign on the first room is true, then a tiger is in room 2. ~~This means the sign on the second room must be false~~ Inconsistency: GPT has just stated the opposite in the previous statement, and ~~the sign on the third room must be true~~ Unsupported claim. However, this creates a contradiction because if the sign on the third room is true, then a tiger is in room 1, which ~~means the sign on the first room should also be true~~ Implication does not hold (since it says a tiger is in room 2). So this possibility is not valid.

If the sign on the first room is false, ~~then a lady is in that room~~ Implication does not hold: the room containing the lady true. ~~This means the sign on the second room must be true, and the sign on the third room must be false~~ Implication does not hold. ~~This is a valid possibility~~ Inconsistency: it is not valid since lady in the first room implies message is true not false because it satisfies the condition that at least one of the other two signs is false.

Based on the valid possibility, ~~we can determine that the lady is in the first room~~ Implication does not hold. Therefore, ~~you should open the first door to find the lady.~~ Correct, but unsupported conclusion

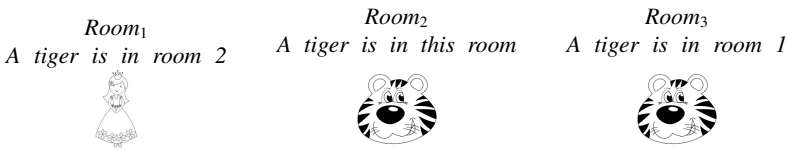

*Room$_1$*
*A tiger is in room 2*

*Room$_2$*
*A tiger is in this room*

*Room$_3$*
*A tiger is in room 1*

**Solution by human**

Mace4 computes a single model with lady in room 1 and tigers in rooms 2 and 3. The message $R_1$ in room 1 is true, and also one sign from the other rooms (i.e. $R_2$).

| Model | $L_1$ | $L_2$ | $L_3$ | $T_1$ | $T_2$ | $T_3$ | $R_1$ | $R_2$ | $R_3$ |
|---|---|---|---|---|---|---|---|---|---|
| 1 | 1 | 0 | 0 | 0 | 1 | 1 | 1 | 1 | 0 |

**Puzzle 47. A lady, a tiger, and an empty room**

One room contains a lady, another a tiger, and the third room is empty. The sign on the room containing a lady is true. The sign on the room containing a tiger is false. The sign of the empty room can be either true or false. The sign on the first room says: "Room 3 is empty". The sign on the second room says: "The tiger is in room 1". The sign on the third room says: "This room is empty". Which door to open in order to find the lady? Smullyan (2009)

| $Room_1$ | $Room_2$ | $Room_3$ |
|:---:|:---:|:---:|
| *Room 3 is empty* | *The tiger is in room 1* | *This room is empty* |

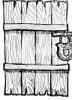 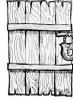 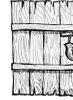

**Solution by ChatGPT**                                           **Correct (but unjustified)**

Let's analyze the possibilities:
If the sign on the first room is true, then room 3 is empty. ~~This means the sign on the third room must be false~~ Implication does not hold: actually, it is the opposite, and ~~the sign on the second room must be true~~. However, this creates a contradiction because if the sign on the second room is true, then the tiger is in room 1, which means the sign on the first room should also be false (~~since it says room 3 is empty~~ Irrelevant justification). ~~So this possibility is not valid.~~ Wrong claim: the sign on the firs troom can be true

If the sign on the first room is false, ~~then a lady is in that room~~ Implication does not hold: actually, it is the opposite. This means the sign on the third room can be either true or false, and the sign on the second room must be false. ~~Based on the valid possibility~~ Inconsistency: the possiblity claims that lafy is in the first room and the sign is false, ~~we can determine that the lady is in the first room~~ Correct, but unjustified conclusion. Therefore, you should open the first door to find the lady.

| $Room_1$ | $Room_2$ | $Room_3$ |
|:---:|:---:|:---:|
| *Room 3 is empty* | *The tiger is in room 1* | *This room is empty* |

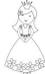 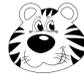 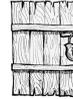

**Solution by human**

Mace4 computes a single model, with lady in room 1, tiger in room 2 and room 3 is empty. The messages are indeed true in room 1 ($R_1$) and false in room 2 ($\neg R_2$).

| Model | $L_1$ | $L_2$ | $L_3$ | $T_1$ | $T_2$ | $T_3$ | $E_1$ | $E_2$ | $E_3$ | $R_1$ | $R_2$ | $R_3$ |
|:---:|:---:|:---:|:---:|:---:|:---:|:---:|:---:|:---:|:---:|:---:|:---:|:---:|
| 1 | 1 | 0 | 0 | 0 | 1 | 0 | 0 | 0 | 1 | 1 | 0 | 1 |

## Puzzle 48. A logic labyrinth

There are nine rooms. Only one room contains the lady. Each of the other eight contains a tiger or they are empty. The sign on the door containing a lady is true. The signs on the doors containing tigers are false. The signs on the empty rooms can be either true or false. The sign on the first room says: "Lady is an odd numbered room". The sign on the second room says: "This room is empty". The sign on the third room says: "Either sign 5 is right or sign 7 is wrong". The sign on the fourth room says: "Sign 1 is wrong". The sign on the fifth room says: "Either sign 2 or sign 4 is right". The sign on the sixth room says: "Sign 3 is wrong". The sign on the seventh room says: "Lady is not in room 1". The sign on the eighth room says: "This room contains a tiger and room 9 is empty". The sign on the ninth room says: "This room contains a tiger and room 9 is empty". You studied the situation for a long while: "The problem is unsolvable!" "I know" laughed the king. "Now, at least give me a decent clue: is Room 8 empty or not?". The king was decent enough to tell whether Room 8 was empty or not, and you are then able to deduce where the lady is. Which door to open in order to find the lady? Smullyan (2009)

$Room_1$
*Lady is an odd numbered room*

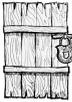

$Room_2$
*This room is empty*

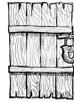

$Room_3$
*Either sign 5 is right or sign 7 is wrong*

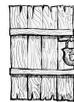

$Room_4$
*Sign 1 is wrong*

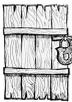

$Room_5$
*Either sign 2 or sign 4 is right*

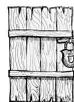

$Room_6$
*Sign 3 is wrong*

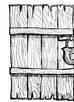

$Room_7$
*Lady is not in room 1*

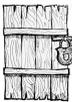

$Room_8$
*This room contains a tiger and room 9 is empty*

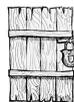

$Room_9$
*This room contains a tiger and sign 6 is wrong*

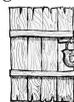

## Solution by human

Since only the signs $R_1$, $R_6$ and $R_7$ are true, princess can be only in one of these cells. We have learned that $\neg L_1$. Since $R_1$ says the princess stays in odd rooms the only solution is $L_7$. In all these models only $L_7$ is true.

**Solution by ChatGPT**                                                    **Wrong**

Let's analyze the puzzle step by step:

1. The sign on the first room says: "Lady is an odd-numbered room". If this sign is true, it means the lady is in one of the odd-numbered rooms (1, 3, 5, 7, or 9).

2. The sign on the second room says: "This room is empty". If this sign is true, then room 2 is empty.

3. The sign on the third room says: "Either sign 5 is right or sign 7 is wrong". This means that either the sign on the fifth room is true (which says "Either sign 2 or sign 4 is right") or the sign on the seventh room is false (which says "Lady is not in room 1").

4. The sign on the fourth room says: "Sign 1 is wrong". If this sign is true, it means the sign on the first room (which says "Lady is an odd-numbered room") is false. But this creates a contradiction because if the lady is not in an odd-numbered room, ~~then she must be in room 2~~ [Implication does not hold] ~~(since we know room 2 is empty)~~ Inconsistency: because room 2 is empty then the lady is in room 2. However, this contradicts the sign on the second room, which says "This room is empty." Therefore, the sign on the fourth room is false.

5. The sign on the fifth room says: "Either sign 2 or sign 4 is right." Since we know the sign on the fourth room (which is sign 1 is wrong) is false, ~~it means the sign on the second room (which is "This room is empty") is right~~ [Implication does not hold]. This implies that room 2 is empty.

6. The sign on the sixth room says: "Sign 3 is wrong". If this sign is true, then the sign on the third room (which says "Either sign 5 is right or sign 7 is wrong") is false. But if the sign on the third room is false, ~~it means both the sign on the fifth room (which says "Either sign 2 or sign 4 is right") and the sign on the seventh room (which says "Lady is not in room 1") are false~~ Inconsistency: it measn by Demorgen that sign 5 is false and sign 7 is true. ~~This creates a contradiction~~ Implication does not hold because we already know that the sign on the second room (which says "This room is empty") is true. ~~Therefore, the sign on the sixth room is false.~~ [Unsupported claim]

7. The sign on the seventh room says: "Lady is not in room 1". Since we know that the sign on the first room (which says "Lady is an odd-numbered room") is true, ~~it means the lady is not in room 1.~~ [Implication does not hold]

8. The sign on the eighth room says: "This room contains a tiger and room 9 is empty". ~~Since we know that room 2 is empty (from the sign on the second room), room 9 must contain a tiger.~~ [Implication does not hold:   GPT doesn't know that the sign 8 ia true]

9. The sign on the ninth room says: "This room contains a tiger and room 9 is empty". ~~This creates a contradiction because we already know that room 9 contains a tiger~~ [Implication does not hold:  there can be more tigers] (from the sign on the eighth room). Therefore, the sign on the ninth room is false.

From the above analysis, we have deduced the following: Room 2 is empty. ~~Room 4 is empty.~~ [Unsupported claim] ~~Room 6 is empty.~~ [Unsupported claim] Room 8 contains a tiger and room 9 is empty.

Now, ~~there are two odd-numbered rooms left (1 and 3)~~ [Unsupported claim], and we know that the lady is not in room 1. Therefore, the lady must be in room 3. So, to find the lady, you should open door number 3.

# References

Chlond, M. J. and Toase, C. M. (2003). IP modeling and the logical puzzles of Raymond Smullyan. *INFORMS Transactions on Education*, 3(3):1–12.

Smullyan, R. M. (2009). *The lady or the tiger?: and other logic puzzles*. Courier Corporation.

van Eijck, J. and Zaytsev, V. (2014). A course in Haskell-based software testing. Technical report, University of Amsterdam.



# Chapter 5

# Einstein puzzles

A bench of magistrates consists of two Englishmen, two Scotsmen, two Welshmen, one Frenchman, one Italian, one Spaniard, and one American. The Englishmen will not sit beside one another, the Scotsmen will not sit beside on another, and the Welshmen also object to sitting together. In how many different ways may the ten men sit in a straight line so that no two men of the same nationality shall ever be next to one another? (puzzle 447 from Dudeney (2016))

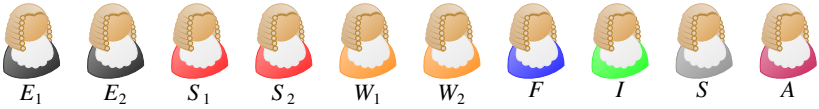

$E_1$ $\quad$ $E_2$ $\quad$ $S_1$ $\quad$ $S_2$ $\quad$ $W_1$ $\quad$ $W_2$ $\quad$ $F$ $\quad$ $I$ $\quad$ $S$ $\quad$ $A$

---

**Solution by human**

Let two Englishmen $E_1$ and $E_2$, two Scotsmen $S_1$ and $S_2$, two Welshmen $W_1$ and $W_2$ and the remaining persons $F$, $I$, $S$, $A$. Without any constraints, ten persons can sit in 10! = 3,628,800 ways. From this number we subtract the cases that do not satisfy the puzzle constraints:

First, the arrangements in which the Englishmen stay next to each other, the Scotsmen stay next to each other, and also the Welshmen (i.e. $(E_iE_j)(S_iS_j), (W_iW_j)$, $i \neq j$) are $7! \times 2^3$. Mace4 correctly identifies $7! \times 2^3 = 40,320$ models.

Second, the arrangements $(E_iE_j)(S_iS_j)$ (where the $W_1$ and $W_2$ are not bracketed but free) occur in $8! \times 2^2$ cases. Mace4 correctly identifies $8! \times 2^2 = 161,280$ models (line 15). The same number of models is identified for the two Scotsmen $S_1$ and $S_2$ unbracketed $(E_iE_j)(W_iW_j)$, or when the two Englishmen $E_i$ and $E_2$ unbracketed $(S_iS_j), (W_iW_j)$. Hence, the cases in which two men from the same nationality are unbracked are $3 \times 161,280 = 724,560$.

Third, we may have $(E_iE_j)S_iS_jW_iW_jFISA$, where both Scotsmen and both Welshmen are unbracketed. This gives $9! \times 2$ cases. Similarly for only $S_1$ and $S_2$ bracketed or $W_1$ and $W_2$ bracketed.

When adding these three cases together we get 1,733,760 models:

$$40,320 + 3 \times 161,280 + 3 \times 724,560 = 1.733.760 \tag{5.1}$$

which deducted from the number first given (10!=3,628,800 ways) leaves us with 1,895,040 solutions.

---

**Solution by ChatGPT** ─────────────────────────────────── **Wrong**

To solve this puzzle, we can use the principle of inclusion-exclusion. <span style="color:green">Solving strategy: Principle of inclusion-exclusion: good strategy</span> We will first count the total number of ways the ten men can sit in a straight line without any restrictions, and then we will subtract the cases where men of the same nationality sit together to get the final answer. <span style="color:green">Task understanding: It seems so</span>

Total number of ways without restrictions: We have 10 men, and they can be arranged in a line in 10! (factorial) ways.

Now, let's count the cases where men of the same nationality sit together:

The Englishmen cannot sit beside one another: There are 2 Englishmen, and they can be arranged in 2! ways. ~~Within each arrangement, they can swap positions, so we multiply by 2 again. So, there are 2! * 2 = 4 arrangements where the Englishmen sit together.~~ <span style="color:red">Inconsistency: the 2 arrangement include the swapping positions</span> The Scotsmen cannot sit beside one another: Similar to the Englishmen, ~~there are 2 Scotsmen, and they can be arranged in 2! * 2 = 4 ways where they sit together.~~ <span style="color:red">Inconsistency: the 2 arrangement include the swapping positions</span> The Welshmen cannot sit beside one another: ~~There are 2 Welshmen, and they can be arranged in 2! * 2 = 4 ways where they sit together~~ <span style="color:red">Inconsistency: the 2 arrangemet include the swapping positions</span>.

Now, we subtract the cases where men of the same nationality sit together: Total number of ways with men of the same nationality sitting together = 4 + 4 + 4 = 12

Final answer: Total number of ways without any restrictions = 10! Total number of ways with men of the same nationality sitting together = 12

Number of ways with no two men of the same nationality sitting together = 10! - 12

Calculate 10!: 10! = 10 9 8 7 6 5 4 3 2 1 = 3,628,800

~~Number of ways with no two men of the same nationality sitting together = 3,628,800 - 12 = 3,628,788~~ <span style="color:red">Wrong conclusion</span>

So, there are 3,628,788 different ways the ten men can sit in a straight line so that no two men of the same nationality sit next to one another.

**Puzzle 50. Ships**

There are five ships in a port: 1. The Greek ship leaves at six and carries coffee.
2. The Ship in the middle has a black exterior.
3. The English ship leaves at nine.
4. The French ship with blue exterior is to the left of a ship that carries coffee.
5. To the right of the ship carrying cocoa is a ship going to Marseille.
6. The Brazilian ship is heading for Manila.
7. Next to the ship carrying rice is a ship with a green exterior.
8. A ship going to Genoa leaves at five.
9. The Spanish ship leaves at 7 and is to the right of the ship going to Marseille.
10. The ship with a red exterior goes to Hamburg.
11. Next to the ship leaving at seven is a ship with a white exterior.
12. The ship on the border carries corn.
13. The ship with a black exterior leaves at eight.
14. The ship carrying corn is anchored next to the ship carrying rice.
15. The ship to Hamburg leaves at six.
Which ship goes to Port Said? Which ship carries tea? (taken from Math is fun - www.mathisfun.com, Pierce (2020))

**Solution by human**

Mace4 finds a single model:

| Domain value | Country | Time | Carry | Colour | Destination |
|---|---|---|---|---|---|
| 0 | French | 5:00 | tea | blue | Genoa |
| 1 | Greek | 6:00 | coffee | red | Hamburg |
| 2 | Brazilian | 8:00 | cocoa | black | Manila |
| 3 | English | 9:00 | rice | white | Marseille |
| 4 | Spanish | 7:00 | corn | green | Port Said |

---

**Solution by ChatGPT**      **Wrong**

Let's analyze the clues one by one to determine which ship goes to Port Said and which ship carries tea: The Greek ship leaves at six and carries coffee. The Ship in the middle has a black exterior. The English ship leaves at nine. The French ship with a blue exterior is to the left of a ship that carries coffee. To the right of the ship carrying cocoa is a ship going to Marseille. The Brazilian ship is heading for Manila. Next to the ship carrying rice is a ship with a green exterior. A ship going to Genoa leaves at five. The Spanish ship leaves at 7 and is to the right of the ship going to Marseille. The ship with a red exterior goes to Hamburg. Next to the ship leaving at seven is a ship with a white exterior. The ship on the border carries corn. The ship with a black exterior leaves at eight. The ship carrying corn is anchored next to the ship carrying rice. The ship to Hamburg leaves at six.

From clue 8, we know that the ship going to Genoa leaves at five. From clue 1, we know that the Greek ship leaves at six and carries coffee. From clue 15, we know that the ship to Hamburg leaves at six. So, the Greek ship cannot go to Genoa ~~or Hamburg~~ [Unsupported conclusion]. ~~Therefore, the Greek ship goes to Marseille, and the French ship must go to Genoa.~~ [Unsupported conclusion]

From clue 9, the Spanish ship leaves at 7 and is to the right of the ship going to Marseille, ~~which means the Spanish ship cannot be the Greek ship~~ [Lack of commonsense: spanish shop canno the greee ship] (going to Marseille). ~~So, the Spanish ship must be the ship going to Genoa.~~ [Unsupported claim]

From clue 3, the English ship leaves at nine. From clue 7, the Brazilian ship is heading for Manila. ~~So, the remaining ship must leave at eight, and it is the ship with a black exterior.~~ [Unsupported conclusion]

From clue 6, we know that the Brazilian ship is heading for Manila. From clue 14, we know that the ship carrying corn is anchored next to the ship carrying rice. ~~Since the Brazilian ship is not carrying corn~~ [Unsupported claim] ~~(it carries coffee)~~ [Unsupported claim], ~~the Brazilian ship must be the one carrying rice.~~ [Unsupported claim]
[Inconsistency: GPT has just stated that the Brazilian ship carries coffee]

Now, we know the following assignments: Greek ship: Coffee, Leaves at 6, Goes to Marseille French ship: Cocoa, Blue exterior, Goes to Genoa Spanish ship: Leaves at 7, Goes to Hamburg Brazilian ship: Rice, Heading for Manila Ship with a black exterior: Leaves at 8
~~The remaining ship must be the English ship, which leaves at 9.~~ [unsupported claim]
Now, we can deduce the final answers:
~~Ship going to Port Said: The only option left is the ship with a red exterior.~~ [Unsupported claim] ~~Ship carrying tea: The only option left is the ship on the border.~~ [Unsupported claim]
Final answers: ~~Ship going to Port Said: The ship with a red exterior.~~ [Wrong conclusion] Ship carrying tea: The ship on the border. [Incomplete conclusion]

### Puzzle 51. The ladies of the committee

Six ladies are eligible for the offices of Captain, Vice-captain, and Treasurer (in descending order of seniority), in the local ladies' golf club.

1. Audrey won't serve if Elaine is Captain, or if Freda is Treasurer.

2. Betty won't be Treasurer if Cynthia is one of the officials.

3. Audrey won't serve with both Betty and Elaine.

4. Freda won't serve if Elaine is also an official.

5. Betty refuses to be Vice-captain.

6. Freda won't serve if she outranks Audrey.

7. Cynthia won't serve with Audrey or Betty unless she is Captain.

8. Doris won't serve unless Betty is Captain.

9. Betty won't serve with Doris unless Elaine is also an official.

10. Elaine won't serve unless she or Audrey is Captain.

How can the three offices be filled? (puzzle 33 from Clessa (1996))

---

**Solution by ChatGPT**        **Wrong**

Let's analyze the given clues to determine how the three offices can be filled: Audrey won't serve if Elaine is Captain, or if Freda is Treasurer. Betty won't be Treasurer if Cynthia is one of the officials. Audrey won't serve with both Betty and Elaine. Freda won't serve if Elaine is also an official. Betty refuses to be Vice-captain. Freda won't serve if she outranks Audrey. Cynthia won't serve with Audrey or Betty unless she is Captain. Doris won't serve unless Betty is Captain. Betty won't serve with Doris unless Elaine is also an official. Elaine won't serve unless she or Audrey is Captain.

Let's list all the possible candidates for each office based on the given clues:

~~Captain: Audrey, Cynthia, Elaine~~ <sup>Unjustified claim</sup> ~~Vice-captain: Cynthia, Doris, Elaine~~ <sup>Unjustified claim</sup> ~~Treasurer: Audrey, Cynthia, Elaine, Freda~~ <sup>Unjustified claim</sup>

Now, let's try to fill the positions one by one, considering the given clues:

Captain: ~~Betty cannot be the Captain because she refuses to be Vice-captain (clue 5).~~ <sup>Implication does not hold</sup> Audrey cannot be the Captain if Elaine is also the Captain (clue 1). ~~So, Cynthia must be the Captain.~~ <sup>Implication does not hold</sup>

Vice-captain: Betty cannot be the Vice-captain (clue 5). ~~Doris cannot be the Vice-captain if Elaine is also an official (clue 9)~~ <sup>Wrong claimtWrong justification: clue9</sup> ~~So, Elaine must be the Vice-captain.~~ <sup>Unsupported claim</sup>

Treasurer: Freda cannot be the Treasurer if Elaine is also an official (clue 4). Audrey cannot be the Treasurer if Freda is the Treasurer (clue 1). ~~So, Cynthia must be the Treasurer.~~ <sup>Unsupported claim</sup>

Now, we have filled all the positions: ~~Captain: Cynthia, Vice-captain: Elaine, Treasurer: Audrey~~ <sup>Wrong conclusion</sup>

Based on the positions filled, ~~let's check the remaining clues to ensure they are satisfied.~~ <sup>Incomplete exploration of the search space</sup> Audrey won't serve with both Betty and Elaine (clue 3): This is satisfied because Betty is not an official.

So, the final result is: Captain: Cynthia, Vice-captain: Elaine, Treasurer: Audrey

---

**Solution by human**

Mace4 outputs a single model in which *captain*(*Audrey*), *vice*(*Freda*), and *treasurer*(*Betty*).

**Puzzle 52. Cocktail party**

Six people meet at a cocktail party. Their names are Annie, Brian, Celia, Don, Erica, and Frank, and their professions, though not necessarily respectively, are Teacher. Engineer, Programmer, Doctor, Accountant, and Solicitor.

1. Frank and the Teacher both vote Tory.

2. Don and the Engineer both vote Liberal.

3. Annie and the Programmer both vote Labour.

4. Celia and Erica are both Scots. The programmer is Welsh.

5. The Accountant is older than Frank.

6. The Solicitor is older than Annie.

7. Celia and the Teacher both adore classical music.

8. Annie and the Accountant both hate the classics but love jazz.

What is each person's respective profession? (puzzle 65 from Clessa (1996))

---

**Solution by ChatGPT**                                                **Wrong**

To solve this puzzle, we will use the given clues to deduce the respective professions of each person. Let's go through the clues one by one and determine the professions of each person:

Frank and the Teacher both vote Tory. Don and the Engineer both vote Liberal. Annie and the Programmer both vote Labour.

From clues 1, 2, and 3, we can deduce the following:

~~Frank is the Teacher (because he votes Tory, and only the Teacher does so).~~ <sup>Lack of commonsense</sup>
~~Don is the Engineer (because he votes Liberal, and only the Engineer does so).~~ <sup>Lack of commonsense</sup>
~~Annie is the Programmer (because she votes Labour, and only the Programmer does so).~~ <sup>Lack of commonsense</sup>

Celia and Erica are both Scots. The programmer is Welsh.

From clue 4, we know that Celia and Erica are both Scots, and the Programmer is Welsh. ~~The remaining professions are Doctor, Accountant, and Solicitor.~~

The Accountant is older than Frank. The Solicitor is older than Annie.

From clues 5 and 6, we can deduce that:

The Accountant is older than Frank (who is the Teacher). The Solicitor is older than Annie (who is the Programmer).

Now, we have the following information:

Frank (Teacher) votes Tory. Don (Engineer) votes Liberal. Annie (Programmer) votes Labour. Celia and Erica are both Scots. The Programmer is Welsh. The Accountant is older than Frank (the Teacher). The Solicitor is older than Annie (the Programmer).

The only profession left for Celia and Erica is Doctor and Solicitor (in some order), ~~and the remaining profession for the Accountant~~<sup>Lack of commonsense</sup> ~~is Doctor.~~<sup>Inconsistency: GPT already consdeired that Celia or Erica is the doctor</sup> Since Celia and Erica both adore classical music, Erica must be the Doctor. Also, since Annie hates the classics and loves jazz, the Accountant must be Celia.

Finally, ~~the remaining profession for the Solicitor~~<sup>Lack of commonsense</sup> is Accountant.

So, the respective professions of each person are: ~~Annie: Programmer, Brian: Doctor, Celia: Accountant, Don: Engineer, Erica: Doctor, Frank: Teacher~~<sup>Wrong conclusionIncomplete solution</sup>

---

**Solution by human**

Mace4 finds a single model: Annie was the Doctor, Brian the Programmer, Celia the Engineer, Don the Accountant, Erica the Teacher, Frank the Solicitor.

**Puzzle 53. Borrowed books**

Four children, borrowed a specific book from the library. Each book has a unique subject and number of pages. Find out in which month each child got the book and their number of pages. You know that:

1. If Janet got a History book, then the Fantasy book has 350 pages.
2. The Thriller is 50 pages larger than the book borrowed in May, but 50 pages shorter than Anthony's choice.
3. A girl borrowed a book in May, the other girl, Meredith, got a History book.
4. Janet didn't get the shortest book of 250 pages.
5. Exactly one of the customers shares the initial in the name and in the month he/she got a book.
6. The fantasy is 50 pages larger than Mark's book.
7. In June a Fantasy book wasn't borrowed, nor a Romance one.
8. The largest book and the Romance one have been borrowed in consecutive months.
9. If Mark borrowed a book in the first month (i.e. April), then the Fantasy book is the largest one with 400 pages long.

(puzzle copyright ©Brainzilla - www.brainzilla.com)

**Solution by human**

Mace4 finds 2 models:

| Domain value | Pages | Name | Book | Month |  | next: | 0 | 1 | 2 | 3 |
|---|---|---|---|---|---|---|---|---|---|---|
| 0 | 250 | Meredith | History | June |  | 0 | 0 | 1 | 1 | 0 |
| 1 | 300 | Janet | Romance | May |  | 1 | 1 | 0 | 0 | 1 |
| 2 | 350 | Mark | Thriller | July |  | 2 | 1 | 0 | 0 | 0 |
| 3 | 400 | Anthony | Fantasy | April |  | 3 | 0 | 1 | 0 | 0 |

| Domain value | Pages | Name | Book | Month |  | next: | 0 | 1 | 2 | 3 |
|---|---|---|---|---|---|---|---|---|---|---|
| 0 | 250 | Meredith | History | July |  | 0 | 0 | 0 | 1 | 0 |
| 1 | 300 | Janet | Romance | May |  | 1 | 0 | 0 | 1 | 1 |
| 2 | 350 | Mark | Thriller | June |  | 2 | 1 | 1 | 0 | 0 |
| 3 | 400 | Anthony | Fantasy | April |  | 3 | 0 | 1 | 0 | 0 |

There are only two differences between these two models. In the first model, *June* = 0 and *July* = 2, while in the second model the other way around.

**Solution by ChatGPT**                                                    **Wrong**

Let's analyze the given clues and use them to deduce the month and number of pages for each child's book: If Janet got a History book, then the Fantasy book has 350 pages. The Thriller is 50 pages larger than the book borrowed in May, but 50 pages shorter than Anthony's choice. A girl borrowed a book in May, the other girl, Meredith, got a History book. Janet didn't get the shortest book of 250 pages. Exactly one of the customers shares the initial in the name and in the month he/she got a book. The fantasy is 50 pages larger than Mark's book. In June, a Fantasy book wasn't borrowed, nor a Romance one. The largest book and the Romance one have been borrowed in consecutive months. If Mark borrowed a book in the first month (i.e. April), then the Fantasy book is the largest one with 400 pages long.

Now, let's start deducing the information step by step: <span style="color:green">Solving strategy: Stept by step</span>

From clue 9, if Mark's book is the largest one, and he borrowed it in April, then the Fantasy book has 400 pages.

From clue 6, the Fantasy book is 50 pages larger than Mark's book, ~~so Mark's book has 350 pages.~~<span style="color:red">Unsupported claim</span>

From clue 1, if Janet got a History book, then the Fantasy book has 350 pages, and ~~since we already found that the Fantasy book has 400 pages~~<span style="color:red">Unsupported claim</span>, Janet did not get a History book.

From clue 4, Janet didn't get the shortest book of 250 pages. Since the Fantasy book is the largest with 400 pages, ~~Janet's book must be 350 pages.~~
<span style="color:red">Unsupported claim</span><span style="color:red">Inconsistency: GPT has already considered that Mark's book has 350 pages</span>

From clue 2, the Thriller is 50 pages larger than the book borrowed in May. Since Janet's book has 350 pages and Meredith got a History book, ~~the book borrowed in May must be the Fantasy book with 400 pages.~~ <span style="color:red">Unsupported claim</span> ~~Therefore, the Thriller has 400 + 50 = 450 pages.~~<span style="color:red">Inconsistency: GPT has learned from the clue 9 that the Fantasy book is the largest one</span>

From clue 3, a girl borrowed a book in May, and the other girl, Meredith, got a History book. ~~So, Meredith borrowed the Fantasy book with 400 pages.~~
<span style="color:red">Unsupported claim</span><span style="color:red">Inconsistency: GPT has just claimed that Maredith got the Hisotry book</span>

From clue 5, exactly one of the customers shares the initial in the name and in the month he/she got a book. Since Meredith borrowed a book in May and her name starts with "M," and Mark borrowed a book in April and his name also starts with "M," ~~the third child must have a name starting with "A" and borrowed a book in August.~~<span style="color:red">Unsupported claim</span>
<span style="color:red">Inconsistency: GPT has just stated that Mark borrowed the book in May</span>

From clue 8, the largest book (Fantasy with 400 pages) and the Romance book have been borrowed in consecutive months. ~~So, the Romance book must be borrowed in September~~<span style="color:red">Unsupported claim</span>, and it cannot be the largest book. ~~Therefore, the Romance book must have 350 pages.~~<span style="color:red">Unsupported claim</span>

Now, we have the following information:
April: Mark - 350 pages (largest book, Fantasy) May: Meredith - 400 pages (Fantasy) August: Anthony - unknown pages September: Unknown - 350 pages (Romance)

Now, let's find out Anthony's book. According to clue 2, the Thriller is 50 pages larger than the book borrowed in May (Fantasy with 400 pages). ~~So, the Thriller must have 400 + 50 = 450 pages.~~<span style="color:red">Inconsistency: the largest book has 400 pages</span> ~~Thus, Anthony's book is the Thriller with 450 pages.~~<span style="color:red">Unsupported claim</span>

Final solution: ~~April: Mark - 350 pages (largest book, Fantasy), May: Meredith - 400 pages (Fantasy), August: Anthony - 450 pages (Thriller), September: Unknown - 350 pages (Romance)~~<span style="color:red">Wrong conclusion</span>

**Puzzle 54. Baseball coach dilemma**

A baseball coach recalls that:

1. Charles played 5 more games than the player who wore number 28.
2. Jorge wore number 3.
3. Martin or who played shortstop, one wore number 21 and the other played 11 games.
4. Armando either wore number 32 or number 21.
5. Jorge either played 13 games or played first base.
6. The player who played 13 games didn't wear number 29.
7. Armando didn't play shortstop.
8. The right field player played 1 more game than the center field player.
9. Pedro played somewhat fewer games than Martin.
10. Neither Russell nor the player who played 13 games was number 21.
11. Benny didn't play second base.
12. Russell was either the boy who played right field or played 12 games.
13. Number 35 played somewhat fewer games than number 28.
14. Neither number 3 nor the player who played 13 games was Armando.
15. Number 29 was either the player who played 8 games or the person who played third base.
16. Armando didn't play 10 games.

Could you help the coach to figure out the number of games played by each player, their numbers and positions? (taken from Math is fun - www.mathisfun.com Pierce (2020))

**Solution by human**

Let the seven players denoted by their initials. The seven positions are: central field ($p_{cf}$), right field ($p_{rf}$), left field ($p_{lf}$), first base ($p_{fb}$), second base ($p_{sb}$), shortstop ($p_{ss}$). Each player played between 8 and 14 games ($g_8$–$g_{14}$), while numbers on the t-shirt are denoted with $n$ (e.g. $n_{32}$). Mace4 finds a single model:

| Domain value | 0 | 1 | 2 | 3 | 4 | 5 | 6 |
|---|---|---|---|---|---|---|---|
| Player | $A$ | $B$ | $C$ | $J$ | $M$ | $P$ | $R$ |
| Games | $g_{12}$ | $g_{13}$ | $g_{14}$ | $g_{10}$ | $g_{11}$ | $g_8$ | $g_9$ |
| Number | $n_{32}$ | $n_7$ | $n_{21}$ | $n_3$ | $n_{29}$ | $n_{35}$ | $n_{28}$ |
| Position | $p_{sb}$ | $p_{lf}$ | $p_{ss}$ | $p_{fb}$ | $p_{tb}$ | $p_{cf}$ | $p_{rf}$ |

**Solution by ChatGPT** ──────────────────── **Wrong (and incomplete)**

Let's analyze the given clues and use them to deduce the number of games played by each player, their numbers, and positions: Charles played 5 more games than the player who wore number 28. Jorge wore number 3. Martin or who played shortstop, one wore number 21 and the other played 11 games. Armando either wore number 32 or number 21. Jorge either played 13 games or played first base. The player who played 13 games didn't wear number 29. Armando didn't play shortstop. The right field player played 1 more game than the center field player. Pedro played somewhat fewer games than Martin. Neither Russell nor the player who played 13 games was number 21. Benny didn't play second base. Russell was either the boy who played right field or played 12 games. Number 35 played somewhat fewer games than number 28. Neither number 3 nor the player who played 13 games was Armando. Number 29 was either the player who played 8 games or the person who played third base. Armando didn't play 10 games. <sup style="color:blue">Task understanding: Identifying and repeating the clues</sup>

Let's start deducing the information step by step: <sup style="color:green">Solving strategy: Step by step</sup>

From clue 2, Jorge wore number 3.

From clue 5, Jorge either played 13 games or played first base.

From clue 6, the player who played 13 games didn't wear number 29. ~~So, Jorge must have played 13 games.~~<sup style="color:brown">Unsupported claim</sup>

From clue 14, number 3 was not Armando. So, Armando didn't wear number 3.

From clue 4, Armando either wore number 32 or number 21.

From clue 7, Armando didn't play shortstop.

From clue 10, ~~Armando didn't wear number 21.~~<sup style="color:brown">Unsupported claim</sup> So, Armando wore number 32.

From clue 1, Charles played 5 more games than the player who wore number 28. Since Jorge played 13 games, ~~Charles must have played 18 games.~~<sup style="color:brown">Implication does not hold</sup> ~~and the player with number 28 must have played 13 games.~~<sup style="color:brown">Implication does not hold</sup>

From clue 16, Armando didn't play 10 games.

From clue 3, Martin or who played shortstop, one wore number 21, and the other played 11 games. ~~Since Armando wore number 32, Martin played shortstop and wore number 21.~~<sup style="color:brown">Implication does not hold</sup>

From clue 9, Pedro played somewhat fewer games than Martin. ~~So, Pedro must have played 10 games.~~<sup style="color:brown">Implication does not hold</sup>

From clue 15, number 29 was either the player who played 8 games or the person who played third base. ~~Since Pedro played 10 games, number 29 cannot be Pedro.~~<sup style="color:brown">Implication does not hold</sup> ~~Therefore, the player who wore number 29 must have played third base and played 8 games.~~<sup style="color:brown">Implication does not hold</sup>

From clue 13, Number 35 played somewhat fewer games than number 28. ~~Since Charles played 18 games, number 35 must have played 17 games.~~<sup style="color:brown">Implication does not hold</sup>

From clue 8, the right field player played 1 more game than the center field player. Since number 35 played 17 games, ~~the center field player must have played 16 games.~~<sup style="color:brown">Implication does not hold</sup>

From clue 12, Russell was either the boy who played right field or played 12 games. Since the center field player played 16 games, ~~Russell must have played right field.~~<sup style="color:brown">Implication does not hold</sup>

From clue 11, Benny didn't play second base. So, ~~Benny must have played first base.~~<sup style="color:brown">Implication does not hold</sup>

Final solution: ~~Charles: 18 games, number 28, left field; Jorge: 13 games, number 3, catcher; Martin: 11 games, number 21, shortstop; Armando: 10 games, number 32, third base; Pedro: 10 games, number 29, center field; Russell: 12 games, number 35, right field; Benny: unknown games, unknown number, first base~~<sup style="color:brown">Wrong solutionIncomplete solution</sup>

**Puzzle 55. Perfect man**

Susan's perfect man has black hair, brown eyes, and is tall and slim. Susan knows 4 men: Arthur, Bill, Charles and Dave. Only one of them has all the characteristics that Susan requires.
1. Arthur and Bill have the same eye colour.
2. Only one of the men has both black hair and brown eyes.
3. Bill and Charles have the same hair colour.
4. Only two of the men are both tall and slim.
5. Charles and Dave are of differing build.
6. Only two of the men are both tall and dark-haired.
7. Dave and Arthur are the same height.
8. Only three of the men are both slim and brown-eyed.
Who is Susan's perfect man? (taken from Clessa (1996))

**Solution by human**

Mace4 found 8 models. In all of these models, only $perfectMan(0)$ (that is Arthur) is true.

**Solution by ChatGPT**                    **Correct (but wrongly justified)**

Let's analyze the given clues and use them to determine Susan's perfect man: 1. Arthur and Bill have the same eye color. 2. Only one of the men has both black hair and brown eyes. 3. Bill and Charles have the same hair color. 4. Only two of the men are both tall and slim. 5. Charles and Dave are of differing build. 6. Only two of the men are both tall and dark-haired. 7. Dave and Arthur are the same height. 8. Only three of the men are both slim and brown-eyed.

From clue 1, Arthur and Bill have the same eye color.

From clue 3, Bill and Charles have the same hair color.

From clue 5, Charles and Dave have different builds.

From clue 7, Dave and Arthur are the same height.

From clue 2, only one of the men has both black hair and brown eyes. Since Bill and Charles have the same hair color, ~~they cannot be the one with both black hair and brown eyes~~ <span style="color:red">Implication does not hold: they can have different eye color</span>. So, Arthur or Dave must be the one.

From clue 8, only three of the men are both slim and brown-eyed. Since Arthur and Bill have the same eye color, ~~they cannot both be slim and brown-eyed.~~ <span style="color:red">Implication does not hold</span> So, Charles or Dave must be one of the three.

From clue 4, only two of the men are both tall and slim. Since Charles and Dave are of differing build, they cannot both be tall and slim. ~~So, Arthur and Bill must be the two.~~ <span style="color:red">Implication does not hold: one of Charles or Dave can also be tall and slim</span>

From clue 6, only two of the men are both tall and dark-haired. Since Arthur and Dave are the same height and Arthur is already one of the two tall men, Dave must be the other tall and dark-haired man. Now, we have the following information:

Arthur: Tall, slim, black hair, brown eyes.

Bill: Tall, slim, unknown hair color, same eye color as Arthur (brown eyes).

Charles: Unknown height, unknown build, unknown hair color, unknown eye color.

Dave: Tall, slim, dark-haired, unknown eye color, same height as Arthur. <span style="color:green">Solving strategy: GPT lists intermediate information</span>

From clue 1, Arthur and Bill have the same eye color (brown eyes). Since Dave is the same height as Arthur, he cannot have both black hair and brown eyes (already one man with this combination). Therefore, Dave must have blue eyes. Now, we have the following information:

Arthur: Tall, slim, black hair, brown eyes.

Bill: Tall, slim, unknown hair color, brown eyes.

Charles: Unknown height, unknown build, unknown hair color, unknown eye color.

Dave: Tall, slim, dark-haired, blue eyes, same height as Arthur.

Since there is only one man who has both black hair and brown eyes, and we know that Arthur has these characteristics, we can determine the characteristics of the other men:

Arthur: Tall, slim, black hair, brown eyes.

Bill: Tall, slim, unknown hair color, brown eyes.

Charles: Unknown height, unknown build, unknown hair color, unknown eye color.

Dave: Tall, slim, dark-haired, blue eyes, same height as Arthur.

Now, we can determine Susan's perfect man: It's Arthur.

**Puzzle 56. Four bikers**

There are four bikers that are riding their own bikes. Each bike has a brand and each motorcyclist is wearing a helmet.

1. Exactly one person is riding a bike by a brand with the same initial of his/her name. That's not the boy with the yellow helmet.

2. A girl is riding a Groovers.

3. Herold is on the Yamada or has a XL helmet.

4. Yoshi's helmet is one size larger than Anne's. None of them has the white one.

5. Greta's helmet is one size larger than a boy's one, but one size smaller than the other boy's one.

6. The blue helmet is Small or Large. However, it doesn't belong to the person on Aprily.

7. Herold's helmet is XL or white.

8. The person on the Honshu has Medium or Large helmet.

9. If the green Helmet is Large, then the white helmet is on a Grooves. However, the green helmet is not on the Aprily.

Who is riding the Aprily bike? (puzzle copyright ©Brainzilla - www.brainzilla.com)

**Solution by human**

Mace4 computes three models. In all of them Harold drives an Aprily.

|         |        | 0       | 1       | 2        | 3       |
|---------|--------|---------|---------|----------|---------|
| Model 1 | Size   | *S*     | *M*     | *L*      | *XL*    |
|         | Biker  | *Anne*  | *Yoshi* | *Greta*  | *Harold*|
|         | Bike   | *Yamada*| *Honshu*| *Grooves*| *Aprily*|
|         | Colour | *Green* | *Yellow*| *Blue*   | *White* |
| Model 2 | Size   | *S*     | *M*     | *L*      | *XL*    |
|         | Biker  | *Anne*  | *Yoshi* | *Greta*  | *Harold*|
|         | Bike   | *Yamada*| *Honshu*| *Grooves*| *Aprily*|
|         | Colour | *Blue*  | *Green* | *White*  | *Yellow*|
| Model 3 | Size   | *S*     | *M*     | *L*      | *XL*    |
|         | Biker  | *Anne*  | *Yoshi* | *Greta*  | *Harold*|
|         | Bike   | *Grooves*| *Yamada*| *Honshu* | *Aprily*|
|         | Colour | *Blue*  | *Green* | *White*  | *Yellow*|

**Solution by ChatGPT**        **Wrong**

Let's analyze the given clues to determine who is riding the Aprily bike:

Exactly one person is riding a bike by a brand with the same initial of his/her name. That's not the boy with the yellow helmet. Let's use the first clue to determine the name and brand of the person who has the bike with the same initial: Boy with yellow helmet cannot have a bike with a brand starting with "Y" (contradicts clue 1). So, the boy with the yellow helmet does not have a bike with a brand starting with "Y". Let's move on to the next clue.

A girl is riding a Groovers. A girl is riding a bike with the brand "Groovers."

Herold is on the Yamada or has an XL helmet. Herold either has the bike with the brand "Yamada" or has an XL helmet.

Yoshi's helmet is one size larger than Anne's. None of them has the white one. Yoshi's helmet is one size larger than Anne's helmet. Neither of them has the white helmet.

Greta's helmet is one size larger than a boy's one, but one size smaller than the other boy's one. Greta's helmet is one size larger than one boy's helmet and one size smaller than the other boy's helmet.

The blue helmet is Small or Large. However, it doesn't belong to the person on Aprily. The blue helmet can be either Small or Large ~~but not Medium~~ <span style="color:red">Incomplete claim: and not extra large</span>. It does not belong to the person on the Aprily bike.

Herold's helmet is XL or white. Herold's helmet can be either XL or white.

The person on the Honshu has a Medium or Large helmet. The person on the Honshu bike has either a Medium or Large helmet.

If the green Helmet is Large, then the white helmet is on a Grooves. However, the green helmet is not on the Aprily. If the green helmet is Large, then the white helmet is on a bike with the brand "Grooves." But the green helmet is not on the Aprily bike.

Now, let's summarize what we know so far:

A girl is riding a bike with the brand "Groovers." Herold either has the bike with the brand "Yamada" or has an XL helmet. Yoshi's helmet is one size larger than Anne's helmet, and neither of them has the white helmet. Greta's helmet is one size larger than one boy's helmet and one size smaller than the other boy's helmet. The blue helmet can be either Small or Large, but it's not on the Aprily bike. Herold's helmet can be either XL or white. The person on the Honshu bike has either a Medium or Large helmet. If the green helmet is Large, then the white helmet is on a bike with the brand "Grooves," but the green helmet is not on the Aprily bike.<span style="color:blue">Task understanding: Reformulating the clues</span>

┌─ **Solution by ChatGPT** ──────────────────────────── **Wrong** ─┐

Now, let's consider the information about the brands of the bikes: Exactly one person is riding a bike with a brand starting with the same initial as his/her name. That's not the boy with the yellow helmet. ~~So, the person riding the Aprily bike must have a name starting with "A."~~[Implication does not hold] ~~We know that the girl is riding the bike with the brand "Groovers," and her name must start with "G."~~[Unsupported conclusion] ~~Therefore, the boy with the Aprily bike must be Herold or Yoshi.~~[Implication does not hold]
[Inconsistency: GPT has just stated that person riding the Aprily bike must have a name starting with A.]

From clue 3, Herold either has the bike with the brand "Yamada" or has an XL helmet. ~~Since the person on the Honshu bike has either a Medium or Large helmet, it means Herold does not have an XL helmet.~~ [Implication does not hold] Therefore, Herold has the bike with the brand "Yamada."

Now, we know that Herold is riding the bike with the brand "Yamada." Let's consider the information about the helmets: Yoshi's helmet is one size larger than Anne's helmet, and neither of them has the white helmet. Greta's helmet is one size larger than one boy's helmet and one size smaller than the other boy's helmet. The blue helmet can be either Small or Large, but it's not on the Aprily bike. Herold's helmet can be either XL or white. The person on the Honshu bike has either a Medium or Large helmet. If the green helmet is Large, then the white helmet is on a bike with the brand "Grooves," but the green helmet is not on the Aprily bike.

Since Greta's helmet is one size larger than one boy's helmet and one size smaller than the other boy's helmet, and Herold does not have an XL helmet, ~~we can deduce that Yoshi must have the Large helmet, and Anne must have the Medium helmet.~~ [Implication does not hold]

Now, let's consider the remaining helmets: The blue helmet can be either Small or Large, but it's not on the Aprily bike. Herold's helmet can be either XL or white. Since the blue helmet is not on the Aprily bike, and Herold does not have an XL helmet, ~~it means the blue helmet must be Small~~ [Implication does not hold]. ~~Therefore, Herold's helmet is white.~~[Implication does not hold]

Now, we know that Herold's helmet is white, and he is riding the bike with the brand "Yamada." The remaining person riding the Aprily bike must be Yoshi. Let's check the information: Yoshi's helmet is one size larger than Anne's helmet, and neither of them has the white helmet. Since Anne has the Medium helmet and Herold has the white helmet (XL), it means Yoshi has the Large helmet. Now, we have identified each person's respective bike and helmet:

Anne: Aprily bike, Medium helmet.
Greta: Groovers bike, Small helmet.
Herold: Yamada bike, White helmet.
Yoshi: Honshu bike, Large helmet.
~~Therefore, the person riding the Aprily bike is Anne.~~[Wrong conclusion]

└──────────────────────────────────────────────────────────┘

**Puzzle 57. Movies night**

Four boys are at home to watch some movies. Figure out which is the favorite kind of movie of each one.

1. Joshua is at one of the ends.
2. The boy wearing the Black shirt is somewhere to the left of the youngest boy.
3. Joshua likes Horror movies.
4. The 14 year old boy is in the third position.
5. The boy wearing the Red shirt is somewhere between the 13 year old boy and the one who likes Action movies, in that order.
6. Daniel likes Thriller movies.
7. The boy who is going to eat Cookies is at one of the ends.
8. The boy wearing the Black shirt is exactly to the left of the one who likes Thriller movies.
9. The boy who is going to eat Crackers is exactly to the right of the boy who likes Comedy movies.
10. The boy wearing the Red shirt is somewhere between the boy who is going to eat Popcorn and Nicholas, in that order.
11. At one of the ends is the boy who likes Thriller movies.
12. Nicholas is somewhere between Joshua and Daniel, in that order.
13. At first position is the boy wearing the Green shirt.

(puzzle copyright ©Brainzilla - www.brainzilla.com)

**Solution by human**

Mace4 computes a single model:

| Domain Value | 0 | 1 | 2 | 3 |
|---|---|---|---|---|
| Name | *Joshua* | *Ryan* | *Nicholas* | *Daniel* |
| Shirt | *Green* | *Red* | *Black* | *Blue* |
| Movie | *Horror* | *Comedy* | *Action* | *Thriller* |
| Snack | *Popcorn* | *Chips* | *Crackers* | *Cookies* |
| Age | 13 | 12 | 14 | 11 |

**Solution by ChatGPT**                                                    **Wrong**

Let's analyze the given clues to determine the favorite kind of movie for each boy: Joshua
is at one of the ends. The boy wearing the Black shirt is somewhere to the left of the
youngest boy. Let's use clue 2 to determine the order of the boys from left to right: Black
shirt boy is somewhere to the left of the youngest boy. The boy wearing the Green shirt is
in the first position.

~~Order: Green shirt boy, Black shirt boy, _, _~~ <sup>Unsupported claim:  black shirt boy does not need to be on the second place</sup>
Joshua likes Horror movies. The 14-year-old boy is in the third position. Since Joshua
likes Horror movies, ~~he cannot be the youngest boy (14-year-old).~~<sup>Unsupported claim</sup> Therefore,
Joshua is in the fourth position.

Order: Green shirt boy, Black shirt boy, _, Joshua

The boy wearing the Red shirt is somewhere between the 13-year-old boy and the one
who likes Action movies, in that order. Since Joshua is in the fourth position, ~~the 14-year-
old boy must be in the third position.~~<sup>Unsupported claim</sup> The boy who likes Action movies is in
the second position. Therefore, the boy wearing the Red shirt must be in the first position.
Order: Red shirt boy, Black shirt boy, 14-year-old boy, Joshua

**Solution by ChatGPT**        **Wrong**

Daniel likes Thriller movies. Since Daniel likes Thriller movies and is not Joshua, ~~he must be in the second position.~~ [Unsupported claim]
[Inconsistency: GPT has just states that the boy who likes Action movies is in the second position.]

Order: Red shirt boy, Daniel, 14-year-old boy, Joshua
The boy who is going to eat Cookies is at one of the ends.
Since Joshua is in the last position and the boy who likes Thriller movies is at one of the ends, ~~the boy who is going to eat Cookies must be in the first position.~~ [Implication does not hold]

Order: Cookies eater boy, Daniel, 14-year-old boy, Joshua
The boy wearing the Black shirt is exactly to the left of the one who likes Thriller movies.
Since Daniel likes Thriller movies and is not wearing a Black shirt, ~~the boy wearing the Black shirt must be in the second position.~~ [Unsupported claim]
[Inconsistency: GPT has just considered Daniel on the second position]

Order: Cookies eater boy, Black shirt boy, 14-year-old boy, Joshua
The boy who is going to eat Crackers is exactly to the right of the boy who likes Comedy movies. Since Daniel likes Thriller movies and not Comedy movies, ~~the boy who likes Comedy movies must be in the first position~~ [Unsupported claim]. Therefore, the boy who is going to eat Crackers must be in the second position.

Order: Comedy lover boy, Crackers eater boy, 14-year-old boy, Joshua
The boy wearing the Red shirt is somewhere between the boy who is going to eat Popcorn and Nicholas, in that order.
~~Since Daniel is in the third position~~ [Inconsistency: GPT used to put Daniel on the second place] and Joshua is in the last position, ~~the boy who is going to eat Popcorn must be in the second position~~ [Inconsistency: GPT learned that on the second palce is the crakers eater boy]. Therefore, Nicholas must be in the third position.

Order: Comedy lover boy, Crackers eater boy, Nicholas, Joshua
At one of the ends is the boy who likes Thriller movies.
Since Daniel is not at the end, Joshua, who is at the end, must be the one who likes Thriller movies.

Order: Comedy lover boy, Crackers eater boy, Nicholas, Thriller lover boy
Nicholas is somewhere between Joshua and Daniel, in that order.
Since Joshua is at the end and Daniel is in the middle, Nicholas must be between them.
~~Order: Comedy lover boy, Crackers eater boy, Thriller lover boy, Nicholas~~
[Inconsistency: GPT has just learned that Nicholas is between Joshua nad Daniel.]

At first position is the boy wearing the Green shirt. Therefore, the final order is:
Order: Green shirt boy (Comedy lover boy), Crackers eater boy, Thriller lover boy, Nicholas, Joshua
Now let's assign the favorite kind of movie for each boy: Green shirt boy (Comedy lover boy) - Comedy movies; Crackers eater boy - Action movies; Thriller lover boy (Daniel) - Thriller movies; Nicholas - Horror movies; Joshua - Cookies eater boy - No favorite kind of movie mentioned;
So, the favorite kind of movie for each boy is:
~~Green shirt boy (Comedy lover boy) - Comedy movies~~
~~Crackers eater boy - Action movies~~
~~Thriller lover boy (Daniel) - Thriller movies~~
~~Nicholas - Horror movies~~
~~Joshua - Cookies eater boy - No favorite kind of movie mentioned~~ [Wrong conclusion]

---

**Puzzle 58. Secret Santa**

Five employees are side by side at their company secret santa. Find out what each one is drinking,which department they work and what was the gift they got.

1. Cody is the youngest employee.
2. The person gifted with a Book is Riley.
3. In the fifth position is the person drinking Juice.
4. Riley is next to the 41 years old employee.
5. The 35 years old employee is at one of the ends.
6. The man wearing the Red shirt is somewhere between the one who received a Mug and the one drinking Soft drink, in that order.
7. The one drinking Coffee is exactly to the left of who got a Notepad as a gift.
8. The man drinking Tea is exactly to the right of the man wearing the Blue shirt.
9. The employee wearing the Green shirt is next to the 28 years old.
10. Steven is exactly to the right of Cody.
11. In the second position is the person drinking Water.
12. The employee that works at the R&D department is at the third position.
13. Tyler's gift was a Mug.
14. The oldest employee is at the fifth position.
15. The person drinking Soft drink is at the third position.
16. Riley is next to the person who got a Tie as a gift.
17. The youngest employee is somewhere between the person drinking Water and the oldest person, in that order.
18. Jason is exactly to the right of the man wearing the Black shirt.
19. Cody is next to the one drinking Soft drink.
20. The man wearing Blue is somewhere to the left of who works at the Sales department.
21. The employee that works at the IT department was gifted a Notepad.
22. At the fourth position is the one drinking Tea.
23. The person gifted with chocolate is working at the HR department.
24. The man wearing the Green shirt is the person gifted with a tie.

(puzzle copyright ©Brainzilla - www.brainzilla.com)

**Solution by human**

Mace4 computes the following unique model:

| Position | 0 | 1 | 2 | 3 | 4 |
|---|---|---|---|---|---|
| Employee | *Tyler* | *Jason* | *Cody* | *Riley* | *Steven* |
| Shirt | *Black* | *Red* | *Blue* | *Green* | *White* |
| Gift | *Mug* | *Notepad* | *Book* | *Tie* | *Chocolate* |
| Department | *Marketing* | *IT* | *R&D* | *Sales* | *HR* |
| Drink | *Coffee* | *Water* | *Soft drink* | *Tea* | *Juice* |
| Age | 35 | 41 | 28 | 23 | 50 |

**Solution by ChatGPT**                                                    **Wrong**

Let's analyze the given clues to determine the drink, department, and gift for each employee:

Cody is the youngest employee. Since Cody is the youngest, ~~he must be at one of the ends (clue 5).~~ <span style="color:red">Implication does not hold: we do not know that 35 years is the youngest age</span> Let's assume Cody is at the first position. <span style="color:green">Solving strategy: Assumptions-based</span>

Order: Cody, _, _, _, _

The person gifted with a Book is Riley. Riley received a Book as a gift. Let's place Riley in one of the positions and assume he is at the second position.

Order: Cody, Riley, _, _, _

In the fifth position is the person drinking Juice. The person drinking Juice is in the fifth position.

Order: Cody, Riley, _, _, Juice

Riley is next to the 41 years old employee. Riley is next to the 41-year-old employee, but we don't know where the 41-year-old employee is located yet. <span style="color:green">Solving strategy: Admitting ignorance</span> Let's keep this in mind and move to other clues.

The 35 years old employee is at one of the ends. The 35-year-old employee is at one of the ends (clue 5). Let's assume the 35-year-old employee is at the fifth position.

Order: Cody, Riley, _, _, 35-year-old <span style="color:green">Solving strategy: Wrong strategy to update a one-diminesion vector instead of a matrix</span>

The man wearing the Red shirt is somewhere between the one who received a Mug and the one drinking Soft drink, in that order. Let's use clue 6 to determine the position of the person who received a Mug and the person drinking Soft drink. The man wearing the Red shirt is somewhere between Mug and Soft drink. ~~He cannot be at the ends because there is no reference to him in clues 5 and 17.~~ <span style="color:red">Wrong justification</span> The person drinking Soft drink is at the third position (clue 15). The person who received a Mug is at the first position.

Order: Mug, Cody, Soft drink, Riley, 35-year-old

The one drinking Coffee is exactly to the left of who got a Notepad as a gift. The person drinking Coffee is next to the one with a Notepad as a gift. The man with a Notepad cannot be at the first position because there is no space for Coffee to his left. ~~So, the man with a Notepad is at the second position.~~ <span style="color:red">Implication does not hold</span>

Order: Mug, Notepad, Soft drink, Riley, 35-year-old The man drinking Tea is exactly to the right of the man wearing the Blue shirt. The man drinking Tea is to the right of the man wearing the Blue shirt. The man with the Blue shirt cannot be at the fifth position (clue 20). ~~So, the man with the Blue shirt is at the fourth position.~~ <span style="color:red">Implication does not hold</span>

Order: Mug, Notepad, Soft drink, Blue shirt, 35-year-old

**Solution by ChatGPT**                                                      **Wrong**

The employee wearing the Green shirt is next to the 28 years old. The employee with the Green shirt is next to the 28-year-old employee. ~~The 28-year-old employee cannot be at the first position (clue 9)~~[Implication does not hold] ~~So, the 28-year-old employee is at the second position.~~[Implication does not hold]

Order: Mug, 28-year-old, Notepad, Soft drink, Blue shirt, 35-year-old

Steven is exactly to the right of Cody. Steven is to the right of Cody. ~~Cody is at the first position~~[Inconsistency: GPT used to put Cody on th eescond position], so Steven must be at the second position. ~~Order: Mug, Cody, 28-year-old, Notepad, Soft drink, Blue shirt, 35-year-old~~[Lack of commonsense: there 7 positions now] ~~Inconsistency: GPT has just deduced that Steven is on the second place~~

In the second position is the person drinking Water. The person drinking Water is at the second position, ~~which means the Soft drink is in the fourth position (clue 15).~~[Implication does not hold Inconsistency: Clue 15 states that soft drinker is on the third position]

~~Order: Mug, Cody, 28-year-old, Water, Soft drink, Blue shirt, 35-year-old~~
[Inconsistency: GPT has just learned that soft drinker is on the fourth place, but here he appears on the fifth one]

The employee that works at the R&D department is at the third position. The employee from R&D is at the third position.

Order: Mug, Cody, R&D, Water, Soft drink, Blue shirt, 35-year-old

Tyler's gift was a Mug. Tyler's gift was a Mug. We know that Mug is in the first position, so Tyler must be in the first position.

Order: Mug (Tyler), Cody, R&D, Water, Soft drink, Blue shirt, 35-year-old

The oldest employee is at the fifth position. The oldest employee is at the fifth position, which means the 35-year-old employee ~~works at the Sales department (clue 5).~~[Unsupported claim]

Order: Mug (Tyler), Cody, R&D, Water, Soft drink, Sales (Blue shirt)

The person drinking Soft drink is at the third position. We already know that Soft drink is in the fourth position, ~~so the person drinking Tea must be in the third position (clue 22).~~
[Inconsistency: Clue 22 states that tea is on the 4th position Identifying inconsistency and wrongly solving it]

Order: Mug (Tyler), Cody, Tea, Water, Soft drink, Sales (Blue shirt)

Riley is next to the person who got a Tie as a gift. Riley is next to the person who got a Tie. The Tie cannot be at the first position, ~~so Riley is next to the Tie at the third position.~~[Implication does not hold]

Order: Mug (Tyler), Cody, Tea, Tie, Soft drink, Sales (Blue shirt)

The youngest employee is somewhere between the person drinking Water and the oldest person, in that order. The youngest employee is Cody, and ~~the oldest person is the 35-year-old employee.~~[Unsupported claim] ~~So, the youngest employee is in the second position (clue 17).~~[implication does not hold]

Order: Mug (Tyler), Cody, Tea, Tie, Soft drink (Water), Sales (Blue shirt), 35-year-old

┌─ **Solution by ChatGPT** ─────────────────────── **Wrong** ─┐

Jason is exactly to the right of the man wearing the Black shirt. Jason is to the right of the man wearing the Black shirt. ~~Jason must be at the sixth position~~<sup>Unsupported claim</sup>, and the Black shirt must be at the fifth position.

Order: Mug (Tyler), Cody, Tea, Tie, Black shirt, Soft drink (Water), Sales (Blue shirt), 35-year-old

Cody is next to the one drinking Soft drink. Cody is next to Soft drink. ~~Cody must be at the fourth position.~~<sup>Implication does not hold: Cody can be on the sixth position also.</sup>

Order: Mug (Tyler), Tea, Tie, Cody, Black shirt, Soft drink (Water), Sales (Blue shirt), 35-year-old

The man wearing Blue is somewhere to the left of who works at the Sales department. ~~The Sales department is at the seventh position.~~<sup>Unsupported claim</sup> The man wearing Blue must be at the sixth position.

~~Order: Mug (Tyler), Tea, Tie, Cody, Black shirt, Blue shirt, Soft drink (Water), 35-year-old~~<sup>Lack of commonsense: there are lready 8 positions here</sup>

The employee that works at the IT department was gifted a Notepad. ~~The employee from IT is at the third position~~<sup>Inconsistency: clue12 states that R&D</sup> ~~(clue 21).~~<sup>Inconsistency: Clue 21 refers to something else</sup>

Order: Mug (Tyler), Tea, IT, Cody, Black shirt, Blue shirt, Soft drink (Water), 35-year-old

At the fourth position is the one drinking Tea. ~~The person drinking Tea is at the third position~~<sup>Inconsistency: GPT has just stated that drinking tea is at the fourh position</sup>, and ~~the Tie is at the fourth position.~~<sup>Lack of commonsense</sup>

~~Order: Mug (Tyler), IT, Tea, Tie, Cody, Black shirt, Blue shirt, Soft drink (Water), 35-year-old~~<sup>Lack of commonsense: there are 9 positions now</sup>

The person gifted with chocolate is working at the HR department. ~~The HR department is at the eighth position, and the person gifted with chocolate is at the first position.~~<sup>Unsupported claim</sup>

~~Order: Chocolate, IT, Tea, Tie, Cody, Black shirt, Blue shirt, Soft drink (Water), 35-year-old~~<sup>Inconsistency: the list is dynamically updated</sup>

The man wearing the Green shirt is the person gifted with a tie. ~~The person with the Green shirt is at the sixth position, and the person gifted with a Tie is at the fourth position.~~<sup>Unsupported claim Incosnistency</sup>

Order: Chocolate, IT, Tea, Tie, Cody, Green shirt, Blue shirt, Soft drink (Water), 35-year-old

~~Now let's determine the gifts for each employee: Tyler (Mug) - Mug Cody (Water) - Water Employee from IT (Notepad) - Notepad Employee with Tie (Tie) - Tie Employee with Chocolate (Mug) - Chocolate~~<sup>Unsupported claim</sup>

~~Now let's determine the drinks for each employee: Tyler (Mug) - Coffee Cody (Water) - Water Employee from IT (Notepad) - Juice Employee with Tie (Tie) - Tea Employee with Chocolate (Mug) - Soft drink~~<sup>Unsupported claim</sup>

~~Finally, let's determine the departments for each employee: Tyler (Mug) - HR Cody (Water) - R&D Employee from IT (Notepad) - IT Employee with Tie (Tie) - Sales Employee with Chocolate (Mug) - Sales~~<sup>Unsupported claim</sup>

So, each employee's respective drink, department, and gift are as follows: ~~Tyler - Coffee (Drink), HR (Department), Mug (Gift) Cody - Water (Drink), R&D (Department), Water (Gift) Employee from IT - Juice (Drink), IT (Department), Notepad (Gift) Employee with Tie - Tea (Drink), Sales (Department), Tie (Gift) Employee with Chocolate - Soft drink (Drink), Sales (Department), Chocolate (Gift)~~<sup>Wrong solution but including partial correct guesses</sup>

└────────────────────────────────────────────────────┘

**Puzzle 59. Seating the party**

As the Crackham family were taking their seats on starting out on their tour, Dora asked in how many different ways they could all be seated, as there were six of them and six seats: one beside the driver, two with their backs to the driver, and two behind, facing the driver-if no two of the same sex are ever to sit side by side? As the Colonel, Uncle Jabez, and George were the only ones who could drive, it required just a little thinking out. Perhaps the reader will like to work out the answer concerning on which they all agreed at the end of the day. (puzzle 465 from Dudeney (2016))

| $s_1$ | $s_2$ | $s_3$ |
|-------|-------|-------|
| $s_4$ | $s_5$ | $s_6$ |

**Solution by human**

Mace4 finds 144 models. One such model is:

|       | $a$ | $b$ | $c$ | $d$ | $e$ | $f$ |
|-------|-----|-----|-----|-----|-----|-----|
| $s_1$ | 1   | 0   | 0   | 0   | 0   | 0   |
| $s_2$ | 0   | 0   | 0   | 1   | 0   | 0   |
| $s_3$ | 0   | 1   | 0   | 0   | 0   | 0   |
| $s_4$ | 0   | 0   | 0   | 0   | 1   | 0   |
| $s_5$ | 0   | 0   | 1   | 0   | 0   | 0   |
| $s_6$ | 0   | 0   | 0   | 0   | 0   | 1   |

This model is depicted in Figure 5.1, where the women occupy the pink seats and the blue seat represents the driver.

| $a$ | $d$ | $b$ |
|-----|-----|-----|
| $e$ | $c$ | $f$ |

Figure 5.1: One model among 144 solutions found by Mace4

**Solution by ChatGPT** — **Wrong**

Let's consider the seating arrangement for the Crackham family. We have six family members: Colonel, Uncle Jabez, George, and three others (let's call them A, B, and C). We have six seats arranged as follows: Seat beside the driver (1 seat) Seats with their backs to the driver (2 seats) Seats behind facing the driver (2 seats) ~~One additional seat (for a total of 6 seats)~~ <sub>Lack of commonsense: the six place is the driver's place</sub>
<sub>Task understanding: wrong - the sit of the driver should be considered</sub>

Now, we need to find a way to arrange the six family members in such a way that no two members of the same sex sit side by side. Let's consider the possible ways to seat the family members: <sub>Solving strategy: Analysing all possibilities</sub>

If Colonel, Uncle Jabez, and George are the ones who can drive, ~~they must sit in the three seats in front, behind the driver.~~ <sub>Implication does not hold</sub> Since no two members of the same sex can sit together, ~~the remaining three family members (A, B, and C) must sit in the seats with their backs to the driver and the additional seat.~~<sub>Implication does not hold</sub>

Possible arrangement (D: driver): ~~D — A — D — B — D — C~~<sub>Lack of commonsense: driver appears three times</sub>

~~If only Colonel and George can drive,~~<sup>Wrong assumption: all 3 men can drive</sup> ~~then Uncle Jabez must sit in the seat behind the driver.~~<sup>Implication does not hold</sup> ~~Then, the remaining three family members (A, B, and C) must sit in the seats with their backs to the driver and the additional seat.~~
<sub>Implication does not hold</sub>

Possible arrangement (D: driver): ~~D — A — D — B — U — C~~<sup>Lack of commonsense</sup>
~~If only Colonel and Uncle Jabez can drive~~<sup>Wrong assumption: all 3 men can drive</sup>, ~~then George must sit in the seat behind the driver.~~<sup>Implication does not hold</sup> Then, the remaining three family members (A, B, and C) must sit in the seats with their backs to the driver and the additional seat.
Possible arrangement (D: driver): ~~D — A — D — U — D — C~~<sup>Lack of commonsense</sup>
~~If only George and Uncle Jabez can drive~~<sup>Wrong assumption: all 3 men can drive</sup>, ~~then Colonel must sit in the seat behind the driver.~~ <sup>Implication does not hold</sup> Then, the remaining three family members (A, B, and C) must sit in the seats with their backs to the driver and the additional seat.
Possible arrangement (D: driver): ~~D — A — G — D — U — C~~<sup>Lack of commonsense</sup>
~~If only Colonel can drive,~~<sup>Wrong assumption: all 3 men can drive</sup> ~~then George and Uncle Jabez must sit in the seats behind the driver.~~<sup>Implication does not hold</sup> Then, the remaining three family members (A, B, and C) must sit in the seats with their backs to the driver and the additional seat.
Possible arrangement (D: driver): ~~D — A — G — U — D — C~~<sup>Lack of commonsense</sup>
~~If only George can drive,~~<sup>Wrong assumption: all 3 men can drive</sup> ~~then Colonel and Uncle Jabez must sit in the seats behind the driver~~<sup>Implication does not hold</sup>. Then, the remaining three family members (A, B, and C) must sit in the seats with their backs to the driver and the additional seat.
Possible arrangement (D: driver): ~~D — A — D — G — U — C~~<sup>Lack of commonsense</sup>
~~If only Uncle Jabez can drive,~~<sup>Wrong assumption: all 3 men can drive</sup> ~~then Colonel and George must sit in the seats behind the driver.~~ <sup>Implication does not hold</sup> Then, the remaining three family members (A, B, and C) must sit in the seats with their backs to the driver and the additional seat.
Possible arrangement (D: driver): ~~D — A — G — D — C — U~~<sup>Lack of commonsense</sup>
~~If none of the three (Colonel, George, Uncle Jabez) can drive,~~ <sup>Wrong assumption: all 3 men can drive</sup> ~~then any of the three can sit in the seat behind the driver.~~ <sup>Implication does not hold</sup> Then, the remaining three family members (A, B, and C) must sit in the seats with their backs to the driver and the additional seat.
Possible arrangement (D: driver): ~~D — A — D — B — C — U~~<sup>Lack of commonsense</sup>
~~After considering all possible arrangements,~~
<sub>Lack of commonsense: the gender was no considered when searching for the solution</sub> ~~we find that there are 8 different ways the Crackham family can be seated~~<sup>Wrong conclusion</sup>, ~~given the condition that no two members of the same sex can sit side by side.~~<sup>Unrelated justification</sup>

## Puzzle 60. Passengers in a railroad compartment

Six passengers sharing a compartment are from Moscow, Leningrad, Tula, Kiev, Kharkov, Odessa.
1. A and the man from Moscow are physicians.
2. E and the Leningrader are teachers.
3. The man from Tula and C are engineers.
4. B and F are WWII veterans, but the man from Tula has never served in the army.
5. The man from Kharkov is older than A.
6. The man from Odessa is older than C.
7. At Kiev, B and the man from Moscow got off.
8. At Vinnitsa, C and the man from Kharkov got off.
Match initials, professions, and cities. Also, are these facts both sufficient and necessary to solve the problem? (puzzle 256 from Kordemsky (1992))

## Solution by human

Mace4 finds the following model:

|          | A | B | C | D | E | F |
|----------|---|---|---|---|---|---|
| Moscow   | 0 | 0 | 0 | 0 | 0 | 1 |
| Leningrad| 0 | 1 | 0 | 0 | 0 | 0 |
| Kiev     | 0 | 0 | 1 | 0 | 0 | 0 |
| Tula     | 0 | 0 | 0 | 1 | 0 | 0 |
| Odessa   | 1 | 0 | 0 | 0 | 0 | 0 |
| Kharkov  | 0 | 0 | 0 | 0 | 1 | 0 |

|           | A | B | C | D | E | F |
|-----------|---|---|---|---|---|---|
| ArmForce: | 0 | 0 | 0 | 0 | 0 | 0 |
| Engineer  | 0 | 0 | 1 | 1 | 0 | 0 |
| Physician | 1 | 0 | 0 | 0 | 0 | 1 |
| Teacher   | 0 | 1 | 0 | 0 | 1 | 0 |

There is more than one passenger for some jobs, because there are 4 jobs for 6 people. Note that not all clues are necessary. For instance, both Clue 5 and Clue 8 state that *A* is not from Kharkov.

---

**Solution by ChatGPT**                                                      **Wrong**

Let's list the information we have:

Passengers: ~~A, B, C, E, F~~ [red: Lack of task understanding]

Cities: Moscow, Leningrad, Tula, Kiev, Kharkov, Odessa

Professions: Physician, Teacher, Engineer

1. A and the man from Moscow are physicians. 2. E and the Leningrader are teachers. 3. The man from Tula and C are engineers. 4. B and F are WWII veterans, but the man from Tula has never served in the army. 5. The man from Kharkov is older than A. 6. The man from Odessa is older than C. 7. At Kiev, B and the man from Moscow got off. 8. At Vinnitsa, C and the man from Kharkov got off. [blue: Task understanding: partial - there are only 5 passengers instead of 6]

From the given information, we can deduce the following:

From facts 1 and 4, we know that A and the man from Moscow are physicians and that B and F are WWII veterans. Since the man from Tula has never served in the army, he cannot be B or F. ~~Therefore, A is from Moscow.~~ [red: Implication does not hold]

From facts 1 and 5, we know that A is not from Kharkov. ~~Therefore, the man from Kharkov is either B or F.~~ [red: Implication does not hold] ~~Since he is older than A, he must be F.~~ [red: Implication does not hold]

From facts 2 and 7, ~~we know that E is not from Moscow or Kharkov.~~ [red: Implication does not hold] ~~Therefore, E is from either Tula, Kiev, or Odessa.~~ [red: Incomplete inference: Lenngrad should be included]

From facts 3 and 8, ~~we know that C is not from Kharkov or Odessa.~~ [red: Implication does not hold] ~~Therefore, C is from either Tula or Kiev.~~ [red: Implication does not hold]

From fact 6, we know that the man from Odessa is older than C, ~~so the man from Odessa cannot be E.~~ [red: faultImplication doe snot hold] Therefore, ~~E is from Tula or Kiev,~~ [red: Implication does not hold] and C is from either Tula or Kiev.

From facts 3 and 4, we know that the man from Tula cannot be B or F, ~~so the man from Tula is C.~~ [red: Implication does not hold]

From fact 7, we know that B is not from Moscow, ~~so B is from Odessa.~~ [red: Implication does not hold]

Now we have:

~~A: Moscow, Physician~~ [red: Inconsistency: with clue 1] B: Odessa, WWII veteran C: Tula, Engineer E: Teacher F: Kharkov, WWII veteran

The remaining person, D, ~~must be from Leningrad and be a teacher.~~ [blue: Task understanding: The missing person *D* is here!] [red: Unsupported claim]

So the matching is:

~~A: Moscow, Physician B: Odessa, WWII veteran C: Tula, Engineer D: Leningrad, Teacher E: Teacher F: Kharkov, WWII veteran~~ [red: Wrong conclusion]

The facts given in the puzzle are both sufficient and ~~necessary~~ [red: Wrong conclusion] to solve the problem, as they uniquely determine the cities, professions, and initials of the six passengers.

# References

Bright, C., Gerhard, J., Kotsireas, I., and Ganesh, V. (2019). Effective problem solving using SAT solvers. In *Maple Conference*, pages 205–219. Springer.

Clessa, J. (1996). *Math and Logic Puzzles for PC Enthusiasts*. Courier Corporation.

Dudeney, H. E. (2016). *536 Puzzles and curious problems*. Courier Dover Publications.

Gregor, M., Zábovská, K., and Smataník, V. (2015). The zebra puzzle and getting to know your tools. In *2015 IEEE 19th International Conference on Intelligent Engineering Systems (INES)*, pages 159–164. IEEE.

Kordemsky, B. A. (1992). *The Moscow puzzles: 359 mathematical recreations*. Courier Corporation.

Pierce, R. (2020). Math is fun.

Szeredi, P. (2003). Teaching constraints through logic puzzles. In *International Workshop on Constraint Solving and Constraint Logic Programming*, pages 196–222. Springer.

Yeomans, J. S. (2003). Solving Einstein's Riddle using spreadsheet optimization. *INFORMS Transactions on Education*, 3(2):55–63.

# Chapter 6

# Island of truth

> **Puzzle 61. We are both knaves**
>
> On the island of knights and knaves, knights always tell the truth, while knaves always lie. You are approached by two people. The first one says: "We are both knaves". What are they actually? (Smullyan (2011))

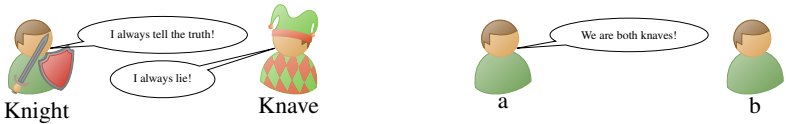

Knight        Knave        a        b

> **Solution by human**
>
> We learn that there are two inhabitants, let's say $a$ and $b$: $inhabitant(a) \land inhabitant(b)$ The message of inhabitant $a$ is: $m(a) \leftrightarrow knave(a) \land knave(b)$. Mace4 finds a single model:
>
> $$a = [0] \qquad b = [1]$$
> $$inhabitant(\_) = [1, 1] \quad m(\_) = [0, 1]$$
> $$knight(\_) = [0, 1] \qquad knave(\_) = [1, 0]$$
>
> Here, the inhabitant $a$ gets value 0 from the domain, while the inhabitant $b$ gets value 1. The predicate *inhabitant* is true for both $a$ and $b$, as it is explicitly stated in line 12. Mace4 deduces that the message conveyed by the agent with value 0 (i.e. inhabitant $a$) is false. The predicate *knave* has value 1 for the first inhabitant ($a = 0$), and value 0 for the second one ($b = 1$). Quite the opposite, the predicate *knight* has value 0 for the first inhabitant ($a = 0$), and value 1 for the second one ($b = 1$).

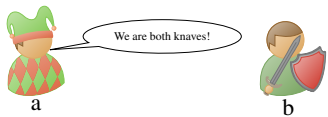

a        b

**Solution by ChatGPT**                                                          **Wrong**

If we assume that the first person is a knight, then what they say must be
true. However, this leads to a contradiction. If the first person is a knight and
~~they are~~ <sup>Language pattern:   he/she is telling</sup> telling the truth, ~~then they are a knight,~~
<sup>Circular reasoning:   if he is a knight then he is a knigh</sup> not a knave. ~~This contradicts their statement "We~~
~~are both knaves."~~<sup>Lack of commonsense:   the statement can be false or true</sup>

If we assume that the first person is a knave, then what ~~they~~
~~say~~<sup>Grammar pattern:   using plural instead of singular</sup> must be false. This is consistent because a
knave lying is expected behavior. If the first person is a knave and they are lying, ~~then~~
~~they are not both knaves.~~<sup>Implication does not hold</sup>

~~This   is   consistent   with   their   statement   "We   are   both~~
~~knaves."~~<sup>Inconsistency:   if the person lying, its statement cannot be true</sup>

~~Therefore, the only consistent interpretation is that both people are knaves.~~<sup>Wrong conclusion</sup>

---

**Puzzle 62. At least one of us is a knave**

On the island of knights and knaves, knights always tell the truth, while knaves always lie. You are approached by two people. The first one says: "At least one of us is a knave". What are they actually? (puzzle 28 from Smullyan (2011))

---

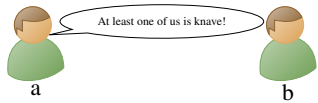

---

**Solution by human**

Mace4 finds a single model:

$$a = [0] \qquad\qquad b = [1]$$
$$inhabitant(\_) = [1, 1] \quad m(\_) = [1, 0]$$
$$knight(\_) = [1, 0] \qquad knave(\_) = [0, 1]$$

Here the inhabitant $a$ gets the value 0 from the domain, while the inhabitant $b$ gets the value 1. The function *inhabitant* is true for both $a$ and $b$. The function *knave* has value 0 for the first inhabitant ($a = 0$), and value 1 for the second one ($b = 1$). The function *knight* has value 1 for the first inhabitant ($a = 0$), and value 0 for the second one ($b = 1$).

---

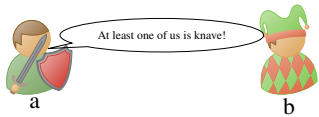

---

**Solution by ChatGPT**                                                        **Wrong**

If we assume that the first person is a knight, then what ~~they say~~ [Language pattern: maybe GPT uses it to avoid gender bias] must be true. However, this leads to a contradiction. If the first person is a knight and they are telling the truth, then both of them can't be knaves ~~as stated.~~ [Using info from previous puzzle] ~~This contradicts their statement "At least one of us is a knave."~~ [False contradiction: the statement "At least one of us is a knave" does not contradict the statement "Both of us are knaves"]

If we assume that the first person is a knave, then what they say must be false. This is consistent because a knave lying is expected behavior. If the first person is a knave and they are lying, then ~~it's possible~~ [Wrong modality: It's should] that neither of them can be a knave. ~~Therefore, the only consistent interpretation is that the first person is a knave and the second person is a knight.~~ [Wrong conclusion] ~~This way, the statement "At least one of us is a knave" made by the knave is false,~~ [Inconsistency: the statement is true] and ~~the second person (a knight) tells the truth.~~ [Lack of understanding: the second person says nothing]

---

ChatGPT API is stateless

**Puzzle 63. Either I am a knave or *b* is a knight**

On the island of knights and knaves, knights always tell the truth, while knaves always lie. You are approached by two people. The first one says: "Either I am a knave or B is a knight.". What are they actually? (puzzle 29 from Smullyan (2011))

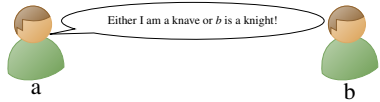

**Solution by human**

Mace4 computes a single model:

$$a = [0] \qquad\qquad b = [1]$$
$$inhabitant(\_) = [1, 1] \quad m(\_) = [1, 1]$$
$$knight(\_) = [1, 1] \qquad knave(\_) = [0, 0]$$

Here the message is true and both inhabitants are knights.

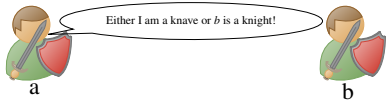

**Solution by ChatGPT**                          **Correct (but lack of task understanding)**

Let's consider the possibilities: <sup>Solving strategy:  Analysing possibilities</sup>
If the first person is a knight, then what ~~they say~~ <sup>Task understanding:  the statement belongs to a single person, not to both of them</sup> must be true. This would mean that either they are a knave (which is a contradiction) or the second person (B) is a knight. This is possible, so we continue to the next step.
If the second person (B) is a knight, then what they say must be true as well. If B is a knight, then either the first person is a knave (contradiction) or B is a knight (which is true). <sup>Task understanding:  GPT considers that B says somehting</sup> This creates a paradox and is not possible.
If the second person (B) is a knave, then what they say must be false. If B is a knave, then either the first person is a knave (which is true) or B is a knight (contradiction). This creates a paradox and is not possible.
Since there is a consistent solution where the first person is a knight and the second person (B) is a knight, that is the valid answer.

---

**Puzzle 64. We are both the same**

On the island of knights and knaves, knights always tell the truth, while knaves always lie. You are approached by two people. The first one says: "We are the same". Is the other person a knight or a knave? (adapted from Smullyan (2011))

---

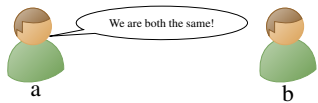

a                                   b

---

**Solution by human**

For $a = 0$ and $b = 1$, Mace computes two models:

|  Model 1 | Model 2 |
|---|---|
| relation(knave(_), [1, 0]) | relation(knave(_), [0, 0]) |
| relation(knight(_), [ 0, 1]) | relation(knight(_), [ 1, 1]) |
| relation(m(_), [ 0, 1]) | relation(m(_), [ 1, 1]) |

Note that in both models $b$ is a knight.

---

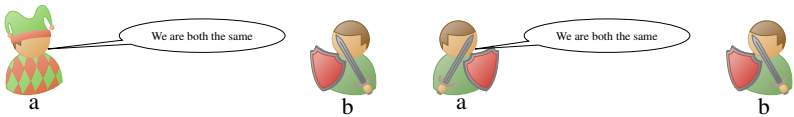

a                    b              a                    b

---

**Solution by ChatGPT**                    **Wrong (claiming not enough information)**

If the first person says "We are the same," this statement can only be true if both of them are the same, either both knights or both knaves. Since we know that knights always tell the truth and knaves always lie, we can analyze the possibilities: <span style="color:blue">Solving strategy: Analyse the possibilities</span>

If the first person is a knight, then ~~their~~<span style="color:blue">Task understanding: The statement belongs to a single person</span> statement is true. This means both of them are knights.

If the first person is a knave, then their statement is false. ~~This means both of them are knaves~~<span style="color:red">Implication does not hold Inconsistency: if both are knaves they are the same, which makes the statement true</span>

So, based on the statement "We are the same," ~~we can determine that both people are either both knights or both knaves.~~<span style="color:red">Implication does not hold</span> ~~We cannot definitively determine if the other person is a knight or a knave from this statement alone.~~<span style="color:red">Wrong conclusion</span>

**Puzzle 65. Three inhabitants and two messages**

On the island of knights and knaves, knights always tell the truth, while knaves always lie. We have three people, A, B, C, each of them is either a knight or a knave. A and B make the following statements:  A: "All of us are knaves".  B: "Exactly one of us is a knight". What are A, B, C? (puzzle 31 from Smullyan (2011))

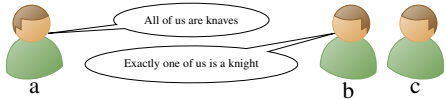

**Solution by human**

To avoid isomorphisms we fix $a = 0$, $b = 1$, $c = 2$ Mace4 finds a single model which includes the following relevant relations:

$$\text{relation(knave}(\_), [\ 1, 0, 1\ ])$$
$$\text{relation(knight}(\_), [\ 0, 1, 0\ ])$$
$$\text{relation(m}(\_), [\ 0, 1, 0\ ])$$

Here, the first message $m(a)$ is false and $m(b)$ is true. Also $b$ is a knight, while $a$ and $c$ are knaves.

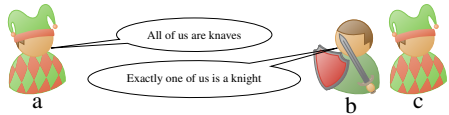

**Solution by ChatGPT**                                                                    **Wrong**

Let's analyze the statements made by A and B:
A says, "All of us are knaves." If A is a knight, then this statement must be true, ~~which means that all of them are indeed knaves.~~ Inconsistency:  GPT has assumed that A is a knight  If A is a knave, then this statement must be false, which contradicts ~~the assumption that all of them are knaves.~~ Wrong assumption:  the assumption holds if A is a knight  ~~So, A must be a knight.~~Implication does not hold
B says, "Exactly one of us is a knight." Since we already established that A is a knight, this means that B is a knave. If B was a knight, then B's statement would be false, since it would imply that both A and B are knights.
~~Given that A is a knight and B is a knave, we cannot determine the identity of C based on the given information alone. C can be either a knight or a knave.~~ Wrong conclusion

**Puzzle 66. Three inhabitants and not enough information**

On the island of knights and knaves, knights always tell the truth, while knaves always lie. We have three people, A, B, C, each of them is either a knight or a knave. A and B make the following statements: A: "All of us are knaves". B: "Exactly one of us is a knave". Can it be determined what B is? Can it be determined what C is? (puzzle 32 from Smullyan (2011))

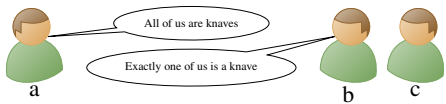

**Solution by human**

To avoid isomorphisms, we fix $a = 0$, $b = 1$, $c = 2$. Mace4 finds two models:

|  | Model 1 | Model 2 |
| --- | --- | --- |
| relation(knave(_), [ 1, 0, 0 ]) | relation(knave(_), [ 1, 1, 0 ]) |
| relation(knight(_), [ 0, 1, 1 ]) | relation(knight(_), [ 0, 0, 1 ]) |

In both models, $a$ is knave (i.e. first position in relation knave is 1 in both models). In both models $c$ is a knight (i.e. third position in relation knight is 1 in both models).

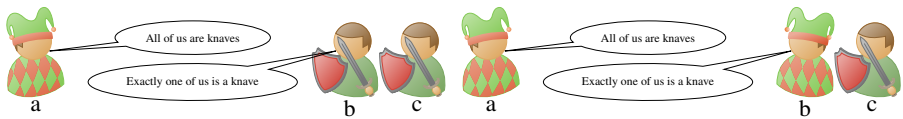

**Solution by ChatGPT**　　　　　　　　　　　　　　　　　　　　　**Wrong**

Let's analyze the statements made by A and B:

A says, "All of us are knaves." ~~If A is a knight, then this statement must be true, which means that all of them are indeed knaves.~~ ^(Inconsistency: A is knight implies A is knave) If A is a knave, then this statement must be false, which ~~contradicts the assumption that all of them are knaves.~~ ^(Wrong assumption) So, A must be a knight. ^(Implication does not hold)

B says, "Exactly one of us is a knave." If B is a knight, then this statement must be true. This means that one of them is a knave and the other two are knights. ~~However, since A is a knight,~~ ^(Wrong assumption) ~~this would imply that C is a knight as well,~~ ^(Implication does not hold) which contradicts the assumption that there is only one knave. Therefore, B cannot be a knight. Since B cannot be a knight, B must be a knave.

As for C, since we know that A is a knight and B is a knave, ~~we cannot definitively determine whether C is a knight or a knave based on the given information alone.~~ ^(Wrong conclusion: If A would be a knigth, his statement should be true, hence C should be a knave.)

**Puzzle 67. Three inhabitants and two of the same type**

On the island of knights and knaves, knights always tell the truth, while knaves always lie. We have three people, A, B, C, each of them is either a knight or a knave. A and B make the following statements: A: "B is a knave". B: "A and C are of the same type". What is C? (puzzle 34 from Smullyan (2011))

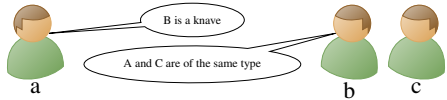

**Solution by human**

To avoid isomorphisms we fix $a = 0$, $b = 1$, $c = 2$. Mace4 finds two models:

|                              | Model 1                    | Model 2                    |
| ---------------------------- | -------------------------- | -------------------------- |
| relation(knave(_),           | [ 1, 0, 1 ])               | [ 0, 1, 1 ])               |
| relation(knight(_),          | [ 0, 1, 0 ])               | [ 1, 0, 0 ])               |

Observe that in both models, $c$ is knave (i.e. the third position corresponding to $c = 2$ in relation knave is 1 in both models).

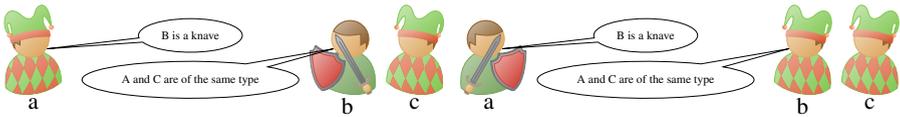

**Solution by ChatGPT** — **Wrong (partial correctness but not backed by reasoning**

Let's analyze the statements made by A and B:

A says, "B is a knave." If A is a knight, then this statement must be true, which means that B is indeed a knave. If A is a knave, then this statement must be false, which would imply that B is a knight, ~~leading to a contradiction.~~ [False contradiction: the statement is false] ~~Therefore, A must be a knight, and B is a knave.~~ [Implication does not hold]

B says, "A and C are of the same type." Since we know that A is a knight and B is a knave, for B's statement to be true, both A and C must be of the same type. This means that either both A and C are knights, or both A and C are knaves.

Since A is a knight and A's statement about B being a knave is true, it implies that B is indeed a knave. ~~Therefore, C must also be a knave to satisfy B's statement that A and C are of the same type.~~ [Inconsistency:  B is a kanve, so his statement should be false]

In summary:  ~~A is a knight. B is a knave. C is a knave.~~ [Unsupported conclusion] [Partial solution]

**Puzzle 68. Jim, Jon, and Joe**

On the island of knights and knaves, knights always tell the truth, while knaves always lie. You are approached by three people: Jim, Jon and Joe. Jim says: "Joe is a knave or I am a knight". Jon says, "Jim could claim that I am a knave". Joe says, "Neither Jim nor Jon are knights". What are they actually? (taken from Popular mechanics - www.popularmechanics.com/science/math)

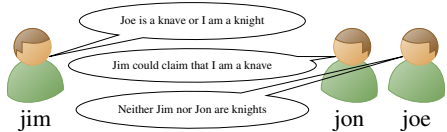

**Solution by human**

Mace4 finds a single model (see Figure **??**):

$$\text{relation(knave(\_), [ 1, 0, 1 ])}$$
$$\text{relation(knight(\_), [ 0, 1, 0 ])}$$

So *Jim* = 0 (i.e. first position) and *Jon* = 2 (i.e. third position) are knaves, while *Joe* = 1 (i.e second position) is a knight.

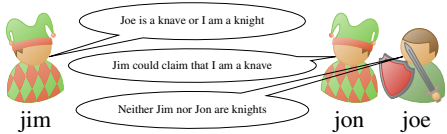

**Solution by ChatGPT**                                    **Wrong**

Let's analyze the statements made by Jim, Jon, and Joe:
Jim says, "Joe is a knave or I am a knight." If Jim is a knight, then his statement is true, ~~which means Joe must be a knave.~~ Implication does not hold If Jim is a knave, then his statement is false, which would imply that Joe is a knight. ~~This leads to a contradiction, so Jim must be a knight.~~ False contradiction
Jon says, "Jim could claim that I am a knave." ~~Since we know that Jim is a knight,~~ Wrong assumption Jon's statement must be true. This means that Jim could indeed claim that Jon is a knave.
Joe says, "Neither Jim nor Jon are knights." Since we already established that Jim is a knight, Joe's statement must be false. This implies that Joe is a knave.
In summary: ~~Jim is a knight. Jon is a knight. Joe is a knave.~~ Wrong conclusion

**Puzzle 69. A spy appears**

On the island of knights and knaves a new type on inhabitants has settled: spies. Spies can lie or tell the truth at will. Knights always tell the truth, while knaves always lie. You are approached by three people wearing different colored clothes. You know that one is a knight, one is a knave, and one is a spy. They speak in the following order: The man wearing blue says: "I am a knight". The man wearing red says: "He speaks the truth". The man wearing green says: "I am a spy". Who is the knight, who is the knave, and who is the spy? (taken from Popular mechanics - www.popularmechanics.com/science/math)

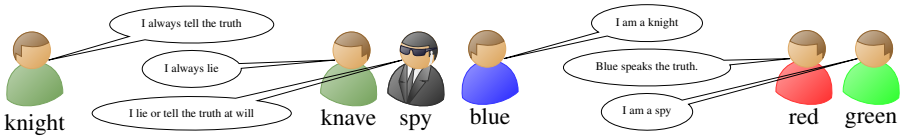

**Solution by human**

Mace4 computes a single model in which *knight*(*blue*), *knave*(*green*), and *spy*(*red*). In this model, the spy's message *m*(*spy*) is true.

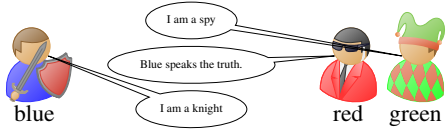

**Solution by ChatGPT**                                                                    **Wrong**

Let's analyze the statements made by the individuals:
The man wearing blue says, "I am a knight." If the man in blue is a knight, then his statement is true, but if he is a knave or a spy, ~~his statement is false.~~
Implication does not hold:  spies can lie at will

The man wearing red says, "He speaks the truth." If the man in red is a knight, then he must be speaking the truth. If he is a knave or a spy, ~~then his statement would be false.~~
Implication does not hold:  spies can lie at will

The man wearing green says, "I am a spy." If the man in green is a knight or a spy, ~~then his statement is false.~~ Implication does not hold:  spies can lie at will ~~If he is a knave, then his statement is true.~~ Inconsistency

Given the information provided, ~~the only consistent solution is that the man in red is the knight, the man in blue is the knave, and the man in green is the spy.~~ Unsupported conclusion
Wrong conclusion

## Puzzle 70. Who is the spy?

On the island of knights and knaves a new type on inhabitants has settled: spies. Spies can lie or tell the truth at will. Knights always tell the truth, while knaves always lie. You are approached by three people wearing different colored clothes. You know that one is a knight, one is a knave, and one is a spy. "Who is the spy?" you ask. The man wearing blue says: "That man in red is the spy". The man wearing red says: "No, the man in green is the spy". The man wearing green says: "No, the man in red is in fact the spy." (taken from Popular mechanics - www.popularmechanics.com/science/math)

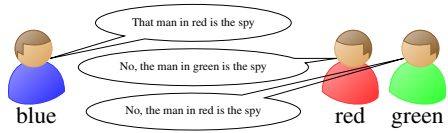

## Solution by human

Mace4 computes a single model in which *knight*(*red*), *knave*(*blue*), and *spy*(*green*) are true. In this model, the spy's message *m*(*spy*) is false.

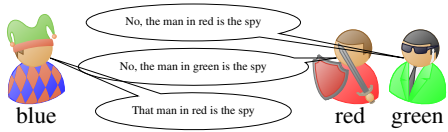

## Solution by ChatGPT — Wrong

Given the statements:
The man in blue says the man in red is the spy. The man in red says the man in green is the spy. The man in green says the man in red is the spy.
Since we know there's only one spy, one knight, and one knave, and all of them are giving different statements, ~~the consistent solution is that the man in blue is the knight, the man in red is the spy, and the man in green is the knave.~~ Unsupported conclusionWrong conclusion

---

**Puzzle 71. The whole truth and nothing but the truth**

On a famous island in the Pacific Ocean (whose name I forgot) live three tribes: the *Whites*, who always tell the truth; the *Blacks*, who always tell lies; and the *Grays*, who lie and tell the truth alternately (although their first answer may be either truth or lie). These natives always gather in groups of three, with one representative of each tribe in the group. I approached such a group of three natives, and had the following conversation with the first native: 'Are you the White, the Black, or the Gray?' 'I am the Gray.' 'And what about your friend here?' 'He is a Black.' 'So, your other friend is the White?' 'Of course.' Was the 'other friend' a White, and if not, what was he? (Clessa (1996))

---

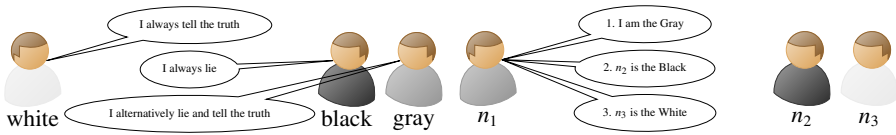

---

**Solution by human**

The functions $n_1$, $n_2$, and $n_3$ represent the three natives. Mace4 outputs a single model: $n_1$ (i.e. the person that answers) is black, $n_2$ is white, while $n_3$ is gray.

---

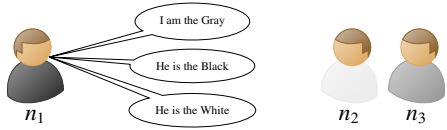

---

**Solution by ChatGPT**                                                                                      **Wrong**

Let's analyze the statements: Native 1 claims to be Gray. Native 1 claims Native 2 to be Black. Native 1 claims Native 3 to be White. Since we know that one of them is White, one is Black, and one is Gray, and they are from different tribes, let's evaluate the possibilities for Native 1: If Native 1 is White: Statement 2: ~~Native 2 cannot be Black because Whites tell the truth.~~ <span style="color:red">Inconsistency:   s2 claims that n2 is black</span> Statement 3: This would mean Native 3 is White, but that contradicts the initial condition that each tribe is represented. If Native 1 is Black: Statement 2: ~~This would mean Native 2 is Gray~~ <span style="color:red">Implication does not hold:   he can be white also</span>, ~~which would make Statement 3 false~~ <span style="color:red">Implication does not hold</span> ~~(Gray alternating truth and lie).~~ <span style="color:red">Unrelated justification:   the assumption is that n1 is black, hence he alsways lie</span> Statement 3: Native 3 cannot be White because Blacks lie. If Native 1 is Gray: Statement 2: ~~This would mean Native 2 is Black,~~ <span style="color:red">Implication does not hold:   s1 is true, hence the second one should be false</span> ~~which fits the fact that Grays alternate truth and lie.~~ <span style="color:red">Inconsistency:   the assumption is that n1 is gray and his s1 is true</span> Statement 3: ~~This would mean Native 3 is Gray, which also fits the alternating truth and lie of Grays.~~ <span style="color:red">Implication does not hold</span> ~~So, it is consistent if Native 1 is Gray, Native 2 is Black, and Native 3 is Gray.~~ <span style="color:red">Wrong conclusion</span>

## Puzzle 72. Three goddesses

Three goddesses were sitting in an old Indian temple. Their names were Truth (always telling the truth), Lie (always lying) and Wisdom (sometimes lying). A visitor asked the one on the left: "Who is sitting next to you?" "Truth," she answered. Then he asked the one in the middle: "Who are you?" "Wisdom." Last, he asked the one on the right: "Who is your neighbour?" "Lie," she replied. And then it became clear who was who. (taken from www.brainden.com)

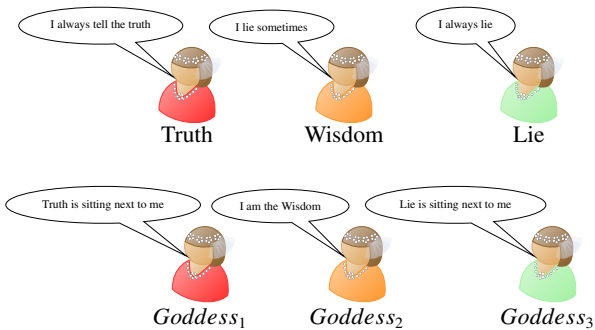

## Solution by human

Let Truth ($T$), Wisdom ($W$), Lie ($L$) are in left (i.e. 1), middle (i.e. 2) or right (i.e. 3) positions. Mace4 finds a model in which Lie is in the middle, Truth on the right, and Wisdom on the left:

| Model | $L_1$ | $L_2$ | $L_3$ | $T_1$ | $T_2$ | $T_3$ | $W_1$ | $W_2$ | $W_3$ |
|-------|-------|-------|-------|-------|-------|-------|-------|-------|-------|
| 1     | 0     | 1     | 0     | 0     | 0     | 1     | 1     | 0     | 0     |

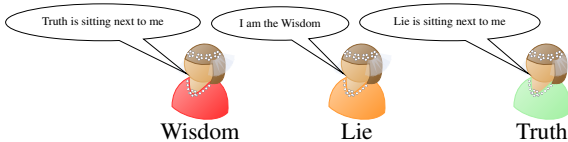

**Solution by ChatGPT** — **Wrong**

Let's analyze the statements:

The goddess on the left said her neighbor is "Truth."

The goddess in the middle said she is "Wisdom."

The goddess on the right said her neighbor is "Lie."

If the goddess on the left is "Truth," then her statement about her neighbor being "Truth" would be false, which contradicts her nature of telling the truth. So, the goddess on the left cannot be "Truth."

If the goddess on the left is "Lie," ~~then her statement about her neighbor being "Truth" would be true,~~ Implication does not hold: she can be "Wisdom" which also contradicts her nature of lying. So, the goddess on the left cannot be "Lie."

Therefore, the only option left for the goddess on the left is to be "Wisdom."

Now, if the goddess in the middle is "Truth," then her statement about being "Wisdom" would be false, which contradicts her nature of telling the truth. So, the goddess in the middle cannot be "Truth."

If the goddess in the middle is "Lie," ~~then her statement about being "Wisdom" would be true,~~ Implication does not hold which also contradicts her nature of lying. So, the goddess in the middle cannot be "Lie."

Therefore, the only option left for the goddess in the middle is to be "Wisdom."

Now, ~~with the goddess on the left being "Wisdom" and the one in the middle being "Wisdom,"~~ Inconsistency: there a retwo wisdom ~~the only option left for the goddess on the right is "Lie."~~ Implication does not hold

So, based on their responses ~~The goddess on the left is "Wisdom." The goddess in the middle is "Wisdom." The goddess on the right is "Lie."~~ Wrong conclusion

# References

Aszalós, L. (2001). Automated solution of the riddle of Dracula and other puzzles. In *Proc. Int. Joint Conf. on Automated Reasoning (IJCAR'01)*, pages 1–10.

Baumgartner, P., Furbach, U., and Stolzenburg, F. (1995). Model elimination, logic programming and computing answers. In *IJCAI*, pages 335–341.

Ciraulo, F. and Maschio, S. (2020). Solving knights-and-knaves with one equation. *The College Mathematics Journal*, 51(2):82–89.

Clessa, J. (1996). *Math and Logic Puzzles for PC Enthusiasts*. Courier Corporation.

Smullyan, R. M. (2011). *What is the name of this book? The riddle of Dracula and other logical puzzles*. Dover Publications.



# Chapter 7

# Love and marriage

**Puzzle 73. Looking at unmarried people**

There are three friends staying on the couch in Central Perk: Rachel, Ross, and Monica. Monica is looking at Ross. Ross is looking at Rachel. Monica is married; Rachel is not. Is a married person looking at an unmarried person?

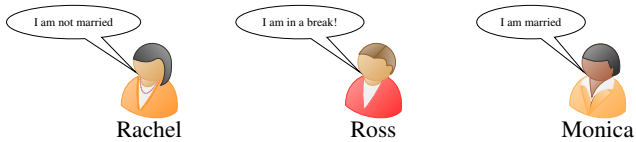

**Solution by human**

There are two models in which Ross is unmarried and respectively married. In both of them, a married person is looking to an unmarried one.

**Solution by ChatGPT**                    **Correct (but wrong justification)**

Yes, a married person is looking at an unmarried person. ~~Ross is married to Rachel~~ The worlds collide: GPT uses information from Friends TV series and he is looking at her. ~~Since Rachel is unmarried,~~ Inconsistency: GPT has just stated that Ross is married to Rachel this scenario fits the given information.

**Puzzle 74. Married people do not lie**

Married people always tell the truth, while unmarried people always lie. There are four friends staying on the couch in Central Perk: Ross, Monica, Rachel and Chandler. They made the following statements:
Ross: "Rachel and I have different marital status".
Rachel: "Chandler is not married".
Monica: "I am married to Chandler".
Chandler: "Monica is lying".
How many are married, and how many are unmarried?

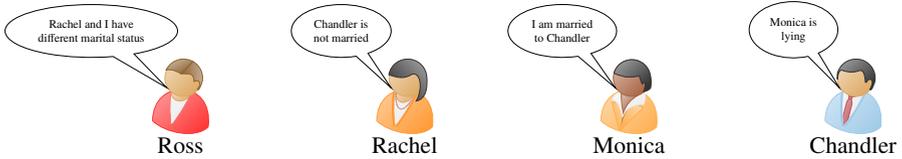

**Solution by human**

Mace4 finds a single model in which only Chandler and Ross are married.

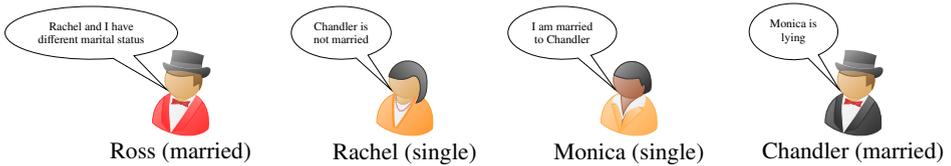

**Solution by ChatGPT**                                                    **Wrong**

Let's analyze the statements one by one:
Ross: "Rachel and I have different marital status". This statement can be true only if one of them is married and the other is unmarried.
Rachel: "Chandler is not married". ~~Since Rachel is unmarried,~~ <sup>Unsupported claim</sup> ~~if she's telling the truth,~~ <sup>Inconsistency: unmarried people always lie</sup> Chandler must be unmarried.
Monica: "I am married to Chandler". If Monica is married to Chandler, then her statement must be true, and since we already determined that Chandler is unmarried, this statement is false.
Chandler: "Monica is lying". If Monica is lying, then this statement is true, which means Monica is indeed lying.
From the above analysis, we can conclude that Monica is married, Chandler is unmarried, and ~~Rachel is unmarried.~~ <sup>Unsupported claim</sup> Ross's statement about having different marital status with Rachel is true, so Ross must be married.
In summary: ~~Married: Ross, Monica. Unmarried: Rachel, Chandler~~<sup>Wrong conclusion</sup>

**Puzzle 75. Minos and aminos: we are both married**

News has reached me, Auspicious King, of a curious town in which every inhabitant is either a Mino or an Amino. "Oh my goodness, what are they?" asked the king. "The Minos are worshipers of a good god; whereas the Aminos worship an evil god. The Minos always tell the truth—they never lie. The Aminos never tell the truth-they always lie. All members of one family are of the same type. Thus given any pair of brothers, they are either both Minos or both Aminos. Now, I heard a story of two brothers, Bahman and Perviz, who were once asked if they were married. They gave the following replies:
Bahman: We are both married.
Perviz: I am not married.
Is Bahman married or not? And what about Perviz? (taken from Smullyan (1996))

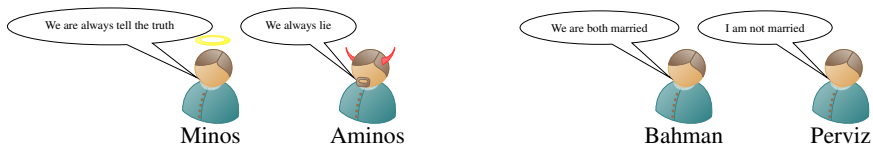

**Solution by human**

A single model is found by Mace4. Both brothers are amino, both statements *m* are false, married relation is false for domain value 0 (i.e. Bahman) and true for domain value 1 (i.e. Perviz).

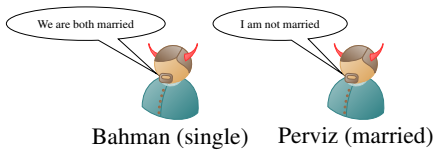

Bahman (single)    Perviz (married)

**Solution by ChatGPT**                                              **Wrong**

Let's analyze the statements:
Bahman: "We are both married." This statement can be true only if both Bahman and Perviz are married. However, if Bahman is an Amino, he would be lying, ~~which means they are both unmarried.~~ Implication does not hold: one unmarried suffices This creates a contradiction.
Perviz: "I am not married." ~~This statement can be true only if Perviz is an Amino~~ Implication does not hold: Aminos alway lie and indeed not married. ~~If Perviz is a Mino, he would be lying,~~ Implication does not hold: Minos always tell the truth which means he is married.
From the analysis, we can conclude: Perviz is an Amino and is not married. Bahman is a Mino and is also not married.
In summary: ~~Bahman is not married. Perviz is not married.~~ Wrong conclusion

### Puzzle 76. Minos and aminos: we are both married or both unmarried

News has reached me, Auspicious King, of a curious town in which every inhabitant is either a Mino or an Amino. "Oh my goodness, what are they?" asked the king. "The Minos are worshipers of a good god; whereas the Aminos worship an evil god. The Minos always tell the truth—they never lie. The Aminos never tell the truth-they always lie. All members of one family are of the same type. Thus given any pair of brothers, they are either both Minos or both Aminos. Now, I heard a story of two brothers, Bahman and Perviz, who were once asked if they were married. They gave the following replies:
Bahman: "We are both married or both unmarried".
Perviz: "I am not married".
What can be deduced about Bahman and Perviz? (Smullyan (1996))

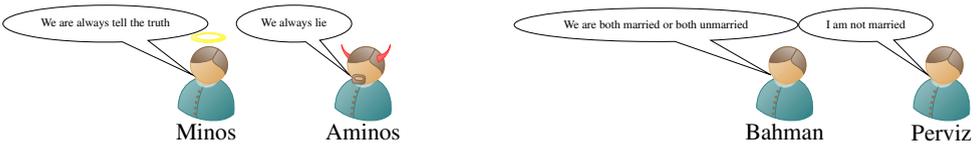

Minos          Aminos                          Bahman          Perviz

### Solution by human

There are two models: In the first model, both sentences are true. In the second model, both sentences are false. Note that in both models, Bahman is not married.

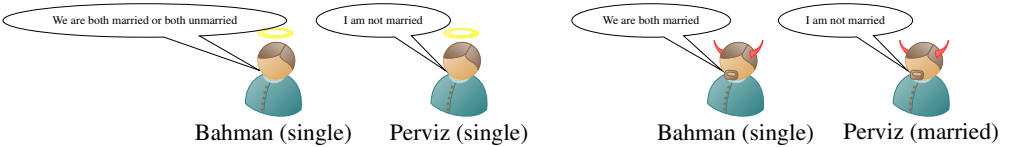

Bahman (single)   Perviz (single)          Bahman (single)   Perviz (married)

### Solution by ChatGPT                                    Wrong

Let's analyze the statements:
Bahman: "We are both married or both unmarried." This statement indicates that either both Bahman and Perviz are married or both are unmarried. This implies that they cannot have different marital statuses. If Bahman is a Mino, this statement would be true, and both of them would have the same marital status. If Bahman is an Amino, he would be lying, ~~which creates a contradiction.~~ <sup>False contradiction</sup>
Perviz: "I am not married." ~~This statement can be true only if Perviz is an Amino~~<sup>Implication does not hold</sup> and indeed not married. ~~If Perviz is a Mino, he would be lying,~~<sup>Implication does not hold</sup> which means he is married.
From the analysis, we can conclude: ~~Perviz is an Amino and is not married.~~<sup>Unsupported claim</sup>
Since Bahman's statement leads to a contradiction, we cannot deduce anything specific about Bahman's marital status from the information given.
In summary: <sup>Wrong conclusion</sup> ~~Perviz is not married. Bahman's marital status cannot be determined with certainty based on the information provided.~~<sup>Wrong undeterminancy</sup>

**Puzzle 77. Marriage in company**

Bob, Janet, and Shirley hold the positions of director, engineer, and accountant, but not necessarily in that order. The accountant is an only child and earns the least. Shirley is married to Bob's brother and earns more than the engineer. What job does each person have?

**Solution by human**

Mace4 finds a single model: *function(job( _ ), [ 1, 0, 2 ])*. Here *job*(0) = 1 means *job*(*bob*) = *engineer*, *job*(1) = 0 means *job*(*janet*) = *accountant*, and *job*(2) = 2 means *job*(*shirley*) = *director*.

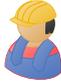   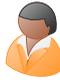   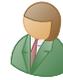

Bob (engineer)          Janet (accountant)          Shirley (director)

**Solution by ChatGPT**                                                **Wrong**

From the information given:
The accountant is an only child and earns the least. Shirley is married to Bob's brother and earns more than the engineer.
Let's analyze the information step by step:
Since the accountant is an only child and earns the least, ~~they~~ <sup>Grammar pattern: avoiding bias?</sup> ~~cannot be Bob or Shirley,~~ <sup>Unsupported claim</sup> ~~as both are mentioned to be married to someone.~~
<sup>Unsupported premise: only Sherly is married</sup> Therefore, the accountant must be Janet.
Now, we have two possibilities for the remaining positions: Bob is the director and Shirley is the engineer. Shirley is the director and Bob is the engineer.
According to the second statement, Shirley earns more than the engineer. Since we know that Janet is the accountant, she earns the least. ~~This means Shirley cannot be the engineer~~<sup>Implication does not hold: Shirly cannot be an engineer because she earns more than an engineer</sup> leaving only one possibility:
Bob is the director, Shirley is the accountant, and ~~Janet is the engineer~~<sup>Inconsistency: GPT has just infered that Shirley cannot be an engineer</sup>
So, the job of each person is: ~~Bob: Director Shirley: Accountant Janet: Engineer~~<sup>Wrong conclusion</sup>

**Puzzle 78. What is my relationship to Teresa?**

Teresa's daughter is my daughter's mother. What is my relationship to Teresa? Are you:
1. Teresa's grandmother
2. her mother
3. her daughter
4. her granddaughter
5. Teresa
(taken from Brainzilla - www.brainzilla.com)

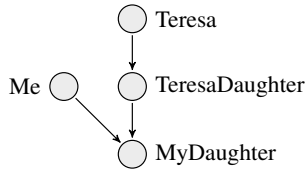

**Solution by human**

There are four named individuals: *Teresa*, *TeresaDaughter*, *MyDaughter* and I (*Me*). Hence we start with a domain size of 4. However, we do not know whether these individuals are distinct or not.

For *Me* = *TeresaDaughter*, Mace4 finds the model depicted below, where the arrow represents the *hasDaughter* relation. In this model, *Me* is the same with Teresa's daughter, since both variables are linked to value 0. Note that *Me* = 0 and *TeresaDaughter* = 0, hence these variables represent the same individual. Thus, this is the answer to the quiz. Mace4 finds no model for the other four options of the quiz, even if the domain size is increased accordingly.

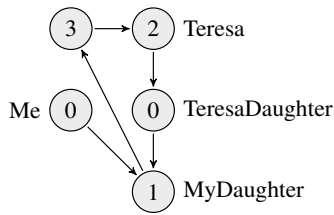

**Solution by ChatGPT**                                                          **Correct**

Your relationship to Teresa is her daughter.
Explanation:
Teresa's daughter is your daughter's mother. Since your daughter's mother is you, it means that you are Teresa's daughter.

**Puzzle 79. Who is Helen's husband?**

Yesterday evening, Helen and her husband invited their neighbours (two couples) for a dinner at home. The six of them sat at a round table. Helen tells you the following: "Victor sat on the left of the woman who sat on the left of the man who sat on the left of Anna. Esther sat on the left of the man who sat on the left of the woman who sat on the left of the man who sat on the left of the woman who sat on the left of my husband. Jim sat on the left of the woman who sat on the left of Roger. I did not sit beside my husband." The question: What is the name of Helen's husband? (http://www.puzzlesite.nl)

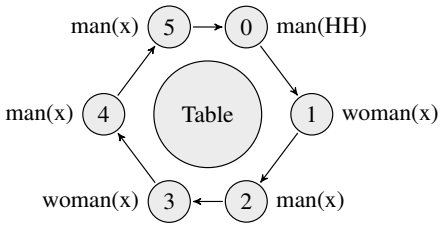 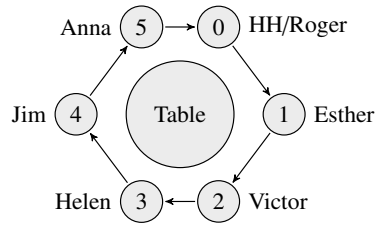

**Solution by human**

We assume Roger, Jim and Victor are men, while Anna, Helen and Esther are women. We note Helen's husband with *HH* and we assume he is a man. Mace4 finds a single model:

| Model | *Anna* | *Esther* | *HH* | *Helen* | *Jim* | *Roger* | *Victor* |
|-------|--------|----------|------|---------|-------|---------|----------|
| 1     | 5      | 1        | 0    | 3       | 4     | 0       | 2        |

Since *HH* and *Roger* have the same value 0, Roger is Helen's husband.

**Solution by ChatGPT**                                    **Correct (but unjustified)**

The name of Helen's husband is Roger.
Explanation: Victor sat on the left of the woman who sat on the left of the man who sat on the left of Anna. Esther sat on the left of the man who sat on the left of the woman who sat on the left of the man who sat on the left of the woman who sat on the left of Helen's husband. Jim sat on the left of the woman who sat on the left of Roger. Helen did not sit beside her husband.
From these statements, we can deduce the seating arrangement:
~~Anna - Man 1 - Woman 1 - Victor - Esther - Roger - Jim - Woman 2~~ Unjustified claim Lack of commonsense: there are 8 persons here instead of 6

Therefore, Helen's husband is Roger.

**Puzzle 80. Arranged royal marriage**

You are a young prince in situation to choose marrying one the three princesses: one from the north neigbouring country, one from the south, or one from the east. Each princess secretly sent you some messages. The first one says: "If you marry the princess from the south, there will be war". The second one says: "Marrying any of us will not bring peace". You learn from the third princess that: "Marry me and it will be peace, marry the south princess and there will be war". Given that you know that all the princesses are liars, can you choose a princess that brings peace for sure?

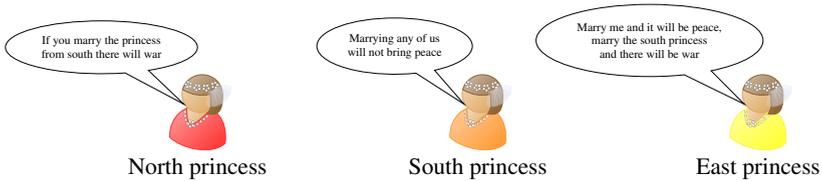

North princess                      South princess                      East princess

**Solution by human**

Mace4 finds a single model in which *east* = 0, *north* = 1, *south* = 2, and the corresponding married relation is *married*(_) = [0, 0, 1]. That is, the South princess.

**Solution by ChatGPT**                                              **Wrong**

Yes, you can choose a princess who brings peace for sure.
~~The princess from the north will bring peace.~~ <sup>Wrong conclusion</sup>
Explanation:
~~If the princess from the south is telling the truth,~~ <sup>Wrong assumption: all princesses lie</sup> then marrying her will result in war. But since all princesses are liars, this statement is false. So, the princess from the south is lying.
~~If the princess from the east is telling the truth,~~ <sup>Wrong assumption: all princesses lie</sup> then marrying any princess will not bring peace. But since all princesses are liars, this statement is false. So, the princess from the east is lying.
~~If the princess from the north is telling the truth,~~ <sup>Wrong assumption: all princesses lie</sup> then marrying her will bring peace. Since all princesses are liars, this statement is false. So, the princess from the north is lying.
~~Since only the statement of the princess from the north can be false,~~ <sup>Unjustified claim</sup> ~~she must be telling the truth.~~ <sup>Inconsistency: GPT has just stated that the princess from the north is lying</sup> . ~~Therefore, marrying the princess from the north will bring peace.~~ <sup>Implication does not hold</sup>

**Puzzle 81. Two single persons at the end of the row**

Four married men and three unmarried men are seated in a row at random. What are the chances that the two men at the ends of the row will be single? (adapted from puzzle 470 from Dudeney (2016))

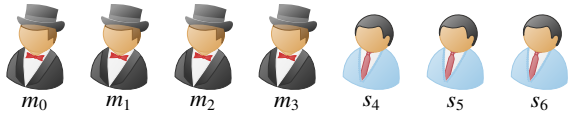

$m_0$ $m_1$ $m_2$ $m_3$ $s_4$ $s_5$ $s_6$

**Solution by human**

For the possible models in which the seven persons can be arranged, Mace4 returns 5,040 models. For the favorable models in which two single persons are at the end of the row, Mace4 finds 720 models. Hence, the probability is 720/5040.

**Solution by ChatGPT**      **Wrong**

To solve this probability problem, let's consider the different ways the married and unmarried men can be seated at random.

There are a total of 7 men: 4 married and 3 unmarried.

To calculate the probability of having the two men at the ends of the row be unmarried, we can consider the total number of possible arrangements where this condition is satisfied. Let's analyze the possible arrangements:

The first and last men are unmarried ~~(3 possibilities)~~[Incomplete analyse: there are more then 3 possibilities].

~~U _ _ _ _ U~~[Inconsistency: there are 7 men, but here there are only 6 places]

~~U _ _ _ U _~~[Inconsistency: the last men does not have U]

~~U _ _ U _ _~~[Inconsistency: the last men does not have U]

The second and second-to-last men are unmarried ~~(3 possibili-ties)~~[Incomplete analyse: there are more then 3 possibilities]:

~~_ U _ _ _ U~~[Inconsistency: the second to last men does not have U]

_ U _ _ U _

~~_ U _ U _ _~~[Inconsistency: the second to last men does not have U]

Both first and last men are unmarried ~~(1 possibility)~~[Incomplete analyse: there are more then 3 possibilities]

[Inconsistency: GPT has already analysed this case] U _ _ _ _ U

~~Total favorable arrangements: 3 + 3 + 1 = 7~~[Wrong computation for favorable models]

Total possible arrangements: 7! (7 factorial, which is 7 x 6 x 5 x 4 x 3 x 2 x 1)

So, ~~the probability that the two men at the ends of the row will be unmarried is: 7/7! 0.00198 or about 0.2%~~[Wrong conclusion].

**Puzzle 82. Five couples**

My wife and I recently attended a party at which there were four other married couples. Various handshakes took place. No one shook hands with oneself, nor with one's spouse, and no one shook hands with the same person more than once. After all the handshakes were over, I asked each person, including my wife, how many hands he (or she) had shaken. To my surprise each gave a different answer. How many hands did my wife shake? (puzzle 1.18 from Gardner (2005))

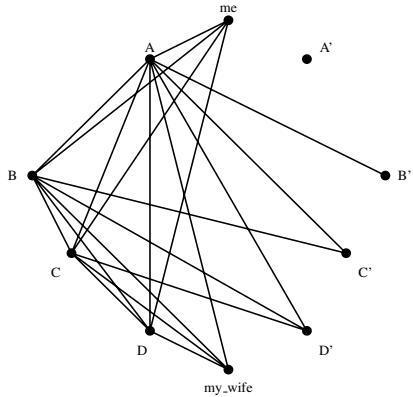

**Solution by human**

If we assume that "all the handshakes were over" means that each couple saluted each other, we have a single model: only *me* and *my_wife* shook the same number of hands

| Model | *me* | *my_wife* | $H_2$ | $H_3$ | $H_4$ | $H_5$ | $W_2$ | $W_3$ | $W_4$ | $W_5$ |
|-------|------|-----------|-------|-------|-------|-------|-------|-------|-------|-------|
| 1     | 4    | 4         | 3     | 2     | 1     | 0     | 5     | 6     | 7     | 8     |

**Solution by ChatGPT**                                                      **Wrong**

Let's analyze the information provided step by step: There are 4 other married couples present at the party ~~which makes a total of 8 couples including you and your wife.~~^Lack of commonsense In each handshake, two people are involved. So, the total number of handshakes can be calculated as the number of ways to choose 2 people out of the total 16 individuals (8 couples). Total Handshakes = 16 choose 2 = 16! / (2! * (16 - 2)!)  = 120 Now, let's consider the situation of each person and how many handshakes they could have had: You (1 handshake): ~~You shook hands with each of the 8 couples~~^Lack of commonsense:  confusion between couples and persons, which makes 8 handshakes. Your wife: The question is about how many hands your wife shook. Since every handshake involves two people, ~~the total number of hands she shook must be equal to the total number of handshakes minus the number of hands you shook (8 hands)~~^Implication does not hold. ~~Your wife shook 120 - 8 = 112 hands.~~^Lack of commonsense:  there are no so many hands in the puzzle The other 15 people: Since each person's answer was different, and no one shook hands with themselves or their spouse, each of the other 15 people shook a different number of hands ranging from 0 to 14. ~~So, your wife shook hands with 112 people at the party.~~^Wrong conclusion

**Puzzle 83. A family tree**

I was going through some old family photos in the attic when I stumbled upon our family tree. I studied it for a couple minutes then went back down stairs to tell my mom about the family tree. The problem is I didn't study it long enough to remember the whole thing. I only remembered a couple of things about it, and recent memories. Can you help me figure out my family tree? There are two grandparents, who had two children, who both got married and had 2 more children each. In total, there are 10 people: Alex, Bob, Caty, David, Elton, John, Lincoln, Mary, Sonia, and Tina.

Clue$_1$:   One of Elton's ancestors was David.
Clue$_2$:   John's sister gave birth to Tina.
Clue$_3$:   Mary went bowling with her nephew last Saturday.
Clue$_4$:   Alex is cousin with one of the girls.
Clue$_5$:   Bob married Mary.
Clue$_6$:   Caty is not an ancestor, nor cousin of Tina.
Clue$_7$:   Lincoln's brother showed Bob's son his baseball cards.

(adapted from www.braingle.com/brainteasers/23381/family-tree.html)

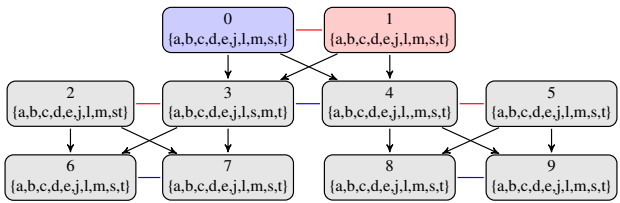

Figure 7.1: A family of ten members over three generations. Red arrows represent the *marry* relation, blue arrows represent the *sibling* relation, and black arrows represent the *ancestor* relation. Blue nodes are males, while pink nodes are females. Unknown gendre is depicted with gray nodes.

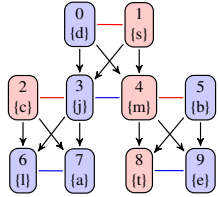

Figure 7.2: One model found by Mace4 (where $d = 0$)

**Solution by human**

Sonia and David were the grandparents. They gave birth to John and Mary. John married Caty, who gave birth to Lincoln and Alex. Mary married Bob and gave birth to Tina and Elton. These fixpoints can be checked by adding the test cases from Listing **??**. If we fix *John* = 3, *David* = 0, and *Tina* = 8, 6 isomorphic models remain. One such model is depicted in Figure 7.3).

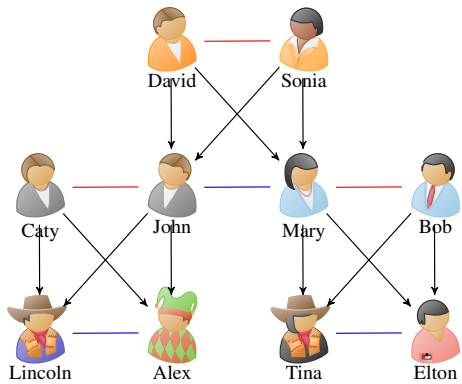

Figure 7.3: A family tree

---

**Solution by ChatGPT** — **Wrong**

Based on the clues provided, we can deduce the following relationships:
There are two grandparents. These grandparents had two children, making them parents to the next generation. Each of these children got married and had two more children each, forming the third generation. In total, there are 10 people: Alex, Bob, Caty, David, Elton, John, Lincoln, Mary, Sonia, and Tina.

Let's analyze each clue:

Clue1: One of Elton's ancestors was David. This means that Elton is younger than David.

Clue2: John's sister gave birth to Tina. ~~This indicates that John and Tina are siblings.~~ Lack of commonsense: John is the uncle of Tina

Clue3: Mary went bowling with her nephew last Saturday. ~~Since John is the only male sibling mentioned so far~~ Wrong assumption: male siblings could appear in the following clues, Mary must be his sister. This makes Tina her daughter, and ~~Tina's brother would be Mary's nephew.~~ Lack of commonsense: Tina's brother would be Mary's son.

Clue4: Alex is cousin with one of the girls. Since John and Tina are siblings, ~~Alex must be their cousin.~~ Implication does not hold ~~This places Alex in the same generation as John and Tina's parents.~~ Implication does not hold

Clue5: Bob married Mary. This gives us the relationship between Bob and Mary.

Clue6: Caty is not an ancestor, nor cousin of Tina. ~~This indicates that Caty cannot be part of the same generation as Tina or her ancestors.~~ Implication does not hold

Clue7: Lincoln's brother showed Bob's son his baseball cards. ~~This implies that Bob's son is Lincoln's nephew.~~ Implication does not hold ~~This nephew is also Mary's nephew according to Clue 3.~~ Implication does not hold

Putting all this information together, we can create the family tree:

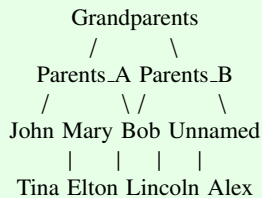

Key points: John and Mary are siblings. ~~John is Tina's parent.~~ Wrong conclusion Tina is Alex's cousin. ~~Elton is one of John's ancestors.~~ Wrong conclusion ~~Lincoln is Bob's son~~ Wrong conclusion and Mary's nephew. Caty's place in the family tree is not specified, ~~but we know she's not in the same generation as Tina or her ancestors.~~ Wrong conclusion Please note that there's an unnamed child of Parents_B. This is based on the information provided, as the specific gender or name of this child is not mentioned.

**Puzzle 84. Uncles and aunts**

The family tree consists of two grandparents, who had three children, each of whom got married and had two children. Males are Cole, Cristian, Jason, Neil, and Steve. Females are Amanda, Ashley, Beth, Erin, Kaitlyn, Katherine, Makayla, Payton, and Tammy. The clues are:

Clue$_1$   One of Makayla's cousins is Jason's son.
Clue$_2$   One of Ashley's aunts is Tammy.
Clue$_3$   Tammy's brother-in-law is Neil's son.
Clue$_4$   Kaitlyn's sister is Ashley's cousin.
Clue$_5$   Ashley's uncle, Steve, is Erin's brother-in-law.
Clue$_6$   The three uncles are Payton's dad, Cristian, and Katherine's son.
Clue$_7$   The three aunts are Kaitlyn's mom, Ashley's mom, and Cristian's sister-in-law.
Clue$_8$   Jason's brother is Ashley's dad.
Clue$_9$   Amanda's sister is Steve's niece.
Clue$_{10}$   Beth is not Cole's aunt.

(adapted from www.braingle.com/brainteasers/24668/family-tree-2.html)

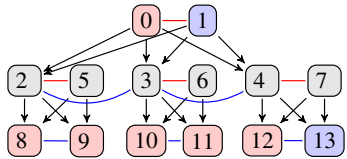

Figure 7.4: A family of 14 members over three generations. Red arrows represent the *marry* relation, blue arrows are the *sibling* relation, and black arrows are the *ancestor* relation. There are 5 men (blue) and 9 women (pink). We do not know if the grandparents 0 and 1 have daughters or sons.

**Solution by human**

Mace4 found 4 isomorphic models. The four models are given by Makayla and Kaitlyn can be 8 or 9, and Ashley and Amanda can be 10 or 11. The grandparents are Katherine and Neil, who had 3 sons: Cristian, Jason, and Steve. Cristian married Erin, and had Ashley and Amanda. Jason married Beth, and had Payton and Cole. Steve married Tammy, and had Makayla and Kaitlyn.

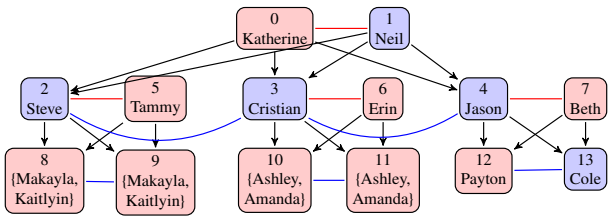

**Solution by ChatGPT** **Wrong**

Let's analyze each clue and use the information to construct the family tree:

Clue1: One of Makayla's cousins is Jason's son. ~~This means that Makayla and Jason are in the same generation~~<sup>Implication does not hold</sup>, and Jason has a son who is Makayla's cousin.

Clue2: One of Ashley's aunts is Tammy. This places Ashley and Tammy in different generations. Ashley's aunt could be one of Tammy's sisters.

Clue3: Tammy's brother-in-law is Neil's son. ~~This means that Tammy is Neil's sister~~<sup>Implication does not hold</sup>, and Neil has a son.

Clue4: Kaitlyn's sister is Ashley's cousin. Kaitlyn and Ashley are in the same generation, and Kaitlyn's sister is Ashley's cousin.

Clue5: Ashley's uncle, Steve, is Erin's brother-in-law. This means Ashley and Erin are in different generations. Steve's relationship to Erin is through Erin's spouse, so Erin is married to Steve's sibling.

Clue6: The three uncles are Payton's dad, Cristian, and Katherine's son. This tells us the relationships of three uncles. Payton's dad is one uncle, Cristian is another uncle, and Katherine's son is the third uncle.

Clue7: The three aunts are Kaitlyn's mom, Ashley's mom, and Cristian's sister-in-law. This tells us the relationships of three aunts. Kaitlyn's mom is one aunt, Ashley's mom is another aunt, and Cristian's sister-in-law is the third aunt.

Clue8: Jason's brother is Ashley's dad. ~~This tells us that Jason and Ashley are in the same generation~~<sup>Implication does not hold</sup>, and Jason's brother is Ashley's dad.

Clue9: Amanda's sister is Steve's niece. This means that Amanda and Steve are in different generations. Steve's niece could be Amanda's sister's daughter.

Clue10: Beth is not Cole's aunt.

This rules out Beth being Cole's aunt.

Based on these clues, let's construct the family tree:

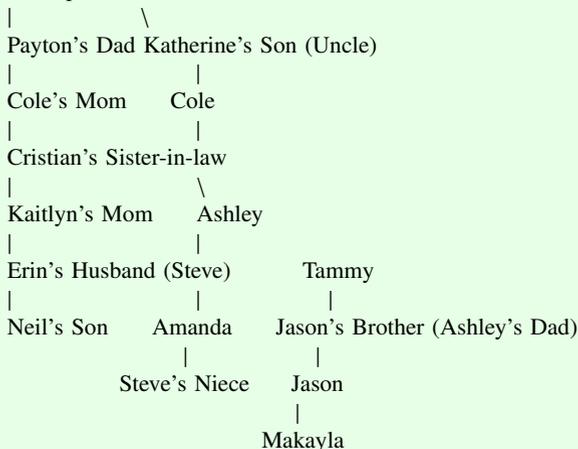

Grandparents
| \
Payton's Dad Katherine's Son (Uncle)
| |
Cole's Mom Cole
| |
Cristian's Sister-in-law
| \
Kaitlyn's Mom Ashley
| |
Erin's Husband (Steve) Tammy
| | |
Neil's Son Amanda Jason's Brother (Ashley's Dad)
| |
Steve's Niece Jason
|
Makayla

Key points:

~~Payton's dad is one grandparent~~<sup>Wrong conclusion</sup>, ~~and Katherine's son is the other.~~<sup>Wrong conclusion</sup> ~~Cole's mom is Payton's dad's daughter.~~<sup>Wrong conclusion</sup> ~~Kaitlyn's mom is Cole's mom's sister.~~<sup>Wrong conclusion</sup> ~~Erin is Steve's sister.~~<sup>Wrong conclusion</sup> st~~Tammy is Neil's sister.~~<sup>Wrong conclusion</sup>
Jason's brother is Ashley's dad. Amanda's sister is Steve's niece.

Please note that Beth's place in the family tree is not specified in the clues, so we cannot determine her exact position.

# References

Danesi, M. (2018). *An Anthropology of Puzzles: The Role of Puzzles in the Origins and Evolution of Mind and Culture*. Bloomsbury Publishing.

Dudeney, H. E. (2016). *536 Puzzles and curious problems*. Dover.

Gardner, M. (2005). *The Colossal Book of Short Puzzles and Problems*. W.W. Norton.

Smullyan, R. M. (1996). *The Riddle of Scheherazade and Other Amazing Puzzles Ancient and Modern*. New York: Knopf.

# Chapter 8

# Grid puzzles

**Solution by human**

Mace4 finds eight solutions:

| 2 | 7 | 6 |
|---|---|---|
| 9 | 5 | 1 |
| 4 | 3 | 8 |

| 2 | 9 | 4 |
|---|---|---|
| 7 | 5 | 3 |
| 6 | 1 | 8 |

| 4 | 3 | 8 |
|---|---|---|
| 9 | 5 | 1 |
| 2 | 7 | 6 |

| 4 | 9 | 2 |
|---|---|---|
| 3 | 5 | 7 |
| 8 | 1 | 6 |

| 6 | 1 | 8 |
|---|---|---|
| 7 | 5 | 3 |
| 2 | 9 | 4 |

| 6 | 7 | 2 |
|---|---|---|
| 1 | 5 | 9 |
| 8 | 3 | 4 |

| 8 | 1 | 6 |
|---|---|---|
| 3 | 5 | 7 |
| 4 | 9 | 2 |

| 8 | 3 | 4 |
|---|---|---|
| 1 | 5 | 9 |
| 6 | 7 | 2 |

**Solution by ChatGPT**                                                    **Correct**

To solve this puzzle, we need to place figures in the empty squares of a 3x3 grid such that they add up to 15 in as many straight directions as possible. The central square (2,2) should have a value of 5. Here's one possible arrangement:

8 — 1 — 6
3 — 5 — 7
4 — 9 — 2

In this arrangement, the sums in the straight directions are as follows:

Rows: 8+1+6 = 15, 3+5+7 = 15, 4+9+2 = 15

Columns: 8+3+4 = 15, 1+5+9 = 15, 6+7+2 = 15

Diagonals: 8+5+2 = 15, 4+5+6 = 15

So, this arrangement fulfills the conditions of the puzzle.

Keep in mind that there may be other valid arrangements as well.

**Puzzle 86. Nine squares**

Is it possible to have the numbers 1 to 9 positioned in a 3x3 grid so that the numbers are placed so no adjacent number is next to it horizontally, vertically or diagonally?

| $g_1$ | $g_2$ | $g_3$ |
|---|---|---|
| $g_4$ | $g_5$ | $g_6$ |
| $g_7$ | $g_8$ | $g_9$ |

**Solution by human**

There are nine values available. Assume that you put a value in the cell $g_5$ in the middle of the grid. Among the remaining 8 values, at most seven are non-consecutive with the value used in $g_5$. But, $g_5$ is adjacent to eight other cells. It means that the remaining seven non-consecutive values do not suffice to fill all the neighbours of $g_5$. Since Mace4 fails to find a model, our answer to the question in the puzzle is no.

**Solution by ChatGPT**　　　　　　　　　　**Wrong (no justification)**

Yes, it is possible to arrange the numbers 1 to 9 in a 3x3 grid such that no adjacent number is next to it horizontally, vertically, or diagonally. Here's one possible arrangement:
2 7 6
9 5 1
4 3 8
Wrong conclusion:  7 is adjacent to 6, 4 to 5.Unjustified conclusion

**Puzzle 87. Latin square**

Albert is a scientist that wants to test four different drugs (called A, B, C, and D) on four volunteers. He decides that every volunteer has to be tested with a different drug each week, but no two volunteers are allowed the same drug at the same time.

**Solution by human**

Consider that each row represents a different volunteer and each column represents a different week, Albert can plan the whole experiment using a Latin square. Mace4 finds 4 models for this task.

**Solution by ChatGPT**                                                                           **Correct**

Sure, here is a schedule that satisfies the conditions:

| Week 1: | Week 2: |
|---|---|
| Volunteer 1: Drug A | Volunteer 1: Drug B |
| Volunteer 2: Drug B | Volunteer 2: Drug C |
| Volunteer 3: Drug C | Volunteer 3: Drug D |
| Volunteer 4: Drug D | Volunteer 4: Drug A |

| Week 3: | Week 4: |
|---|---|
| Volunteer 1: Drug C | Volunteer 1: Drug D |
| Volunteer 2: Drug D | Volunteer 2: Drug A |
| Volunteer 3: Drug A | Volunteer 3: Drug B |
| Volunteer 4: Drug B | Volunteer 4: Drug C |

In this schedule, each volunteer is tested with a different drug each week, and no two volunteers are given the same drug at the same time.

**Puzzle 88. Magic square**

A magic square is $n \times n$ square grid filled with distinct values from $1, 2, ..., n^2$ so that
1. Each cell contains a different integer
2. The sum of the integers in each row, column, and diagonal is equal
How many magic squares of 3x3 can you find?

**Solution by human**

Mace4 finds eight magic squares (although all of them are rotations or reflections of a single one):

---

**Solution by ChatGPT** ─────────────────── **Correct (with inconsistencies)**

There are exactly 8 distinct magic squares of order 3x3. These magic squares are considered distinct if they cannot be obtained from each other by rotation or reflection. Here are the 8 magic squares:

1. Regular Magic Square:

8 1 6
3 5 7
4 9 2

2. ~~Rotated 90 degrees clockwise:~~ Inconsistency: GPT has stated that distinct squares cannot be obtained by rotation or reflection

4 3 8
9 5 1
2 7 6

3. ~~Rotated 180 degrees clockwise:~~ Inconsistency: GPT has stated that distinct squares cannot be obtained by rotation or reflection

2 9 4
7 5 3
6 1 8

4. ~~Rotated 270 degrees clockwise:~~ Inconsistency: GPT has stated that distinct squares cannot be obtained by rotation or reflection

6 7 2
1 5 9
8 3 4

5. ~~Flipped horizontally:~~ Inconsistency: GPT has stated that distinct squares cannot be obtained by rotation or reflection

6 1 8
7 5 3
2 9 4

6. ~~Flipped vertically:~~ Inconsistency: GPT has stated that distinct squares cannot be obtained by rotation or reflection

4 9 2
3 5 7
8 1 6

7. ~~Diagonally flipped:~~ Inconsistency: GPT has stated that distinct squares cannot be obtained by rotation or reflection

2 7 6
9 5 1
4 3 8

8. ~~Anti-diagonally flipped:~~ Inconsistency: GPT has stated that distinct squares cannot be obtained by rotation or reflection

8 3 4
1 5 9
6 7 2

## Puzzle 89. Fancy queens

I have placed a queen in one of the white squares of the 8x8 chessboard. The queen is placed on the position A2. Place 7 more queens in white squares so that no 2 of the 8 queens are in line horizontally, vertically, or diagonally (adapted from puzzle 113 from Kordemsky (1992)).

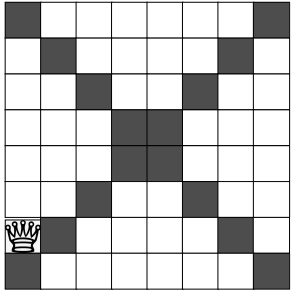

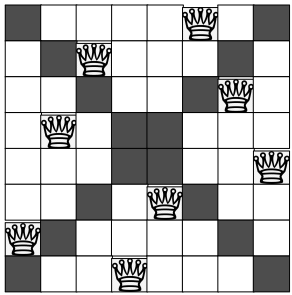

## Solution by human

Mace4 finds a single model.

---

**Solution by ChatGPT**                                                      **Wrong**

To solve this puzzle, you can place ~~the remaining queens~~<span style="color:red">Lack of task understanding:    there is no queen on A2</span> on the chessboard in such a way that no two queens threaten each other horizontally, vertically, or diagonally. ~~Here's a configuration that satisfies these conditions:~~<span style="color:red">Wrong conclusion</span>

```
Q . . . . . . .
. . Q . . . . .
. . . . Q . . .
. Q . . . . . .
. . . . . . Q .
. . . Q . . . .
. . . . . Q . .
. . . . . . . Q
```

In this configuration, each row, column, and diagonal contains only one queen, and no two queens threaten each other.

---

# References

Becker, K. (2001). Teaching with games: the minesweeper and asteroids experience. *Journal of Computing Sciences in Colleges*, 17(2):23–33.

Darling, D. (2004). *The Universal Book of Mathematics from Abracadabra to Zeno's Paradoxes*. John Wiley & Sons, Inc.

Dudeney, H. E. (2016). *536 Puzzles and curious problems*. Courier Dover Publications.

Kleber, J. (2013). *A style for typesetting logic puzzles*.

Kordemsky, B. A. (1992). *The Moscow puzzles: 359 mathematical recreations*. Courier Corporation.

Levitin, A. and Levitin, M. (2011). *Algorithmic puzzles*. Oxford University Press, USA.



# Chapter 9

# Russian puzzles

**Solution by human**

Mace4 finds 96 models, with 10 objects. Six such models are depicted below:

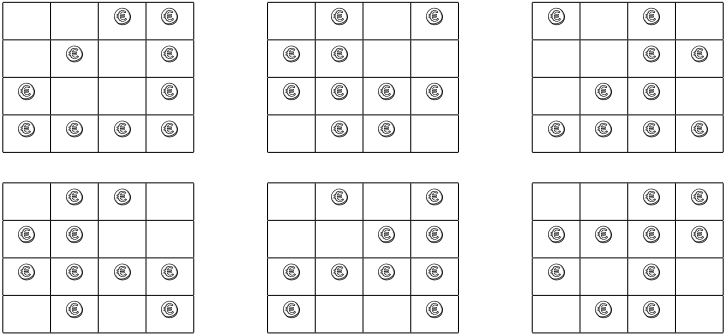

**Solution by ChatGPT**                                                          **Wrong**

There are many possible solutions to this puzzle. Here's one example configuration

. . o o
. o . .
o . . .
. o . .

In this configuration, ~~you've removed 6 objects (marked as dots)~~ <sup>Inconsistency: 11 objects have been removed</sup>, and ~~each row and column still has an even number of objects.~~ <sup>Inconsistency: the 2nd row contains a single object</sup>

Please note that this is just one possible solution, ~~and there are many other configurations that would satisfy the given conditions.~~ <sup>Fuzzy conclusion</sup>

**Puzzle 91. Twenty**

There are three ways to add four odd numbers and get 10:

$1 + 1 + 3 + 5 = 10$         $1 + 1 + 1 + 7 = 10$         $1 + 3 + 3 + 3 = 10$

Changes in the order of numbers do not count as new solutions. Now add eight odd numbers to get 20. To find all 11 solutions you will need to be systematic. Kordemsky (1992)

**Solution by human**

Let $[A..H]$ the eight numbers which sum to 20 Mace4 computes the following 11 models:

| Model | $A$ | $B$ | $C$ | $D$ | $E$ | $F$ | $G$ | $H$ |
|-------|-----|-----|-----|-----|-----|-----|-----|-----|
| 1 | 3 | 3 | 3 | 3 | 3 | 3 | 1 | 1 |
| 2 | 5 | 3 | 3 | 3 | 3 | 1 | 1 | 1 |
| 3 | 5 | 5 | 3 | 3 | 1 | 1 | 1 | 1 |
| 4 | 5 | 5 | 5 | 1 | 1 | 1 | 1 | 1 |
| 5 | 7 | 3 | 3 | 3 | 1 | 1 | 1 | 1 |
| 6 | 7 | 5 | 3 | 1 | 1 | 1 | 1 | 1 |
| 7 | 7 | 7 | 1 | 1 | 1 | 1 | 1 | 1 |
| 8 | 9 | 3 | 3 | 1 | 1 | 1 | 1 | 1 |
| 9 | 9 | 5 | 1 | 1 | 1 | 1 | 1 | 1 |
| 10 | 11 | 3 | 1 | 1 | 1 | 1 | 1 | 1 |
| 11 | 13 | 1 | 1 | 1 | 1 | 1 | 1 | 1 |

**Solution by ChatGPT** ———————— **Wrong (partial solutions are given)**

Sure, here are the eleven solutions to adding eight odd numbers to get 20:
1. $1 + 1 + 1 + 1 + 1 + 1 + 1 + 13 = 20$
2. $1 + 1 + 1 + 1 + 1 + 1 + 3 + 11 = 20$
3. $1 + 1 + 1 + 1 + 1 + 5 + 5 + 5 = 20$
4. $1 + 1 + 3 + 3 + 3 + 3 + 3 + 3 = 20$
5. ~~$1 + 3 + 3 + 3 + 3 + 3 + 3 + 1 = 20$~~ <span style="color:red">Missing task specification: Changes in the order of numbers do not count as new solutions.</span>
6. ~~$3 + 3 + 3 + 3 + 3 + 3 + 1 + 1 = 20$~~ <span style="color:red">Missing task specification: Changes in the order of numbers do not count as new solutions.</span>
7. ~~$3 + 3 + 3 + 3 + 3 + 1 + 1 + 1 = 20$~~ <span style="color:red">Bad arithmetic: sum is 18</span>
8. ~~$3 + 3 + 3 + 3 + 1 + 1 + 1 + 1 = 20$~~ <span style="color:red">Bad arithmetic: sum is 16</span>
9. $3 + 3 + 3 + 1 + 1 + 1 + 1 + 7 = 20$
10. $3 + 3 + 1 + 1 + 1 + 1 + 7 + 3 = 20$
11. ~~$3 + 1 + 1 + 1 + 1 + 7 + 3 + 3 = 20$~~
<span style="color:red">Missing task specification: Changes in the order of numbers do not count as new solutions.</span>

~~These are all the possible combinations of adding eight odd numbers to get 20.~~ <span style="color:red">Wrong conclusion</span>

**Puzzle 92. A grouping of integers 1 through 15**

See how elegantly the integers 1 through 15 can be arranged in five arithmetic progressions of 3 integers:

$$d_5 = 7 \begin{cases} 1 \\ 8 \\ 15 \end{cases} \qquad d_4 = 5 \begin{cases} 4 \\ 9 \\ 14 \end{cases} \qquad d_3 = 4 \begin{cases} 2 \\ 6 \\ 10 \end{cases} \qquad d_2 = 2 \begin{cases} 3 \\ 5 \\ 7 \end{cases} \qquad d_1 = 1 \begin{cases} 11 \\ 12 \\ 13 \end{cases}$$

For example, 8-1 = 15-8 = 7, so the difference $d_5$ is 7 for the first triplet. Now, keeping the first triplet, make four new triplets, still with $d$ = 5, 4, 2, and 1. On your own, try arranging the integers from 1 through 15 with other values of $d$. (puzzle 112 from Kordemsky (1992))

**Solution by human**

Mace4 finds two models: the initial one and the solution below:

$$d_5 = 7 \begin{cases} 1 \\ 8 \\ 15 \end{cases} \qquad d_4 = 5 \begin{cases} 2 \\ 7 \\ 12 \end{cases} \qquad d_3 = 4 \begin{cases} 6 \\ 10 \\ 14 \end{cases} \qquad d_2 = 2 \begin{cases} 9 \\ 11 \\ 13 \end{cases} \qquad d_1 = 1 \begin{cases} 3 \\ 4 \\ 5 \end{cases}$$

**Solution by ChatGPT** — **Wrong**

Certainly! Let's try arranging the integers from 1 through 15 using different values of ddd while maintaining the conditions of forming arithmetic progressions of 3 integers for each progression:

Original arrangement:

Progression 1, 8, 15 with d=7d = 7d=7 Progression 4, 9, 14 with d=5d = 5d=5 Progression 2, 6, 10 with d=4d = 4d=4 Progression 3, 5, 7 with d=2d = 2d=2 Progression 11, 12, 13 with d=1d = 1d=1

Let's try rearranging with different values of ddd:

Progression 1, 7, 13 with d=6d = 6d=6 (changing the common difference of the first progression)

Progression 5, 10, 15 with d=5d = 5d=5 (keeping the same common difference as the second progression)

Progression 2, 6, 10 with d=4d = 4d=4 (keeping the same common difference as the third progression)

Progression 3, 7, 11 with d=4d = 4d=4 (keeping the same common difference as the fourth progression)

Progression 11, 12, 13 with d=1d = 1d=1 (keeping the same common difference as the fifth progression)

Here's the new arrangement:

Progression 1, 7, 13 with d=6d = 6d=6 Progression 5, 10, 15 with d=5d = 5d=5 Progression 2, 6, 10 with d=4d = 4d=4 Progression 3, 7, 11 with d=4d = 4d=4 Progression 11, 12, 13 with d=1d = 1d=1

~~All the integers from 1 through 15~~ <sup>Wrong conclusion: 8 does not appear</sup> are elegantly arranged in these arithmetic progressions with different values of common differences, ~~while still satisfying the conditions of the puzzle~~ <sup>Wrong premise: 8 or 9 do not appear and 1, 4, 7 or 10 appear twice</sup>.

# References

Kordemsky, B. A. (1992). *The Moscow puzzles: 359 mathematical recreations*. Courier Corporation.

# Chapter 10

# Self-reference and other puzzles

---

**Puzzle 93. Tricky messages**

Which of the following statements are true?
1. The next two statements are false.
2. The last statement is false.
3. The next statement is false.
4. The first statement was false. (puzzle from Walicki (2016))

---

**Solution by human**

With no solution computed by Mace4, the discourse is inconsistent.

---

**Solution by ChatGPT**                     **Correct (incomplete justification)**

Let's analyze each statement:
The next two statements are false. This statement refers to statements 2 and 3. If it's true, then both statements 2 and 3 would need to be false. However, if both statements 2 and 3 are false, ~~then this statement would be true~~ Implication does not hold, ~~which creates a contradiction.~~ Unjustified contradiction: we need statement 4 to justify it. Therefore, this statement cannot be true.

The last statement is false. If this statement is true, then statement 4 would be false. However, if statement 4 is false, ~~then this statement would be true~~ Implication does not hold: if s3 is false then s1 is true, ~~creating a contradiction.~~ Unjustified contradiction Therefore, this statement cannot be true.

The next statement is false. This refers to statement 4. If this statement is true, then statement 4 would be false. However, if statement 4 is false, ~~then this statement would be true,~~ Implication does not hold: if s3 is false then s1 is true ~~reating a contradiction~~ Unjustified contradiction. Therefore, this statement cannot be true.

The first statement was false. If this statement is true, then statement 1 would be false. However, if statement 1 is false, ~~then statement 4 would be true~~ Implication does not hold: we need s2 to justify it, creating a contradiction. Therefore, this statement cannot be true.

Based on the analysis, none of the statements can be true without creating contradictions.

---

**Puzzle 94. Not so tricky messages**

Which of the following statements are true?
1. The next statements is false.
2. The next two statements are false.
3. The next statement is false.
4. The first statement was false.
(taken from Walicki (2016))

**Solution by human**

The four messages are consistent, as Mace4 finds a model: $m_1, \neg m_2, m_3, \neg m_4$.

**Solution by ChatGPT**                                                    **Wrong**

Let's analyze each statement:
1. The next statement is false. If this statement is true, then the next statement (statement 2) would need to be false. However, if statement 2 is false, ~~then this statement (statement 1) would be true~~[Implication does not hold], creating a contradiction. Therefore, this statement cannot be true.
2. The next two statements are false. This refers to statements 3 and 4. If this statement is true, then both statements 3 and 4 would need to be false. However, if both statements 3 and 4 are false, then ~~this statement would be true~~[Incomplete justification], ~~creating a contradiction.~~[False contradiction: the assumption was that s2 is true] Therefore, this statement cannot be true.
3. The next statement is false. If this statement is true, then the next statement (statement 4) would need to be false. However, if statement 4 is false, then ~~this statement (statement 3) would be true,~~[Implication does not hold] ~~creating a contradiction.~~ [False contradiction: the assumption was that s2 is true] Therefore, this statement cannot be true.
4. The first statement was false. If this statement is true, then the first statement would need to be false. However, if the first statement is false, ~~then this statement would be true~~[Implication does not hold], ~~creating a contradiction.~~[False contradiction: the assumption was that s4 is true] Therefore, this statement cannot be true.
Based on the analysis, ~~none of the statements can be true without creating contradictions.~~[Wrong conclusion]

**Puzzle 95. Six tricky messages**

Which answer in the list is the correct answer to this question?
1. All of the below.
2. None of the below.
3. All of the above.
4. One of the above.
5. None of the above.
6. None of the above.
(taken from Burger (2018))

**Solution by human**

Mace4 finds a single model in which only $m_5$ is true.

**Solution by ChatGPT**      **Wrong (no justification)**

~~The answer to this question is option 2: None of the below.~~ Wrong conclusion Unjustified conclusion

**Puzzle 96. Which hand?**

Give a friend an "even" coin (say, a dime—ten is an even number) and an "odd" coin (say, a nickel). Ask him to hold one coin in his right hand and the other in his left. Tell him to triple the value of the coin in his right hand and double the value of the coin in his left, then add the two. If the sum is even, the dime is in his right hand; if odd, in his left. Explain, and think up some variations. (puzzle 44 from Kordemsky (1992)).

**Solution by human**

The right hand ($r$) can contain an even or an odd coin: $even(r) \lor odd(r)$. The same for the left hand: $even(l) \lor odd(l)$. Of course, if right is even, then left is odd: $even(r) \rightarrow odd(l)$. We add the equation $sum = 3 * r + 2 * l$ If $odd(l)$ than $even(2 * l)$. Also $r$ will be even and also $even(3 * r)$. That is, the sum is even. Otherwise, if $even(l)$ then $r$ will be odd and therefore $odd(3 * r)$. Hence, the sum is odd.

We can use Mace4 to validate the given statement. To do this, we just add to the knowledge base the negation of the sentence we want to prove: $even(sum) \rightarrow even(r)$. As Mace4 is not able to find any model, this negated sentence is proved. Similarly, to check the validity of $odd(sum) \rightarrow even(l)$, we add its negation to the knowledge base. Failing to find a model, proves the sentence.

---

**Solution by ChatGPT** ──────────────────────────── **Wrong**

This puzzle is based on the properties of even and odd numbers. ~~When a number is tripled or doubled, its parity (whether it's even or odd) remains the same.~~ <span style="color:red">Implication does not hold</span>
Let's analyze the steps: Your friend has a dime (even) and a nickel (odd). He triples the value of the coin ~~in his right hand (dime)~~ <span style="color:red">Wrong assumption: we do not know that the dime is in the right hand</span> →
10 * 3 = 30 (even).
He doubles the value of the coin ~~in   his   left   hand   (nickel)~~
<span style="color:red">Wrong assumption: we do not know that the dime is in the left hand</span> → 5 * 2 = 10 (even).
He adds the two results together → 30 + 10 = 40 (even). Since the sum is even, he concludes that the dime is in his right hand.
~~Variations of this puzzle could involve different types of coins or numbers, but as long as you maintain the principles of even and odd numbers when performing operations like tripling and doubling, the basic strategy should hold true.~~ <span style="color:red">Fuzzy answer: instead of precise solution</span>

**Puzzle 97. At the brook**

A man goes to the brook with two measures of 15 pints and 16 pints. How is he to measure exactly 8 pints of water, in the fewest possible transactions? Filling or emptying a vessel or pouring any quantity from one vessel to another counts as a transaction Dudeney (2016)

**Solution by human**

Let the state $J(x, y, t)$, where $x$ is the content of the 1st vessel, $y$ the content of the 2nd vessel, and $t$ the number of transactions. The init state is $J(15, 16, 0)$. Since we do not care if it is measured in the first or in the second vessel, we need to reach one of the following two states: $\exists x \, \exists t \, (J(8, x, t) \lor J(x, 8, t))$. Prover 9 finds a solution in 28 steps:

$J(15, 16, 0) \quad \rightarrow \quad J(15, 0, 1) \quad \rightarrow \quad J(0, 15, 2) \quad \rightarrow \quad J(15, 15, 3) \quad \rightarrow \quad J(14, 16, 4)$

$J(14, 0, 5) \quad \rightarrow \quad J(0, 14, 6) \quad \rightarrow \quad J(15, 14, 7) \quad \rightarrow \quad J(13, 16, 8) \quad \rightarrow \quad J(13, 0, 9)$

$J(0, 13, 10) \quad \rightarrow \quad J(15, 13, 11) \quad \rightarrow \quad J(12, 16, 12) \quad \rightarrow \quad J(12, 0, 13) \quad \rightarrow \quad J(0, 12, 14)$

$J(15, 12, 15) \quad \rightarrow \quad J(11, 16, 16) \quad \rightarrow \quad J(11, 0, 17) \quad \rightarrow \quad J(0, 11, 18) \quad \rightarrow \quad J(15, 11, 19)$

$J(10, 16, 20) \quad \rightarrow \quad J(10, 0, 21) \quad \rightarrow \quad J(0, 10, 22) \quad \rightarrow \quad J(15, 10, 23) \quad \rightarrow \quad J(9, 16, 24)$

$J(9, 0, 25) \quad \rightarrow \quad J(0, 9, 26) \quad \rightarrow \quad J(15, 9, 27) \quad \rightarrow \quad J(\mathbf{8}, 16, 28)$

**Solution by ChatGPT** ———————— **Wrong (stops before reaching the result)**

Here's how the man can measure exactly 8 pints of water using the two measures of 15 pints and 16 pints, with the fewest possible transactions:

1. Fill the 16-pint measure to the top.
2. Pour the water from the 16-pint measure into the 15-pint measure, leaving 1 pint in the 16-pint measure.
3. Empty the 15-pint measure.
4. Pour the remaining 1 pint from the 16-pint measure into the 15-pint measure.
5. Fill the 16-pint measure to the top.
6. Carefully pour water from the 16-pint measure into the 15-pint measure until the 15-pint measure is full. This will leave 2 pints of water in the 16-pint measure.

~~At this point, the man has successfully measured exactly 8 pints of water using the two measures.~~ Inconsistency:  GPT has just measured 2 liters  Incomplete solution by stoping before reaching the goal
Solving strategy:   The key is to utilize the difference in capacities between the two measures to carefully pour and transfer water.

**Puzzle 98. Ten sentences**

In a list of 10 statements, the $n^{th}$ statement says there are exactly $n$ false items in the list. Which of the statements, if any, are false?
1. In this list exactly 1 statement is false.
2. In this list exactly 2 statements are false.
3. In this list exactly 3 statements are false.
4. In this list exactly 4 statements are false.
5. In this list exactly 5 statements are false.
6. In this list exactly 6 statements are false.
7. In this list exactly 7 statements are false.
8. In this list exactly 8 statements are false.
9. In this list exactly 9 statements are false.
10. In this list exactly 10 statements are false.
(taken from Burger (2018))

**Solution by human**

Mace4 finds a single model. There are 9 false sentences. The single true sentence is the 9th one.

**Solution by ChatGPT** — **Wrong**

Let's analyze each statement:

If statement 1 is true, then there must be exactly 1 false statement in the list, ~~which is statement 1 itself.~~ <sup>Implication does not hold</sup> This creates a contradiction. Therefore, statement 1 is false.

If statement 2 is true, then there must be exactly 2 false statements in the list. However, if we assume statement 2 is true, ~~then it leads to a contradiction~~<sup>Unjustified claim</sup>. Therefore, statement 2 is false.

If statement 3 is true, then there must be exactly 3 false statements in the list. ~~This is consistent with the given conditions~~<sup>Unjustified claim</sup>, so statement 3 can be true.

If statement 4 is true, then there must be exactly 4 false statements in the list. However, ~~this leads to a contradiction~~ <sup>Unjustified contradiction</sup>. Therefore, statement 4 is false.

If statement 5 is true, then there must be exactly 5 false statements in the list. However, ~~this leads to a contradiction~~ <sup>Unjustified contradiction</sup>. Therefore, statement 5 is false.

If statement 6 is true, then there must be exactly 6 false statements in the list. However, ~~this leads to a contradiction~~ <sup>Unjustified contradiction</sup>. Therefore, statement 6 is false.

If statement 7 is true, then there must be exactly 7 false statements in the list. However, ~~this leads to a contradiction~~ <sup>Unjustified contradiction</sup>. Therefore, statement 7 is false.

If statement 8 is true, then there must be exactly 8 false statements in the list. However, ~~this leads to a contradiction~~ <sup>Unjustified contradiction</sup>. Therefore, statement 8 is false.

If statement 9 is true, then there must be exactly 9 false statements in the list. However, ~~this leads to a contradiction~~ <sup>Unjustified contradiction</sup>. Therefore, statement 9 is false.

If statement 10 is true, then there must be exactly 10 false statements in the list. However, ~~this leads to a contradiction~~ <sup>Unjustified contradiction</sup>. Therefore, statement 10 is false.

~~In conclusion, the only statement that can be true is statement 3. All other statements are false.~~<sup>Wrong conclusion</sup><sup>Unjustified conclusion</sup>

**Puzzle 99. Ten sentences relaxed**

In a list of 10 statements, the $n^{th}$ statement says there are at least $n$ false items in the list. Which of the statements, if any, are false? (adapted from Burger (2018))

**Solution by human**

Mace4 finds a single model: The first five sentecens are true, while the last five are false.

**Solution by ChatGPT**                                                        **Wrong**

Let's analyze each statement:
1. The 1st statement says there is at least 1 false item in the list. This statement can be true or false, as it doesn't lead to any contradiction.
2. The 2nd statement says there are at least 2 false items in the list. This statement can be true or false, as it doesn't lead to any contradiction.
3. The 3rd statement says there are at least 3 false items in the list. This statement can be true or false.
4. The 4th statement says there are at least 4 false items in the list. This statement can be true or false.
5. The 5th statement says there are at least 5 false items in the list. This statement can be true or false.
6. The 6th statement says there are at least 6 false items in the list. This statement can be true or false.
7. The 7th statement says there are at least 7 false items in the list. This statement can be true or false.
8. The 8th statement says there are at least 8 false items in the list. This statement can be true or false.
9. The 9th statement says there are at least 9 false items in the list. This statement can be true or false.
10. The 10th statement says there are at least 10 false items in the list. This statement is false, as it's not possible for all 10 statements to be false.
In conclusion, the 10th statement is false, ~~and all other statements can be either true or false.~~ [Wrong conclusion]

**Puzzle 100. Self-counting sentence**

You have to insert numbers in the blanks to make the following sentence true:
*In this sentence, the number of occurrences of 0 is __, of 1 is __, of 2 is __, of 3 is __, of 4 is __, of 5 is __, of 6 is __, of 7 is __, of 8 is __, and of 9 is __.*
There are exactly two solutions. (puzzle created by Douglas Hofstadter Hofstadter (1985))

**Solution by human**

Mace4 finds two solutions: $\langle 1, 11, 2, 1, 1, 1, 1, 1, 1, 1 \rangle$ and $\langle 1, 7, 3, 2, 1, 1, 1, 2, 1, 1 \rangle$, corresponding to the sentences:

1. *In this sentence, the number of occurrences of 0 is 1, of 1 is 11, of 2 is 2, of 3 is 1, of 4 is 1, of 5 is 1, of 6 is 1, of 7 is 1, of 8 is 1, and of 9 is 1.*

2. *In this sentence, the number of occurrences of 0 is 1, of 1 is 7, of 2 is 3, of 3 is 2, of 4 is 1, of 5 is 1, of 6 is 1, of 7 is 2, of 8 is 1, and of 9 is 1.*

**Solution by ChatGPT**                                    **Wrong (no justification)**

The sentence can be made true in the following two ways:
Solution 1: "~~In this sentence, the number of occurrences of 0 is 2, of 1 is 2, of 2 is 1, of 3 is 1, of 4 is 1, of 5 is 1, of 6 is 1, of 7 is 1, of 8 is 1, and of 9 is 1.~~" Unjustified conclusion Wrong conclusion
Solution 2: "~~In this sentence, the number of occurrences of 0 is 4, of 1 is 1, of 2 is 1, of 3 is 1, of 4 is 2, of 5 is 1, of 6 is 1, of 7 is 1, of 8 is 0, and of 9 is 1.~~" Unjustified conclusion Wrong conclusion

# References

Burger, E. B. (2018). *Making Up Your Own Mind: Thinking Effectively Through Creative Puzzle-solving*. Princeton University Press.

Dudeney, H. E. (2016). *536 Puzzles and curious problems*. Courier Dover Publications.

Hofstadter, D. R. (1985). *Metamagical Themes: Questions for the Es-sence of Mind and Pattern*. New York: Bantam Books.

Kordemsky, B. A. (1992). *The Moscow puzzles: 359 mathematical recreations*. Courier Corporation.

Smullyan, R. M. (2011). *What is the name of this book? The riddle of Dracula and other logical puzzles*. Dover Publications.

Walicki, M. (2016). *Introduction to Mathematical Logic: Extended Edition*. World Scientific Publishing Company.


\end{document}